\documentclass{article}
\PassOptionsToPackage{numbers, compress}{natbib}

\usepackage[main, final]{neurips_2026}

\usepackage[utf8]{inputenc} %
\usepackage[T1]{fontenc}    %
\usepackage{hyperref}       %
\usepackage{url}            %
\usepackage{booktabs}       %
\usepackage{amsfonts}       %
\usepackage{nicefrac}       %
\usepackage{microtype}      %
\usepackage[table]{xcolor}  %
\usepackage{amsmath}
\usepackage{adjustbox}
\usepackage{graphicx}
\usepackage{subcaption}
\usepackage{multirow}
\usepackage{float}
\usepackage{etoc}
\usepackage{wrapfig}
\usepackage{pifont}
\usepackage{caption}
\usepackage{enumitem}

\definecolor{icmlblue}{rgb}{0.21,0.49,0.74}
\definecolor{myorchid}{RGB}{224,149,219}
\definecolor{mylavender}{RGB}{151,156,250}
\definecolor{myorange}{RGB}{255,171,64}
\definecolor{mygreen}{RGB}{75,184,52}

\title{CORAL: Correspondence Alignment for Improved Virtual Try-On}

\author{%
  \begin{tabular}[t]{c}
  \textbf{Jiyoung Kim}$^{1}$ \quad
  \textbf{Youngjin Shin}$^{1}$ \quad
  \textbf{Siyoon Jin}$^{1}$ \quad
  \textbf{Dahyun Chung}$^{1}$ \quad
  \textbf{Jisu Nam}$^{1}$ \\
  \textbf{Tongmin Kim}$^{1}$ \quad
  \textbf{Jongjae Park}$^{2}$ \quad
  \textbf{Hyeonwoo Kang}$^{2}$ \quad
  \textbf{Seungryong Kim}$^{1}$ \\[0.5em]
  \mdseries $^{1}$KAIST AI \qquad $^{2}$NC AI \\[0.3em]
  \mdseries \url{https://cvlab-kaist.github.io/CORAL/}
  \end{tabular}
}

\begin{document}

\maketitle
{
    \captionsetup{type=figure}
    \setlength{\abovecaptionskip}{2pt}
    \setlength{\belowcaptionskip}{0pt}
    \vspace{-10pt}
    \noindent\centering
    \setkeys{Gin}{width=\linewidth}
    \includegraphics{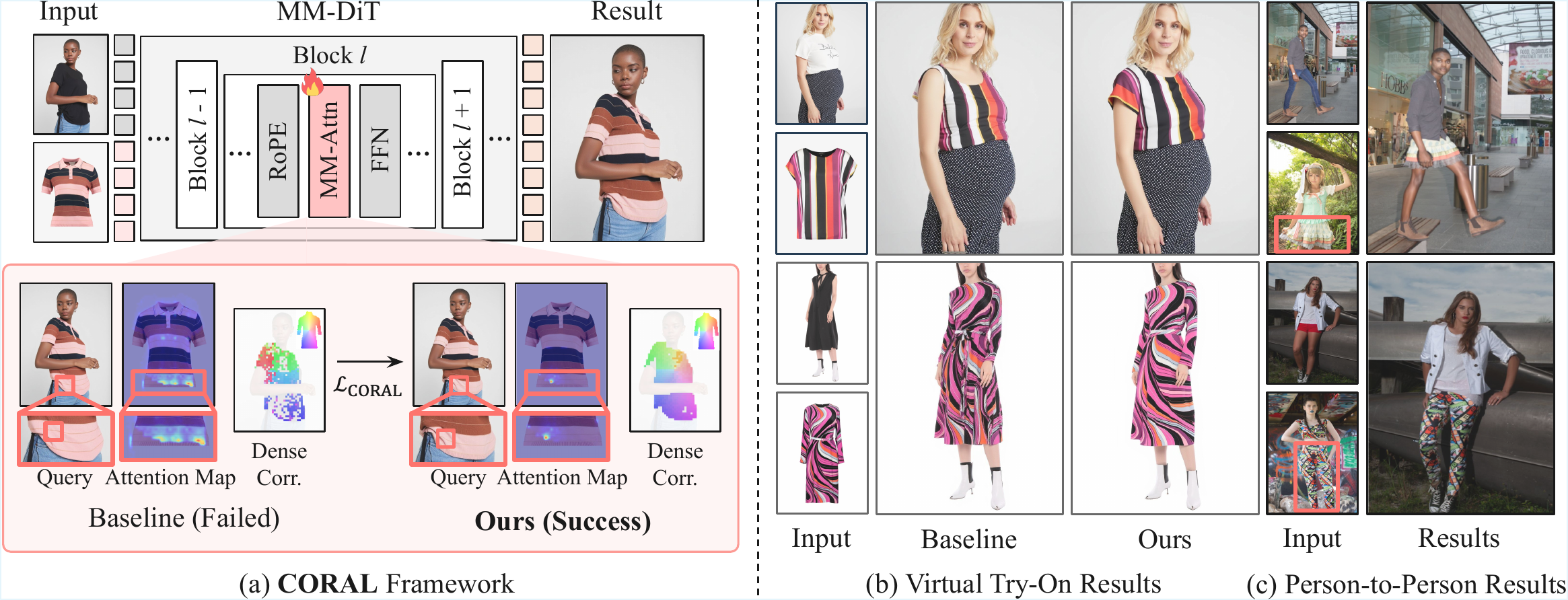}
    \vspace{-5pt}
    \captionof{figure}{\textbf{Teaser.} (a) CORAL is a Diffusion Transformer-based framework that explicitly enhances person$\rightarrow$garment correspondence within the 3D attention. This leads to more accurate local details in the generated results, such as fewer artifacts like duplicated garment hems, as shown in the baseline. (b) Virtual Try-On results on VITON-HD and DressCode, comparing outputs without and with CORAL. (c) Person-to-Person garment transfer results on PPR10K in challenging in-the-wild images.
    }
    \label{teaser}
    \vspace{10pt}
}

\begin{abstract}
Existing methods for Virtual Try-On (VTON) often struggle to accurately transfer garment appearance, especially in unpaired settings where accurate person--garment correspondence is required.
These methods do not explicitly supervise person--garment alignment, leaving correspondence to be learned implicitly within the generation model.
In this paper, we first analyze full self-attention in DiT-based architectures and show that person$\rightarrow$garment query--key matching aligned with local geometric correspondence is closely related to try-on quality.
Building on this insight, we introduce \textbf{\underline{COR}}respondence \textbf{\underline{AL}}ignment (\textbf{CORAL}), a DiT-based framework that explicitly aligns query--key matching with robust external correspondences.
CORAL integrates two complementary components: a correspondence distillation loss that aligns reliable matches with person$\rightarrow$garment attention, and an entropy minimization loss that sharpens the attention distribution.
For evaluation, we further propose a VLM-based evaluation protocol to better reflect human preference.
CORAL consistently improves over the baseline, enhancing both global shape transfer and local detail preservation.
Extensive ablations validate our design choices.
\end{abstract}

\section{Introduction}
\vspace{-5pt}
Given a pair of person and garment images, Virtual Try-On (VTON) aims to synthesize the same person wearing the given garment, while handling large geometric variations between them, including differences in pose, silhouette, and garment type or shape~\cite{choi2024improvingdiffusionmodelsauthentic,zhou2024learningflowfieldsattention}. Therefore, a key challenge in VTON is \textit{\textbf{establishing accurate visual correspondence between the person and the garment}}, as inaccurate matching can lead to suboptimal VTON outputs, including distorted garment shapes, misplaced textures, or incorrect fitting.

\begin{figure*}[t!]
  \centering
  \includegraphics[width=\linewidth]{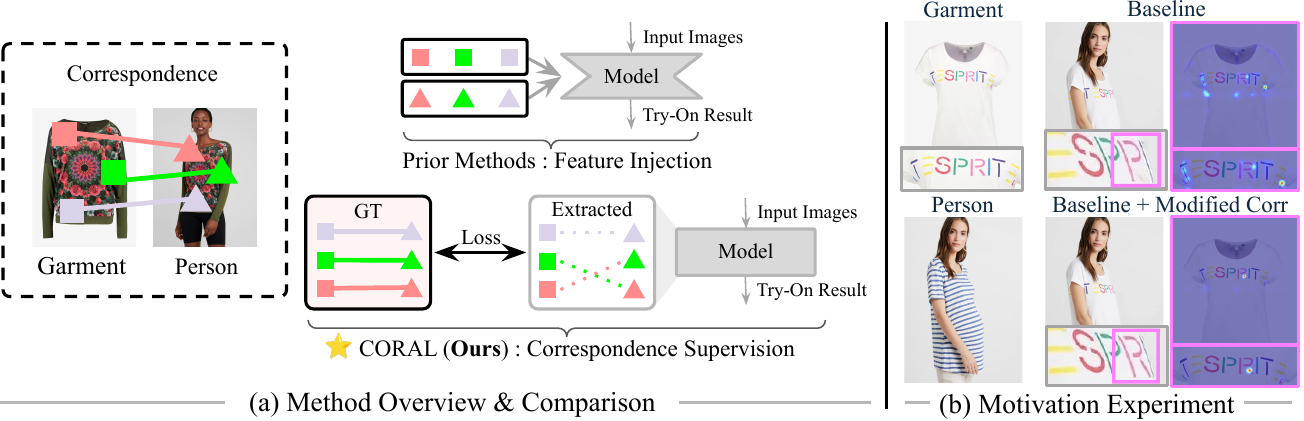}
  \vspace{-15pt}
 \caption{\textbf{Motivation and overview of our approach.} \textcolor{magenta}{Pink} markers denote the query points (the bottom-right corner of the letter ``R''). (a) Prior methods~\cite{wan2025incorporatingvisualcorrespondencediffusion, chen2024wearanywaymanipulablevirtualtryon} inject separate shared features for the garment and person into U-Net-based generative models as auxiliary signals, where correspondence is not explicitly enforced. The matches must therefore be estimated at inference as well. In contrast, our method enforces explicit correspondence supervision in a DiT without modifying the feature representations, and requires no correspondence at inference. (b) We modify the baseline's query-key attention at inference time to focus on pseudo ground-truth correspondences from DINOv3. This intervention recovers the correct text and improves try-on quality. (Best viewed when zoomed in. Full images and experiment details are provided in Appendix~\ref{supple:details_on_motiv_exp}.)
  }
  \vspace{-10pt}
  \label{fig:motivation}
\end{figure*}

Visual correspondence has been a core component of VTON. Early approaches warped garment pixels through geometric transformations~\cite{wang2018characteristicpreservingimagebasedvirtualtryon, choi2021vitonhdhighresolutionvirtualtryon} or learned flow fields~\cite{xie2023gpvtongeneralpurposevirtual}. As illustrated in Fig.~\ref{fig:motivation}a, diffusion-based methods built on U-Net dual-branch architectures incorporate correspondence as an auxiliary conditioning signal, typically by injecting paired point embeddings into the diffusion model~\cite{chen2024wearanywaymanipulablevirtualtryon, wan2025incorporatingvisualcorrespondencediffusion}. However, these methods mainly rely on a reconstruction objective during training, which only implicitly encourages person–garment matching and provides no direct supervision over the correspondence itself, often resulting in misaligned generation. Moreover, these methods take correspondence as an input and thus require it at inference, where accurate matching is far harder to obtain since try-on is unpaired.

Meanwhile, Diffusion Transformers (DiTs)~\cite{peebles2023scalablediffusionmodelstransformers,esser2024scalingrectifiedflowtransformers} have recently gained attention for their strong in-context interactions between input tokens via full 3D attention~\cite{tan2025ominicontrolminimaluniversalcontrol, huang2024incontextloradiffusiontransformers}. Building on this, DiT-based VTON methods adopt a unified design ~\cite{guo2025any2anytryonleveragingadaptiveposition}, concatenating all tokens into a single sequence. This design removes the need for separate dual-branch architectures~\cite{choi2024improvingdiffusionmodelsauthentic, wan2025incorporatingvisualcorrespondencediffusion, chen2024wearanywaymanipulablevirtualtryon, zhou2024learningflowfieldsattention} and enables person and garment tokens to directly interact within self-attention, providing a natural mechanism for person–garment correspondence to emerge.

However, such correspondence is not always reliably established. While the unified DiT design enables rich cross-token interactions and strong try-on synthesis, these interactions are not explicitly optimized for accurate person--garment matching. Since the diffusion loss is defined only on the output, attention tends to rely on appearance similarity, without establishing local geometric correspondence between body positions and garment parts. As a result, the model may produce visually plausible try-on results while relying on imperfect correspondences, often leading to misplaced textures, distorted garment structures, or incorrect fitting, especially when the person and garment differ in pose, scale, or illumination.

To verify the importance of correspondence in DiT-based VTON, we first conduct an inference-time intervention on the baseline model. Specifically, we use DINOv3, a pretrained vision foundation model that provides dense patch-level visual descriptors, to extract high-quality pseudo correspondences between person and garment regions.  We then use these correspondences to locally guide the baseline's person$\rightarrow$garment query--key matching during inference, without updating model weights or modifying feature representations. As shown in Fig.~\ref{fig:motivation}, adding a correspondence bias to the person$\rightarrow$garment query--key logits shifts the attention toward the matched garment regions, improving local garment appearance transfer and overall try-on quality. This suggests that strengthening the internal person--garment matching in DiT self-attention can directly improve VTON results. Motivated by this observation, we explicitly supervise query--key matching during training.

Based on this, we propose a novel DiT-based framework, \textbf{\underline{COR}}respondence \textbf{\underline{AL}}ignment (\textbf{CORAL}), which enhances person–garment correspondence by explicitly supervising query–key matching in the full self-attention of the DiT. Specifically, we introduce two complementary losses: (1) a \textbf{correspondence distillation loss} that aligns reliable local geometric matches obtained from the vision foundation model DINOv3~\cite{siméoni2025dinov3} to person–garment query–key matching in the full self-attention, and (2) an \textbf{entropy minimization loss} that sharpens the attention distribution by minimizing its entropy for localized and precise matching. Since the correspondence serves only as a training target, the model internalizes the matching and requires no correspondence at inference.

We demonstrate the effectiveness of CORAL on standard VTON benchmarks in both paired and unpaired settings. To further evaluate the robustness of our method, we introduce a new evaluation dataset that includes real-world, challenging person-to-person unpaired scenarios. CORAL achieves state-of-the-art performance on SSIM, LPIPS, and FID across all benchmarks in the paired setting, with the largest improvement on the challenging in-the-wild benchmark. We additionally evaluate our method using VLM-based assessment and human evaluation, achieving the best performance among all methods. Extensive ablation studies further validate our design choices.

In summary, our contributions are as follows:
\begin{itemize}[leftmargin=1.5em, itemsep=0.0em]
\item We show that person$\rightarrow$garment query--key matching aligned with local geometric correspondence is closely related to try-on quality in DiT-based VTON, and that correspondence should serve as a verified training target rather than an input required at inference.
\item We introduce {CORAL}, which explicitly enhances person--garment correspondence through cycle-consistent correspondence distillation and attention entropy minimization, supervising both where attention points and how sharply.
\item We achieve consistent gains on standard metrics as well as human and VLM-based evaluations across VTON benchmarks, including the challenging settings we introduce.
\end{itemize}

\section{Related Works}
\label{sec:related_works}
\label{supple:additional_related_works}
\paragraph{Virtual Try-On.} VTON aims to precisely transfer the shape and details of the garment image onto a person image. Several works improve sampling and alignment at inference~\cite{bhunia2023personimagesynthesisdenoising, chong2025catvtonconcatenationneedvirtual, wang2024fldmvtonfaithfullatentdiffusion, li2025enhancingvirtualtryonsynthetic}.
Other efforts~\cite{choi2024improvingdiffusionmodelsauthentic,kim2025promptdresserimprovingqualitycontrollability, xie2023gpvtongeneralpurposevirtual} add structural cues (e.g., parsing maps) to improve robustness against pose variations. Other methods inject global~\cite{kim2023stablevitonlearningsemanticcorrespondence, wan2024improvingvirtualtryongarmentfocused, morelli2023ladivtonlatentdiffusiontextualinversion} or local features~\cite{choi2024improvingdiffusionmodelsauthentic, zhou2024learningflowfieldsattention, xu2024ootdiffusionoutfittingfusionbased} from garment images to enhance detail preservation. Yet, they often lose garment details in unpaired settings that demand robust matching, leading to logo drift or boundary shifts.

\paragraph{Correspondence in Diffusion Models.}
The impressive performance of diffusion models comes from the rich visual representations they learn. Analyses of the internal attention mechanism in both UNet~\cite{nam2024dreammatcher, nam2024diffusionmodeldensematching, jeong2025track4genteachingvideodiffusion, jin2025appearancematchingadapterexemplarbased, hedlin2023unsupervisedsemanticcorrespondenceusing, tang2023emergentcorrespondenceimagediffusion, xiao2024videodiffusionmodelstrainingfree} and DiT-based~\cite{yu2025representationalignmentgenerationtraining, nam2025emergent, jin2025matrixmasktrackalignment, lee2025aligningtextimagediffusion} models show that query--key matching encodes semantic correspondences. Editing and customization studies~\cite{nam2024dreammatcher, cao2023masactrltuningfreemutualselfattention, jin2025appearancematchingadapterexemplarbased, hertz2022prompttopromptimageeditingcross, tumanyan2022plugandplaydiffusionfeaturestextdriven} consistently report that query--key matching captures structure and semantic alignment, while values carry appearance that is routed to attended locations. These findings enable strong layout preservation and controllable appearance transfer, and also support downstream tasks such as tracking~\cite{jeong2025track4genteachingvideodiffusion, nam2025emergent}.

\paragraph{Correspondence in VTON.}
Several VTON works~\cite{wan2025incorporatingvisualcorrespondencediffusion, chen2024wearanywaymanipulablevirtualtryon, huang2022hardposevirtualtryon3daware} inject sparse, point-based correspondences into attention or warp features around a few matched points. We discuss the closest of these~\cite{chen2024wearanywaymanipulablevirtualtryon, wan2025incorporatingvisualcorrespondencediffusion} below. Beyond sparse correspondences, some~\cite{zhou2024learningflowfieldsattention} derive dense flows from attention and train them by comparing the warped garment with the person image in pixel space, but in unpaired settings, this photometric assumption~\cite{truong2021warpconsistencyunsupervisedlearning} fails more often, degrading the performance. StableVITON~\cite{kim2023stablevitonlearningsemanticcorrespondence} places a loss on person--garment attention, but it only encourages smooth attention across neighboring queries without specifying where attention should point. CORAL instead supervises attention through geometric correspondence rather than photometric reconstruction, specifying for each person token the garment position its attention should point to.

\textsuperscript{*}\textit{Relationship to Wear-Any-Way~\cite{chen2024wearanywaymanipulablevirtualtryon} and SPM-Diff~\cite{wan2025incorporatingvisualcorrespondencediffusion}}:~These methods use correspondence as an input, needed at inference as well as in training, whereas CORAL uses it \textit{only as a training target} and takes \textit{no correspondence at inference}. Training uses paired data, where the person already wears the reference garment, so matches can be estimated from shared appearance. In actual use, the person wears a different garment, and appearance no longer reliably determines the correct match. The gap is large: our cycle-consistency check (Eq.~\ref{eq:reliability_mask}) discards roughly 85\% of the proposed matches for unpaired inputs on VITON-HD, against about 30\% for paired ones. Another difference lies in verification. A matcher finds the most similar point, but on a garment full of similar regions, this point is not necessarily the geometrically correct match. CORAL supervises attention only with matches \textit{verified} by the check, rather than using them as they are.

\vspace{-5pt}

\begin{figure*}[t!]
  \centering
\includegraphics[width=\linewidth]{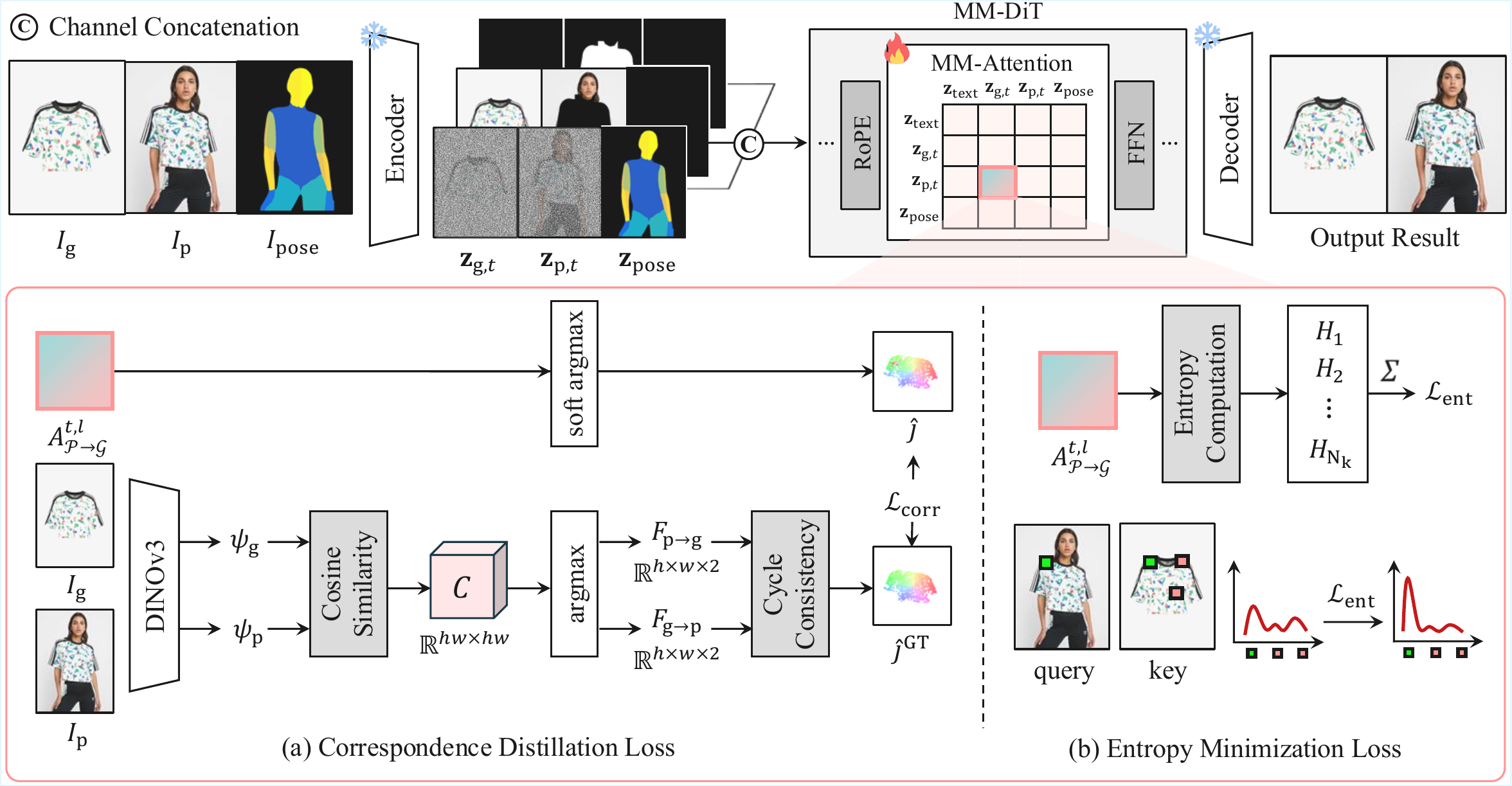}
\caption{\textbf{Overall Architecture.} 
CORAL builds upon a baseline architecture that constructs the noisy latent $\mathbf{z}_t$ by horizontally concatenating the noisy garment latents $\mathbf{z}_{\text{g},t}$ and person latents $\mathbf{z}_{\text{p},t}$, and then channel-wise concatenates the conditioning canvas $\mathbf{z}_{\text{diptych}}$ and mask canvas $\mathbf{m}_{\text{diptych}}$ with $\mathbf{z}_t$ before the input projection layer. Pose is injected by adding $\mathbf{z}_{\text{pose}}$ as tokens, with RoPE set to share spatial positions between person and pose tokens. $\mathcal{L}_\text{CORAL}$ is applied to the person$\rightarrow$garment matching cost $A^{t,l}_{\mathcal{P}\rightarrow\mathcal{G}}$ estimated from MM-Attention within DiT blocks: $\mathcal{L}_{\text{corr}}$ aligns $A^{t,l}_{\mathcal{P}\rightarrow\mathcal{G}}$ to pseudo ground-truth correspondences extracted from DINOv3, while $\mathcal{L}_{\text{ent}}$ is computed on $A^{t,l}_{\mathcal{P}\rightarrow\mathcal{G}}$ to encourage sharper, more localized matches.}
  \vspace{-15pt}
  \label{fig:main_architecture}
\end{figure*}

\section{Preliminaries}
\paragraph{Latent Diffusion Models (LDM).}
LDM~\cite{rombach2022highresolutionimagesynthesislatent} operates in latent space, consisting of VAE and denoiser $\epsilon_\theta(\cdot)$. The VAE encoder maps an image $I \in \mathbb{R}^{H \times W \times 3}$ into a latent representation $\mathbf{z}_0 \in \mathbb{R}^{h \times w \times c}$, where $(H, W, 3)$ and $(h, w, c)$ denote the height, width, and channel dimensions in RGB and latent spaces, respectively. At a timestep $t$, Gaussian noise $\epsilon\sim\mathcal{N}(0, \mathbf{I})$ is added to $\mathbf{z}_0$, producing noisy latent $\mathbf{z}_t $. 
Given conditioning tokens $\mathbf{c}$, we follow rectified flow~\cite{esser2024scalingrectifiedflowtransformers}, 
where the model predicts the velocity transporting a clean latent toward Gaussian noise. 
Specifically, at a timestep $t$, a linearly interpolated latent is constructed as
\begin{equation}
\mathbf{z}_t = (1 - t)\, \mathbf{z}_0 + t\, \boldsymbol{\epsilon}, 
\quad \boldsymbol{\epsilon} \sim \mathcal{N}(0, \mathbf{I}),
\label{eq:latent}
\end{equation}
and the velocity network $\mathbf{v}_\theta(\mathbf{z}_t, t, \mathbf{c})$ 
is trained to approximate the target velocity $(\boldsymbol{\epsilon} - \mathbf{z}_0)$ conditioned on $\mathbf{c}$. 
The velocity training objective $\mathcal{L}_\mathrm{velocity}$ is defined as
\begin{align}
\mathcal{L}_\mathrm{velocity}
= \mathbb{E}_{z_0,\, \boldsymbol{\epsilon},\, t}\Big[
\lVert \mathbf{v}_\theta(\mathbf{z}_t, t, \mathbf{c}) - (\boldsymbol{\epsilon} - \mathbf{z}_0) \rVert_2^2
\Big].
\label{eq:l_diff}
\end{align}

\paragraph{Attention in DiTs.}
Diffusion Transformers (DiTs)~\cite{peebles2023scalablediffusionmodelstransformers, esser2024scalingrectifiedflowtransformers} consist of multiple layers of 3D full attention that jointly process visual and textual information. The noisy latent tokens $\mathbf{z}_t$ and the conditioning tokens $\mathbf{c}_t$ at time step $t$ are concatenated to form the input sequence. The conditioning tokens are constructed by concatenating tokens from multiple conditions, $\mathbf{c}_t = [\,c_{t,1}\|...\| c_{t,N}\,]$, where $N$ denotes the number of conditions and $[\,\cdot \| \cdot \,]$ denotes token-wise concatenation. Therefore, the final input sequence is $ [\,\mathbf{z}_t\| \mathbf{c}_t\,]$. At each timestep $t$ and transformer layer $l$, the concatenated sequence is projected into queries $Q^{t, l}$ and keys $K^{t,l}$:
\begin{equation}
Q^{t,l} = [\,Q^{t,l}_{\mathbf{z}}\, \| \, Q^{t,l}_{\mathbf{c}_1}\, \| \,...\, \| \, Q^{t,l}_{\mathbf{c}_{N}}\,],\quad
K^{t,l} = [\,K^{t,l}_{\mathbf{z}}\, \| \, K^{t,l}_{\mathbf{c}_1}\, \| \,...\, \| \, K^{t,l}_{\mathbf{c}_{N}}\,].
\label{eq:query_key}
\end{equation}
Each token in the sequence is augmented with rotary position embeddings (RoPE)~\cite{su2023roformerenhancedtransformerrotary}. The attention map $A^{t,l}$ at timestep $t$ and layer $l$ is then computed as:
\begin{align}
A^{t,l} = \mathrm{Softmax}\bigg(\frac{Q^{t,l}(K^{t,l})^\top}{\sqrt{d}}\bigg),
\label{eq:full_attention}
\end{align}
where $\mathrm{Softmax}(\cdot)$ is applied over the key dimension for each query, and $d$ denotes a channel dimension. 
To sum up, this full attention determines how each latent or conditioning token attends to every other token in the sequence.
\vspace{-5pt}

\section{Method}
\subsection{Task Definition}
Let $I_\text{p} \in \mathbb{R}^{H \times W \times 3}$ denote a person image, $I_\text{g} \in \mathbb{R}^{H \times W \times 3}$ a garment image, and $I_\text{pose} \in \mathbb{R}^{H \times W \times 3}$ a pose image (estimated from $I_\text{p}$). Let $M_\text{p}, M_\text{g} \in \mathbb{R}^{H \times W}$ be binary masks indicating garment regions in $I_\text{p}$ and $I_\text{g}$, respectively, and $M_\text{e} \in \mathbb{R}^{H \times W \times 3}$ be a binary inpainting mask, specifying the region to be edited in $I_\text{p}$, where $M_\text{p} \subseteq M_\text{e}$. Given the inputs $\{I_\text{p}, I_\text{g}, I_\text{pose}, M_\text{p}, M_\text{g}, M_\text{e}\}$, our goal is to synthesize a try-on image $\hat{I}$ that (i) preserves the appearance of $I_\text{p}$ within the region specified by $M_\text{e}$, (ii) follows the pose defined by $I_\text{pose}$, and (iii) accurately transfers the shape and local details of the garment $I_\text{g}$ according to $M_\text{g}$ into the person image, restricted to $M_\text{e}$.
\vspace{-5pt}

\subsection{Network Architecture}
\label{DIT-based VTON Baseline}
\paragraph{Diptych Formulation.} 
\label{sec:diptcy_formulation}
We first build a DiT-based baseline with a simple conditioning design, leveraging DiT’s in-context conditioning ability for reference-guided generation~\cite{shin2025largescaletexttoimagemodelinpainting, huang2024incontextloradiffusiontransformers}. We adopt a two-panel diptych layout that horizontally concatenates garment and person latents to enable direct token-level interaction through multimodal attention, instead of relying on auxiliary encoders to integrate garment. 
The overall architecture is illustrated in Fig.~\ref{fig:main_architecture}.

Specifically, we encode the garment and person images, $I_\text{g}$ and $I_\text{p}$, with a VAE to obtain $\mathbf{z}_\text{g}$ and $\mathbf{z}_\text{p}$, and form a two-panel latent canvas. At timestep $t$, we construct the noisy diptych latent as:
\begin{equation}
\mathbf{z}_t = [\, \mathbf{z}_{\text{g},t} \, \| \, \mathbf{z}_{\text{p},t} \,],
\end{equation} where $\mathbf{z}_{\text{g},t}$ and $\mathbf{z}_{\text{p},t}$ are noisy garment and person latents at timestep $t$. For conditioning clean garment and masked person image, we use the clean garment  latent and the masked person latent as:
\begin{equation}
\mathbf{z}_{\text{diptych}} = [\, \mathbf{z}_\text{g} \, \| \, \mathbf{z}_\text{p} \odot (1 - \mathbf{m}_\text{e}) \,],
\end{equation}
where $\mathbf{m}_\text{e}$ is obtained by downsampling $M_\text{e}$ into the latent space. We also define a binary mask canvas $\mathbf{m}_{\text{diptych}}$ following the same two-panel layout, assigning ones to the region to be replaced and zeros elsewhere:
\begin{equation}
\mathbf{m}_{\text{diptych}} = [\, \mathbf{0}_{h \times w} \, \| \, \mathbf{m}_\text{e} \,].
\end{equation}

\vspace{-5pt}

\label{sec:pose_injection}
\paragraph{Pose Injection.} Previous try-on methods often suffer from pose hallucination and hand inconsistencies. To mitigate this, a common design in prior methods is to concatenate a pose condition $\mathbf{z}_{\text{pose}}$, obtained by encoding $I_\text{pose}$ with a VAE, with the noisy person latent $\mathbf{z}_{p,t}$ along the channel dimension ~\cite{choi2024improvingdiffusionmodelsauthentic, kim2023stablevitonlearningsemanticcorrespondence}. However, we find this approach suboptimal for strict pose alignment, as the person appearance and pose conditions become entangled in the latent space. 

Following recent DiT-based work~\cite{tan2025ominicontrolminimaluniversalcontrol}, our baseline instead concatenates clean pose condition $\mathbf{z}_{\text{pose}}$ along the token dimension, forming $[\mathbf{z}_{\text{g}, t} || \mathbf{z}_{\text{p}, t} || \mathbf{z}_{\text{pose}}]$ as input. We then modify RoPE~\cite{su2023roformerenhancedtransformerrotary} so that $\mathbf{z}_{\text{p},t}$ and $\mathbf{z}_{\text{pose}}$ share the same positional indices, which provides strong positional correspondence and allows pose to be integrated more coherently through full attention. Finally, $\mathbf{z}_{\text{diptych}}$ and $\mathbf{m}_{\text{diptych}}$ are zero-padded to match the input channel dimension and channel-wise concatenated with $[\mathbf{z}_{t} || \mathbf{z}_{\text{pose}}]$ as the input for the embedding layer, and the resulting tokens are then passed to the DiT. 
\vspace{-4pt}
\subsection{Person-Garment Correspondence Analysis}
\label{correspondence motivation}

\begin{wrapfigure}{r}{0.43\linewidth}
\vspace{-35pt}
  \centering
  \includegraphics[width=\linewidth]{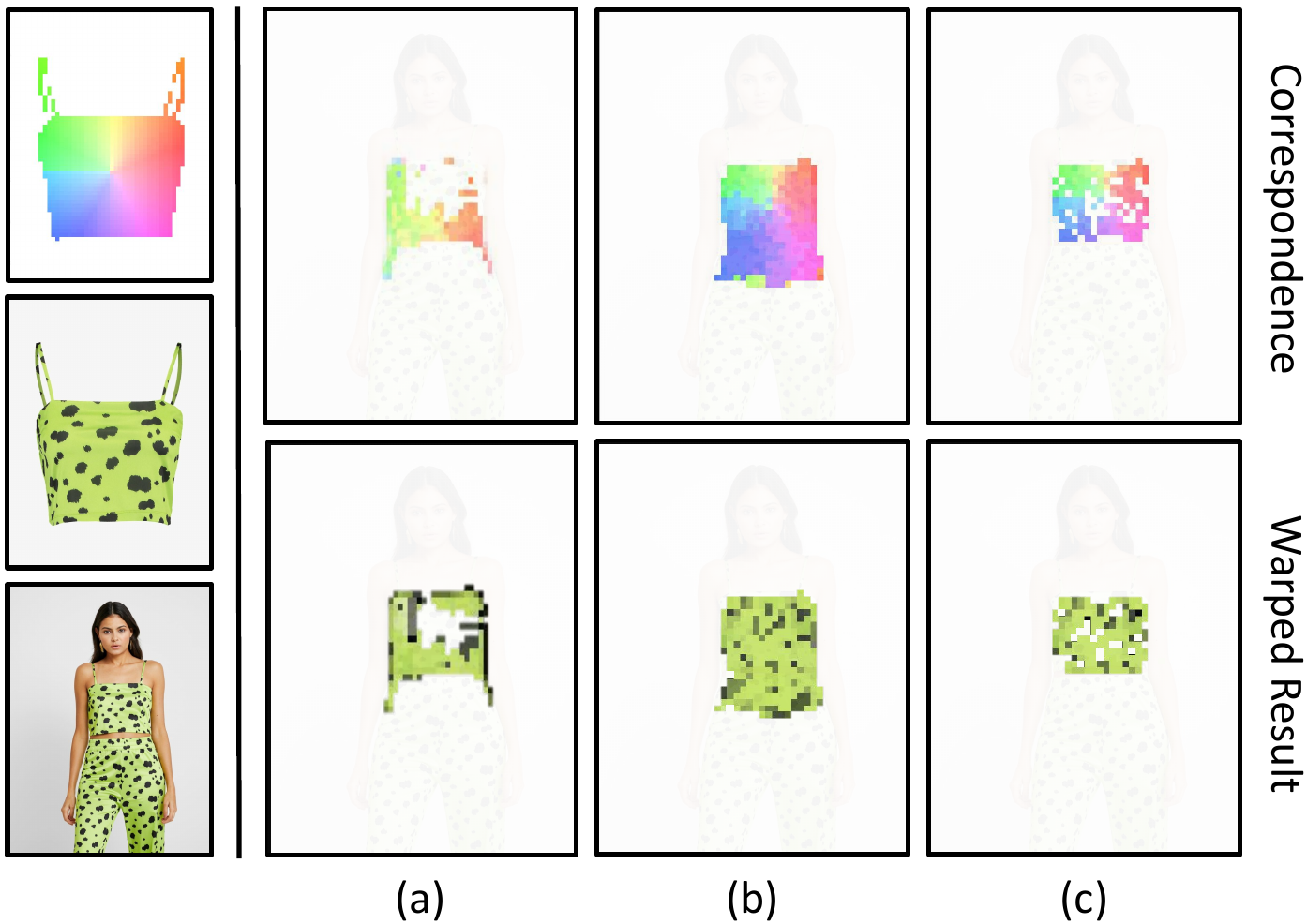}
  \vspace{-15pt}
  \caption{ \textbf{Correspondence Visualization and Warped Results.} 
 We visualize correspondence fields and the resulting warped garment, computed by mapping pixels from the garment reference image to the garment region of the person image. 
(a) Baseline attention-derived correspondence. 
(b) DINOv3 correspondence before the cycle-consistency check. 
(c) Filtered correspondence after the cycle-consistency check. 
The baseline warp shows geometric distortion, while unfiltered DINOv3 can mistakenly match visually similar regions, such as lower-garment areas. 
 }
\vspace{-20pt}
  \label{fig:corr_vis}
\end{wrapfigure}

\paragraph{Motivation} Correspondence has long been a core component of VTON, and prior methods have improved try-on quality by strengthening person–garment matching. In U-Net dual-branch architectures, this is typically achieved by injecting matched features into the generation network~\cite{wan2025incorporatingvisualcorrespondencediffusion, chen2024wearanywaymanipulablevirtualtryon}. However, when directly reproduced on our DiT-based baseline, these feature injection strategies yield only marginal improvements, as shown in Tab.~\ref{tab:spm_diff_comparison}.  

Simply injecting correspondence-aware features provides additional information, but does not explicitly optimize the person-garment matching relation itself. This limitation is particularly relevant in unified DiT architectures, where such features are jointly processed with person, garment, and text tokens within shared self-attention. As a result, the correspondence signal remains as conditioning signal, rather than a training objective that directly guides local person–garment matching. This motivates us to explore correspondence not only as an auxiliary input signal, but also as a direct supervision signal for matching inside DiT.

Another possible direction is to align DiT intermediate features with external matching descriptors (e.g., DINOv3~\cite{siméoni2025dinov3}), following recent representation alignment methods~\cite{yu2025representationalignmentgenerationtraining}. However, this can interfere with
the model's learned generative representations and requires full retraining~\cite{yu2025representationalignmentgenerationtraining}.  Moreover, intermediate features encode appearance information 
from value matrices that weakens spatial relationships, making 
them suboptimal for precise matching~\cite{an2025cross,nam2024dreammatcher}.

To this end, we study query–key attention within the DiT’s full self-attention, which implicitly models interactions between person and garment tokens and can be leveraged for explicit correspondence supervision without disrupting the underlying generative representations.

\paragraph{Correspondence Estimation.} To examine the baseline's correspondence behavior, we extract query–key correspondences from the DiT's full 3D attention. Following Eq.~\ref{eq:query_key}, the query and key at timestep 
$t$ and layer $l$ consist of text, garment, person, and pose 
token components:
\begin{equation}
Q^{t,l} = [\,Q^{t,l}_{\mathbf{z}_{\text{text}}} \| Q^{t,l}_{\mathbf{z}_{\text{g}}} \| Q^{t,l}_{\mathbf{z}_{\text{p}}} \| Q^{t,l}_{\mathbf{z}_{\text{pose}}}\,],\quad
K^{t,l} = [\,K^{t,l}_{\mathbf{z}_{\text{text}}} \| K^{t,l}_{\mathbf{z}_{\text{g}}} \| K^{t,l}_{\mathbf{z}_{\text{p}}} \| K^{t,l}_{\mathbf{z}_{\text{pose}}}\,].
\end{equation} 
Let $m_\text{p}, m_\text{g}$ be the downsampled masks of $M_\text{p}, M_\text{g}$, and $\mathcal{P}, \mathcal{G}$ the corresponding person and garment token index sets within $Q^{t,l}_{\mathbf{z}_{\text{p}}}$ and $K^{t,l}_{\mathbf{z}_\text{g}}$. Using Eq.~\ref{eq:full_attention}, we compute the person$\rightarrow$ garment attention with $Q^{t,l}_{\mathbf{z}_{\text{p}}}$ as queries and $K^{t,l}_{\mathbf{z}_\text{g}}$ as keys:
\begin{equation}
A^{t,l}_{\mathcal{P}\rightarrow\mathcal{G}} := A^{t,l}[\mathcal{P},\mathcal{G}]\in\mathbb{R}^{|\mathcal{P}|\times |\mathcal{G}|}.
\label{eq:sub-attention}
\end{equation}
Following~\cite{nam2024dreammatcher,nam2025emergent}, we 
treat $A^{t,l}_{\mathcal{P}\rightarrow\mathcal{G}}$ as the 
matching cost between person and garment latents. For simplicity, the variables $t$ and $l$ are omitted from the following equations. From these matching costs, we extract dense correspondences from person spatial location $i$ to garment latents by taking $\mathop{\mathrm{arg\,max}}$ over garment tokens: 
\begin{equation}
\hat{j}_i = \mathop{\mathrm{arg\,max}}_{j \in \mathcal{G}} 
{A}_{\mathcal{P} \to \mathcal{G}}(i, j).
\label{eq:predicted_cost_map}
\end{equation}

\paragraph{Pseudo Ground-Truth Construction.}
To enable correspondence supervision and evaluation, we require ground-truth person-garment correspondences. Since such annotations do not exist, we construct pseudo ground-truth matches using DINOv3~\cite{siméoni2025dinov3}, a vision foundation model, shown to be highly effective for dense matching. CORAL only requires reliable local matches and can work with other dense correspondence estimators. Given  $I_{\text{p}}$ and  $I_{\text{g}}$, we extract feature descriptors $\phi(\cdot)$ from DINOv3 and keep only tokens inside the downsampled masks:
\begin{equation}
\psi_{\text{p}} = \phi(I_{\text{p}})\odot m_{\text{p}},
\quad
\psi_{\text{g}} = \phi(I_{\text{g}})\odot m_{\text{g}}.
\label{eq:descriptors}
\end{equation}
Following standard dense matching protocols ~\cite{hong2022cost, hong2022neural, cho2024local, nam2024diffusionmodeldensematching}, we compute the matching cost between descriptors using cosine similarity:
\begin{equation}
C(i,j) = \frac{\psi_{\text{p}}(i) \cdot \psi_{\text{g}}(j)}
{\lVert \psi_{\text{p}}(i) \rVert_2 \ \lVert \psi_{\text{g}}(j) \rVert_2},
\label{eq:costmap}
\end{equation}
where $i$ and $j$ index the spatial locations of the person and garment descriptors, respectively.

To obtain reliable pseudo ground-truth matches, we introduce a reliability mask $\mathbf{m}_{\text{rel}}(i)$ that filters out unstable correspondences using a cycle-consistency ~\cite{jiang2021cotrcorrespondencetransformermatching} constraint. We first estimate bidirectional flow, which are person to garment, $F_{\text{p}\to\text{g}} \in \mathbb{R}^{h\times w\times 2}$, and garment to person, $F_{\text{g}\to\text{p}} \in \mathbb{R}^{h\times w\times 2}$, by taking the $\mathop{\mathrm{\arg\max}}$ over spatial locations of garment and person in the matching cost, respectively. The reliability mask identifies points whose matched locations in the garment features return to the original locations in the person features within a specified threshold $\gamma$:
\begin{equation}
\mathbf{m}_{\text{rel}}(i) =
\begin{cases}
1, & \text{if } \lVert F_{\text{g}\to\text{p}}(F_{\text{p}\to\text{g}}(i)) - i \rVert_2 < \gamma,\\
0, & \text{otherwise.}
\end{cases}
\label{eq:reliability_mask}
\end{equation}
Finally, we compute the mean L2 loss between estimated correspondences and the
pseudo ground-truth matches over the reliable query tokens:
\begin{equation}
\mathcal{L}_{\text{corr}} \;=\; \frac{1}{N_{\text{rel}}}\sum_{i=1}^{N} m_{\text{rel}}(i)\,
\big\|\hat{j}_i - \hat{j}_i^{\text{GT}}\big\|_2^2,
\qquad N_{\text{rel}} = \sum_{i=1}^{N} m_{\text{rel}}(i),
\label{eq:corr_loss}
\end{equation}
where $N$ is the number of query tokens.

\paragraph{Effect of Correspondence Quality.} We leverage constructed pseudo-matches at inference time to investigate how correspondence quality affects VTON performance. Using a pretrained DiT-based VTON model as the baseline, we modify its query-key attention to align with the pseudo-ground truth, sharpening the matching cost to concentrate around the target correspondence locations. As shown in Fig.~\ref{fig:motivation}(b), modifying the attention with target correspondences yields visibly improved try-on quality compared to the baseline, indicating that the attention pattern substantially influences generation quality, and directly correcting it without modifying the underlying features clearly improves results.

However, this inference-time intervention requires pseudo ground-truth correspondences for each input pair and remains applicable only to paired settings. To enable the model to internalize this correspondence prior and generalize to unpaired settings, we propose to directly supervise the query-key matching during training. This motivates CORAL, which we describe in the next section.

\subsection{CORAL: Correspondence Alignment} 
Based on our observation, we introduce CORAL, which integrates reliable pseudo ground-truth matches into the baseline’s query–key matching for robust virtual try-on performance. 

\paragraph{Correspondence Distillation Loss.}
We aim to align the estimated matches obtained from query–key attention with the pseudo ground-truth matches provided by DINOv3~\cite{siméoni2025dinov3}. Because the $\mathrm{argmax}$ operation is not differentiable (Eq.~\ref{eq:predicted_cost_map}), we use a soft $\mathrm{argmax}$ to predict the estimated correspondences from the query–key attention, formulated as follows:
\begin{equation}
\hat{j}_i = \sum_{j=1}^{N} {A}_{\mathcal{P} \to \mathcal{G}}(i, j) \cdot j.
\label{eq:softargmax}
\end{equation}

Finally, we compute the mean L2 loss between estimated correspondences and the pseudo ground-truth matches:
\begin{equation}
\mathcal{L}_{\text{corr}} = \frac{1}{N} 
\sum_{i=1}^{N} 
\lVert \hat{j}_{i} - \hat{j}_i^{\text{GT}} \rVert_2^2 .
\label{eq:corr_loss}
\end{equation}

\begin{figure*}[t]
  \centering
\includegraphics[width=\linewidth]
  {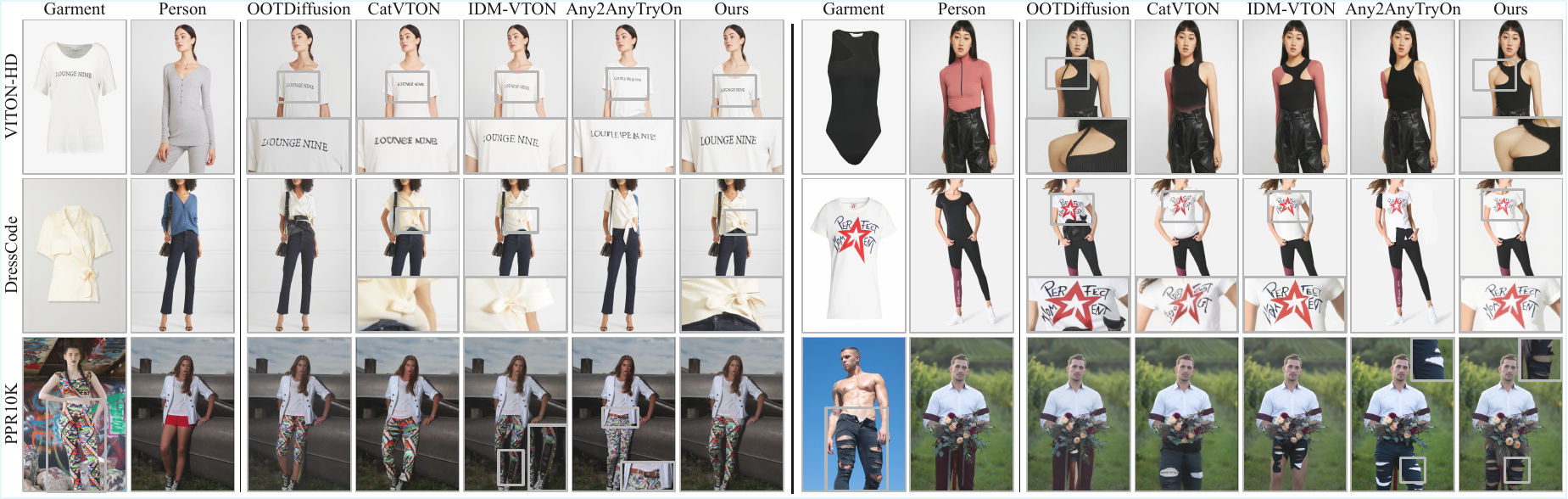}
  \vspace{-15pt}
  \caption{\textbf{Qualitative Comparison.} We show qualitative results on the standard benchmarks VITON-HD~\citep{choi2021vitonhdhighresolutionvirtualtryon} and DressCode~\citep{morelli2022dresscodehighresolutionmulticategory}, as well as in-the-wild evaluation dataset built from PPR10K~\citep{liang2021ppr10klargescaleportraitphoto} (Best viewed when zoomed-in). Additional qualitative results are provided in Appendix~\ref{supple:additional_results}.}
  \vspace{-5pt}
  \label{fig:main_qual}
\end{figure*}

\newcommand{\mc}[1]{\multicolumn{1}{c}{#1}}
\newcommand{\up}{$\uparrow$}
\newcommand{\down}{$\downarrow$}
\begin{table}[!t]
\centering

\begin{minipage}[t]{0.48\linewidth}
\centering

\begin{minipage}{\linewidth}
\centering
\scriptsize
\setlength{\tabcolsep}{3pt}
\resizebox{\linewidth}{!}{
\begin{tabular}{l*{6}{c}}
\toprule
 \multicolumn{1}{c}{\multirow{2}{*}{\textbf{Methods}}}
& \multicolumn{4}{c}{Paired} & \multicolumn{2}{c}{Unpaired} \\
\cmidrule(lr){2-5} \cmidrule(lr){6-7} 
& \mc{SSIM \up} & \mc{LPIPS \down} & \mc{FID \down} & \mc{KID \down} & \mc{FID \down} & \mc{KID \down} \\
\midrule
GPVTON~\cite{xie2023gpvtongeneralpurposevirtual} & 0.878 & 0.067 &  8.938   &  4.257   & 11.993 & 4.570   \\
StableVITON \cite{kim2023stablevitonlearningsemanticcorrespondence} & 0.888 & 0.073 &  8.233 &   0.490 & \underline{9.026} & 3.029   \\
OOTDiffusion \cite{xu2024ootdiffusionoutfittingfusionbased} & 0.842 &  0.087 &  6.619 &  0.845 & 9.938 & 1.302  \\
IDM-VTON \cite{choi2024improvingdiffusionmodelsauthentic}  & 0.866 & 0.062 & 6.009 & 0.838 & 9.198 & 1.203 \\
\text{CatVTON}\cite{chong2025catvtonconcatenationneedvirtual} & 0.874 & 0.058 & \underline{5.458} & \underline{0.439} & 9.076 & \underline{1.184}  \\
Any2AnyTryOn \cite{guo2025any2anytryonleveragingadaptiveposition} & 0.838 & 0.087 & 5.482 & \textbf{0.384} & 9.623 &  1.601 \\
FitDiT~\cite{jiang2024fitditadvancingauthenticgarment} & 0.834 & 0.105 & 8.502 & 1.563 & 10.310 & 1.270 \\
\midrule
\textbf{CORAL (w/o $\mathcal{L}_{\text{CORAL}}$)}  & \underline{0.889}  & \underline{0.055} & 5.543 & 0.870  & 9.641  & 1.323 \\
\textbf{CORAL (w $\mathcal{L}_{\text{CORAL}}$)}
& \textbf{0.907} 
& \textbf{0.048} 
& \textbf{4.962} 
& 0.565 
& \textbf{8.763} 
& \textbf{0.880} \\
\bottomrule
\end{tabular}}
\vspace{3pt}
\captionof{table}{\textbf{Quantitative Comparison on VITON-HD~\cite{choi2021vitonhdhighresolutionvirtualtryon}}.}
\label{tab:main_quan_vt}
\end{minipage}

\vspace{6pt}

\begin{minipage}{0.92\linewidth}
\centering
\small
\setlength{\tabcolsep}{5pt}
\vspace{-10pt}
\resizebox{\linewidth}{!}{
\begin{tabular}{l*{6}{c}}
\toprule
\multicolumn{1}{c}{\multirow{2}{*}{\textbf{Methods}}}
& \multicolumn{4}{c}{Paired} & \multicolumn{2}{c}{Unpaired} \\
\cmidrule(lr){2-5} \cmidrule(lr){6-7} 
& \mc{SSIM \up} & \mc{LPIPS \down} & \mc{FID \down} & \mc{KID \down} & \mc{FID \down} & \mc{KID \down}\\
\midrule
GPVTON~\cite{xie2023gpvtongeneralpurposevirtual} & \underline{0.918} & 0.068 & 8.423  & 2.439  & 9.144 & 3.936   \\
OOTDiffusion~\cite{xu2024ootdiffusionoutfittingfusionbased} & 0.886 & 0.069 & 5.082 & 1.377 & 9.276  &  4.009 \\
IDM-VTON~\cite{choi2024improvingdiffusionmodelsauthentic} & 0.904 & 0.052 & 3.472 & 0.882 & 5.343  & 1.321 \\
CatVTON~\cite{chong2025catvtonconcatenationneedvirtual}  & 0.875 & 0.075 & 5.384 & 1.903 & 7.998 & 3.242 \\
Any2AnyTryOn$^{*}$~\cite{guo2025any2anytryonleveragingadaptiveposition}
 & - & - & - & - & 5.573 & 1.458 \\
\midrule
\textbf{CORAL (w/o $\mathcal{L}_{\text{CORAL}}$)}  &  0.908 & \underline{0.045} & \underline{2.896} & \underline{0.418} & \underline{5.221} & \underline{1.315}  \\
\textbf{CORAL (w $\mathcal{L}_{\text{CORAL}})$}
& \textbf{0.927}
& \textbf{0.029} 
& \textbf{2.333}
& \textbf{0.401}
& \textbf{4.692}
& \textbf{0.846} \\
\bottomrule
\end{tabular}}
\vspace{3pt}
\captionof{table}{\textbf{Quantitative Comparison on DressCode~\cite{morelli2022dresscodehighresolutionmulticategory}.}$^*$ does not support paired setting on specific dataset.
}
\vspace{-20pt}
\label{tab:main_quan_dc}
\end{minipage}

\end{minipage}%
\hfill
\begin{minipage}[t]{0.48\linewidth}
\centering

\begin{minipage}{\linewidth}
\centering
\small
\setlength{\tabcolsep}{5pt}
\vspace{-10pt}
\resizebox{\linewidth}{!}{
\begin{tabular}{l*{6}{c}}
\toprule
 \multicolumn{1}{c}{\multirow{2}{*}{\textbf{Methods}}}
& \multicolumn{4}{c}{Paired} & \multicolumn{2}{c}{Unpaired} \\
\cmidrule(lr){2-5} \cmidrule(lr){6-7} 
& \mc{SSIM \up} & \mc{LPIPS \down} & \mc{FID \down} & \mc{KID \down} & \mc{FID \down} & \mc{KID \down} \\
\midrule
OOTDiffusion~\citep{xu2024ootdiffusionoutfittingfusionbased} & 0.804 & 0.130 & 99.492 & 15.592    & 87.818 & 15.467    \\
IDM-VTON~\citep{choi2024improvingdiffusionmodelsauthentic}  &   0.844 &  0.111 & 64.250 & 1.911  & 63.638 & 3.600  \\
\text{CatVTON}\cite{chong2025catvtonconcatenationneedvirtual}  & 0.743 & 0.147  & 81.722  & 10.979   & 76.417  & 11.207\\
Any2AnyTryOn$^*$ \cite{guo2025any2anytryonleveragingadaptiveposition} & -  & - &  -  & - & \textbf{49.728} & \textbf{0.421} \\
\midrule
\textbf{CORAL (w/o $\mathcal{L}_{\text{CORAL}}$)}  & \underline{0.877}  & \underline{0.078} & \underline{44.117} & \underline{0.015} & 56.644 & 1.202 \\
\textbf{CORAL (w $\mathcal{L}_{\text{CORAL}})$}
& \textbf{0.915}
& \textbf{0.060}
& \textbf{43.648}
&  \textbf{0.011}
&   \underline{53.164}
&  \underline{1.101}  \\
\bottomrule
\end{tabular}}
\vspace{3pt}
\captionof{table}{\textbf{Quantitative Comparison on PPR10K~\cite{liang2021ppr10klargescaleportraitphoto}.} $^*$ does not support paired setting on specific dataset.}
\label{tab:main_quan_additional_dataset}
\end{minipage}

\begin{minipage}{\linewidth}
\centering
\small
\resizebox{\linewidth}{!}{
    \begin{tabular}{lccccccccc}
    \toprule
    \multirow{2}{*}{Train/Test}
    & \multicolumn{3}{c}{VITON-HD/VITON-HD} & \multicolumn{3}{c}{DressCode/DressCode} &  \multicolumn{3}{c}{DressCode/PPR10K} \\
    \cmidrule(lr){2-4} \cmidrule(lr){5-7} \cmidrule(lr){8-10} 
    & \mc{GTC \up} & \mc{TAC \up} & \mc{FPC \up} & \mc{GTC \up} & \mc{TAC \up} & \mc{FPC \up} & \mc{GTC \up} & \mc{TAC \up} & \mc{FPC \up} \\
    \midrule
    GPVTON$^*$~\cite{xie2023gpvtongeneralpurposevirtual} & 3.93 & 4.28 & 4.17 & 3.19 & \underline{3.31} & 3.14 & - & - & -\\
    OOTDiffusion \cite{xu2024ootdiffusionoutfittingfusionbased}  & 3.97 & 4.34 & 4.12 & \underline{3.25} & 3.24 & 3.45 & 1.04 & 2.06 & 1.57\\
    IDM-VTON~\cite{choi2024improvingdiffusionmodelsauthentic} & \underline{3.96} & \underline{4.40} & \underline{4.25} & \underline{3.25} & 3.28 & \underline{3.61} & 1.72 & 2.21 & 2.37 \\
    CatVTON \cite{chong2025catvtonconcatenationneedvirtual} & 3.82 & 4.20 & 4.06 & 3.07 & 3.22 & 3.53 &1.05 & 2.00 & 2.01 \\
    Any2AnyTryOn \cite{guo2025any2anytryonleveragingadaptiveposition} & 3.81 & 4.27 & 4.19 & 2.99 & 3.28 & 3.36 & \underline{1.92} & \underline{2.45} & \underline{2.48} \\
    \midrule
    \textbf{CORAL (Ours)} & \textbf{3.99} & \textbf{4.40} & \textbf{4.26} & \textbf{3.47} & \textbf{3.31} & \textbf{3.83} & \textbf{2.07} & \textbf{2.56} & \textbf{2.89} \\
    \bottomrule
    \end{tabular}}
\label{tab:vlm_quan_all}
\vspace{3pt}
\caption{VLM-based evaluation on unpaired settings. Scores are averages of
per-sample GPT-4o ratings on a continuous 0--5 scale (Appendix~\ref{supple:vlm_eval}).
$^\ast$ does not support inference on specific dataset.}
\label{tab:vlm_quan_all}
\end{minipage}

\end{minipage}

\end{table}

\paragraph{Entropy Minimization Loss.}
While correspondence distillation loss ($\mathcal{L}_{\text{corr}}$) aligns query-key matches with external pseudo-matches, this alignment is unreliable when the attention distribution becomes overly diffuse, as incorrect key positions can then dominate the weighted average. To ensure robust correspondence supervision, we introduce an entropy minimization loss that promotes confident, spatially localized query-key alignments. To implement this, we define the entropy based on the attention distribution. Following Shannon’s definition of entropy~\cite{attanasio2022entropybasedattentionregularizationfrees}, we compute the entropy $H_i$ for a given query $i$ by treating its attention weights ${A}(i, \cdot)$ over the keys as a probability distribution:
\begin{equation}
H_{i} \;=\; - \sum_{j=1}^{N_\text{k}} {A}(i, j) \,\log {A}(i,j) ,
\label{eq:entropy_single}
\end{equation}

where $N_{\text{k}}$ is the total number of key tokens. Averaging $H_i$ over the
reliable query tokens yields the entropy minimization loss:
\begin{equation}
\mathcal{L}_{\text{ent}} \;=\; \frac{1}{N_{\text{rel}}}\sum_{i=1}^{N} m_{\text{rel}}(i)\, H_{i}.
\label{eq:entropy_loss}
\end{equation}
Both losses are thus computed only over query tokens that pass the
cycle-consistency check (Eq.~\ref{eq:reliability_mask}). Tokens without a reliable match
are left to the diffusion objective, as in the baseline, since there is no
correct position to teach there.

Lower entropy corresponds to sharper and more confident attention, whereas higher entropy indicates diffuse or uncertain matching. Entropy minimization loss ($\mathcal{L}_{\text{ent}}$) therefore complements the correspondence distillation loss ($\mathcal{L}_{\text{corr}}$) by enforcing sharper, and more reliable alignments.

\paragraph{Total Loss.}
We define $\mathcal{L}_{\text{CORAL}}=
\lambda_{\text{corr}}\mathcal{L}_{\text{corr}}+
\lambda_{\text{ent}}\mathcal{L}_{\text{ent}}$ and optimize:
\begin{equation}
\mathcal{L}_{\text{total}}
=
\mathcal{L}_{\text{velocity}}
+
\mathcal{L}_{\text{CORAL}}.
\label{eq:train_loss}
\end{equation}
\vspace{-20pt}

\section{Experiments}
\begin{figure*}[!t]
  \centering
  \includegraphics[width=0.99\linewidth]{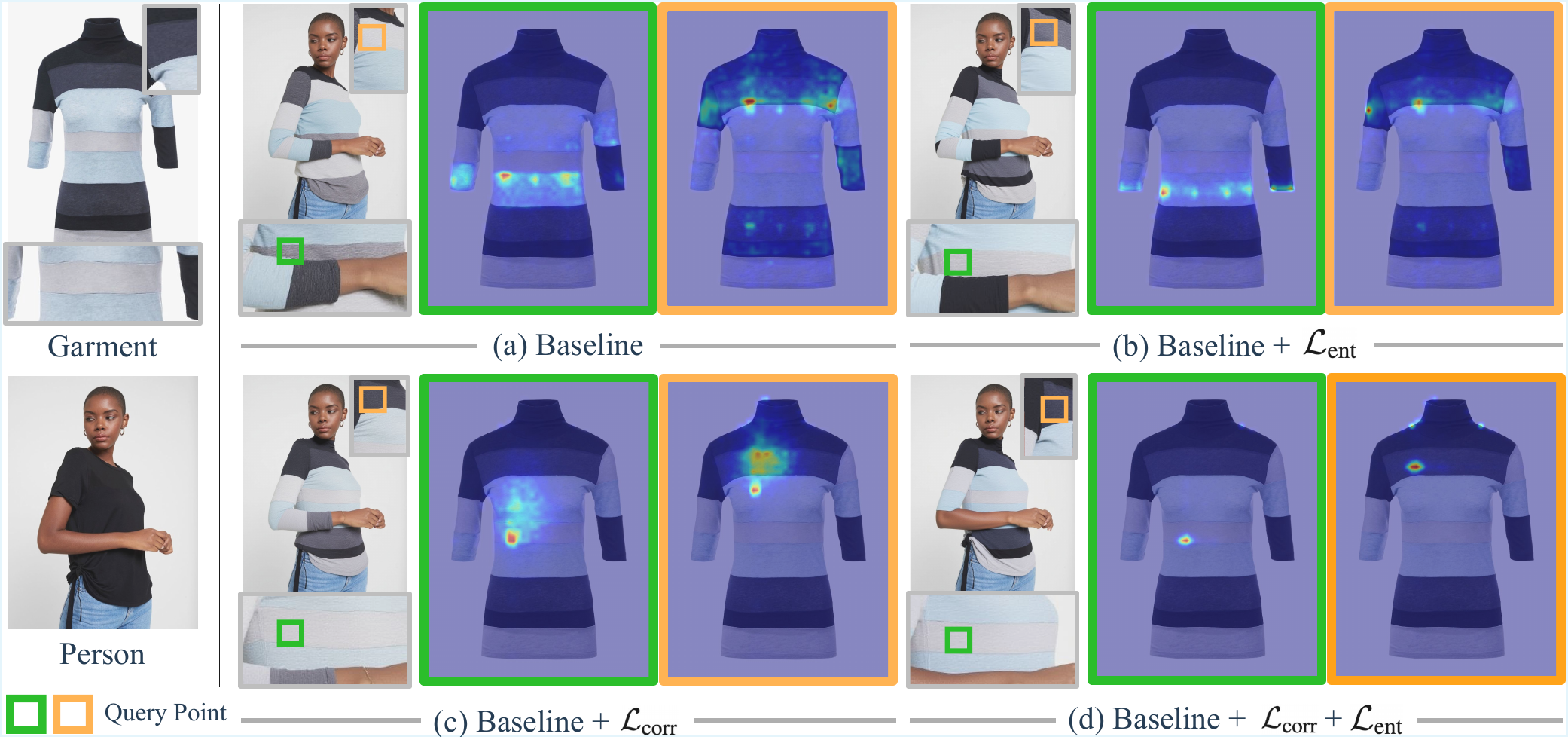}
  \caption{\textbf{Loss Component Ablations.} We show the effectiveness of $\mathcal{L}_\text{ent}$ and $\mathcal{L}_\text{corr}$. \textcolor{myorange}{Orange} and \textcolor{mygreen}{green} markers denote query points, and attention maps outlined in the same colors indicate the matches for each variant. By combining $\mathcal{L}_\text{ent}$ with $\mathcal{L}_\text{corr}$, our model localizes the correct keys most accurately and exhibits the sharpest attention, yielding the best VTON performance.}
    \vspace{-15pt}
  \label{fig:attn_map_vis}
\end{figure*}

\subsection{Evaluation Setting}
\paragraph{Evaluation Dataset.} We train separate models on the standard VTON datasets, \textbf{VITON-HD}~\cite{han2018vitonimagebasedvirtualtryon} and \textbf{DressCode}~\cite{morelli2022dresscodehighresolutionmulticategory}, both at a resolution of $1024 \times 768$, and evaluate them on their respective test sets. For more implementation details, see Appendix~\ref{sec:implementation_details}. Additionally, we evaluate zero-shot generalization on the \textbf{PPR10K}~\cite{liang2021ppr10klargescaleportraitphoto}, which provides less curated inputs than the in-shop product images. For this evaluation, we use a pre-trained checkpoint trained on DressCode. We also include the worn-garment setting, where the reference garment is shown on another person, which is offered in Appendix~\ref{supple:applications}. Additional details on PPR10K evaluation are provided in Appendix~\ref{supple:additional_evaluation}.

\textbf{Metrics.}  
We report conventional metrics SSIM~\cite{cong2022imagequalityassessmentgradient}, LPIPS ~\cite{zhang2018unreasonableeffectivenessdeepfeatures}, FID~\cite{heusel2018ganstrainedtimescaleupdate} and KID~\cite{bińkowski2021demystifyingmmdgans} for paired settings, and measure FID and KID for unpaired settings. We additionally evaluate correspondence accuracy using Percentage of Correct Keypoints (PCK) on the VITON-HD test set. PCK measures whether predicted correspondences fall within a distance threshold $\alpha$ of the pseudo–matches.

\paragraph{VLM-based Evaluation.} As prior studies note~\cite{xiaobin2025vtbenchcomprehensivebenchmarksuite}, FID and KID inadequately reflect garment transfer fidelity in unpaired settings. We therefore evaluate three aspects: Garment Transfer Consistency (GTC), Textual Attribute Consistency (TAC), and Fit Pose Coherence (FPC), capturing garment fidelity, attribute alignment, and natural fit, respectively. For more details, see Appendix~\ref{supple:vlm_eval} and~\ref{supple:human_eval}.
\label{sec:exp-vlm-eval}
\vspace{-4pt}
\subsection{Quantitative Comparison}
\paragraph{Main Results.} 
Tab.~\ref{tab:main_quan_vt} and Tab.~\ref{tab:main_quan_dc} report quantitative comparisons on VITON-HD~\cite{choi2021vitonhdhighresolutionvirtualtryon} and DressCode~\cite{morelli2022dresscodehighresolutionmulticategory}. CORAL achieves the best overall performance on both benchmarks. Tab.~\ref{tab:main_quan_additional_dataset} further shows that CORAL achieves the best results in the paired setting and remains competitive in the unpaired setting on the in-the-wild dataset, outperforming most baselines. Across all benchmarks, CORAL consistently improves every metric, demonstrating robust gains over strong baselines. 
\paragraph{VLM Evaluation Results.}
Tab.~\ref{tab:vlm_quan_all} reports VLM-based evaluation results, where CORAL achieves the best performance across all metrics. By aligning query-key matches to correct person-garment regions, CORAL transfers appearance more accurately, improving GTC. More precise correspondences better preserve garment attributes, yielding higher TAC, and result in more natural wear and pose adherence, reflected in improved FPC. 
\vspace{-5pt}

\subsection{Qualitative Comparison} 
Fig. 5 shows qualitative comparisons between prior methods and CORAL. CORAL preserves global garment structure while recovering fine details missed by baselines, and keeps garment text clear where others produce blur or distortion. It also maintains closer pose alignment to the input, avoiding subtle deviations. For more results, see Appendix~\ref{supple:additional_results}.

\subsection{Ablation Study}
\begin{wraptable}{r}{0.45\columnwidth}
\vspace{-10pt}
\centering
\resizebox{\linewidth}{!}{
\begin{tabular}{c| c c c c c c c c}
\toprule
&
\multirow{3}{*}{$\mathcal{L}_{\text{corr}}$} & \multirow{3}{*}{$\mathcal{L}_{\text{ent}}$} &
\multicolumn{4}{c}{Paired} & \multicolumn{2}{c}{Unpaired} \\
\cmidrule(lr){4-7} \cmidrule(lr){8-9}
& & & \mc{SSIM \up} & \mc{LPIPS \down} & \mc{FID \down} & \mc{KID \down} & \mc{FID \down} & \mc{KID \down} \\
\midrule
(I)  & \textcolor{gray}{\ding{55}} & \textcolor{gray}{\ding{55}} & 0.889 & 0.055 & 5.543 & 0.870 & 9.641 & 1.323 \\
(II) & \textcolor{gray}{\ding{55}} & \ding{51} & 0.905 & 0.052 & 5.482 & 0.986 & 9.012 & 1.011  \\
(III)  & \ding{51} & \textcolor{gray}{\ding{55}} & 0.906 & 0.051 & 5.451 & 1.028 & 8.979  & 1.250  \\
(IV)  & \ding{51} & \ding{51} & 0.907 & 0.048 & 4.962 & 0.565 & 8.763 & 0.880 \\
\bottomrule
\end{tabular}
}
\vspace{-5pt}
\caption{\textbf{ Loss Components Ablation on VITON-HD.}}
\vspace{-16pt}
\label{tab:loss_abl}
\end{wraptable}

\paragraph{Ablation of Loss Components.} Fig.~\ref{fig:attn_map_vis} and Tab.~\ref{tab:loss_abl} highlight the effects of our loss design. (a) and (I) show the baseline, where attention is poorly localized and often dispersed away from query-relevant regions, yielding misaligned matches (e.g., misplaced patterns)  and artifacts. (b) and (II) present the effect of $\mathcal{L}_\text{ent}$, which sharpens attention and improves fine detail transfer. However, since $\mathcal{L}_\text{ent}$ does not correct \emph{where} the model matches, coarse properties like garment length often remain inaccurate, and sharpening on wrong matches can transfer inappropriate appearance. This explains why (II) improves SSIM and LPIPS over (I), while still exhibiting misalignment in challenging cases. (c) and (III) show the effect of $\mathcal{L}_\text{corr}$, which relocates correspondences to the correct garment regions and improves coarse properties (e.g., silhouette, pattern geometry). However, attention remains relatively diffuse, limiting fine detail transfer with low attention mass on the matched key. Compared to (b), (c) achieves more accurate geometry but lags in color fidelity. (d) and (IV), the combination of $\mathcal{L}_\text{ent}$ and $\mathcal{L}_\text{corr}$, integrate these complementary gains. This yields both accurate alignment and sharp detail transfer, leading to the best qualitative and quantitative results.

\begin{wraptable}{r}{0.47\columnwidth}
\vspace{-10pt}
\centering
\setlength{\tabcolsep}{6pt}
\resizebox{\linewidth}{!}{
\begin{tabular}{lccc}
\toprule
Method & SSIM \up & LPIPS \down & FID \down \\
\midrule
Baseline                                             & 0.889 & 0.055 & 5.543 \\
\arrayrulecolor{gray!50}\midrule\arrayrulecolor{black}
Baseline + DINOv3 conditioning (sparse, $\sim$25pts) & 0.889 & 0.054 & 5.889 \\
Baseline + DINOv3 conditioning (dense)               & 0.883 & 0.074 & 5.598 \\
Baseline + ATV loss                                  & 0.892 & 0.054 & 5.652 \\
\midrule
\textbf{CORAL (direct supervision)}                  & \textbf{0.908} & \textbf{0.048} & \textbf{5.009} \\
\bottomrule
\end{tabular}
}
\caption{\textbf{Prior Approaches on VITON-HD}}
\label{tab:spm_diff_comparison}
\vspace{-10pt}
\end{wraptable}

\label{sec:prior_comparison}
\paragraph{Comparison with Prior Approaches.} We compare CORAL with two kinds of prior approaches, reproduced on our DiT baseline, to highlight the importance of direct supervision in DiT-based VTON. The first is feature injection. We reproduce SPM-Diff~\cite{wan2025incorporatingvisualcorrespondencediffusion} by injecting garment features at matched positions using per-layer zero-initialized linear projections, following the original design (see Appendix~\ref{supple:supervision_mechanism} for implementation details), and evaluate both the original sparse setting (25 matched points) and a dense correspondence variant. The second is attention regularization without a correspondence target. We apply the attention total variation (ATV) loss of StableVITON~\cite{kim2023stablevitonlearningsemanticcorrespondence}, which acts on the same person-to-garment attention as ours but only smooths its attention center across neighbouring queries. As shown in Tab.~\ref{tab:spm_diff_comparison}, both feature injection variants yield negligible or negative gains over the baseline. ATV improves reconstruction slightly but worsens FID. It supervises neither of the two things our losses do: it names no position for attention to point at, and it never directly asks attention to sharpen onto one. CORAL improves all metrics. This confirms that auxiliary conditioning is insufficient for DiT-based VTON, and that direct supervision on attention-level correspondences is required, on both where attention points and how sharply.

\begin{wrapfigure}{r}{0.45\linewidth}
\vspace{-20pt}
  \centering
  \includegraphics[width=\linewidth]{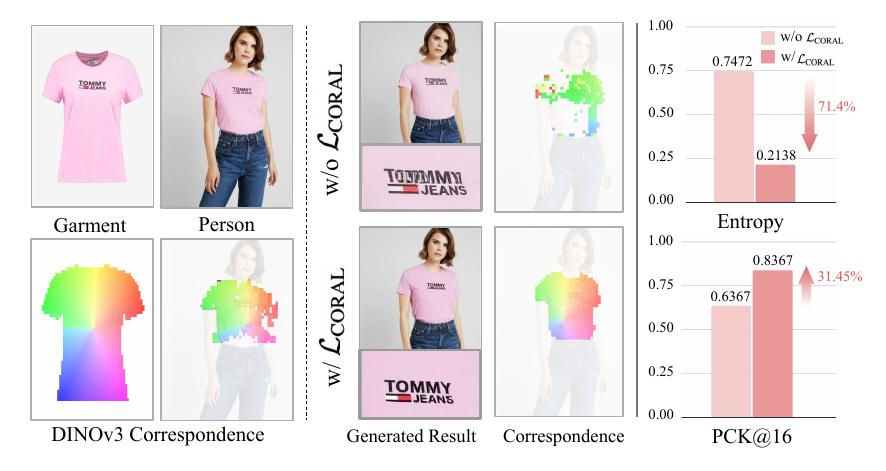}
  \vspace{-20pt}
  \caption{\textbf{Effect of $\mathcal{L}_{\text{CORAL}}$.} We visualize person$\rightarrow$garment correspondences from attention and the resulting try-on images before and after applying $\mathcal{L}_{\text{CORAL}}$. The rightmost plot shows correspondence quality, averaged across timesteps and layers.}
  \vspace{-10pt}
  \label{fig:generalization_across_baselines}
\end{wrapfigure}

\paragraph{Impact of CORAL on Attention Correspondence.} Fig.~\ref{fig:generalization_across_baselines} summarizes how $\mathcal{L}_{\text{CORAL}}$ affects $A^{t,l}[\mathcal{P},\mathcal{G}]$ and generated results. Without $\mathcal{L}_{\text{CORAL}}$, the correspondence maps exhibit mismatches near local textures and garment boundaries, although the vertical alignment appears plausible.  These mismatches lead to unclear logo patterns in the generated result. Adding $\mathcal{L}_{\text{CORAL}}$ consistently corrects the correspondence maps, producing matches that more closely follow the DINOv3 pseudo ground-truth matches, leading to improved text details. Fig.~\ref{fig:generalization_across_baselines} also shows this quantitatively: $\mathcal{L}_{\text{CORAL}}$ improves PCK ($\alpha=16$) by $34\,\%$ while reducing attention entropy, indicating more accurate and sharper correspondences. 
\vspace{-5pt}

\section{Conclusion}
\vspace{-5pt}
We introduce CORAL, a DiT-based framework that aligns person–garment query–key matches with robust DINOv3 matches. CORAL combines a correspondence distillation loss to guide attention  and an entropy minimization loss to sharpen it. As a result, CORAL consistently improves over the baseline across standard and in-the-wild benchmarks, supported by extensive analyses validating our design choices.

\begin{ack}
This work was partly supported by the Institute of Information \& Communications Technology Planning \& Evaluation (IITP) grant funded by the Korea government (MSIT) (No. RS-2025-25441313, Professional AI Talent Development Program for Multimodal AI Agents, Contribution: 50\%) and by National IT Industry Promotion Agency (NIPA) grant funded by the Korea government (MSIT) (Development of Physical Intelligence Action Models and an Integrated Framework, No. RS-2026-25621677, Contribution: 50\%).

\end{ack}

\bibliographystyle{plainnat}
\bibliography{neurips_2026_ref}

\newpage
\appendix
\section*{\LARGE Appendix}
\noindent This appendix provides supplementary material to support the main paper.

\bigskip
\noindent{\Large\textbf{Contents}}\\[2pt]
\noindent\rule{\linewidth}{0.5pt}

\setcounter{section}{0}
\setcounter{table}{0}
\setcounter{figure}{0}
\setcounter{equation}{20}
\renewcommand{\thesection}{\Alph{section}}
\renewcommand{\thefigure}{\thesection.\arabic{figure}}
\renewcommand{\thetable}{\thesection.\arabic{table}}
\newcommand{\tocsec}[2]{\par\noindent\hbox to \linewidth{\textbf{#1}\dotfill\textbf{\pageref{#2}}}\par}
\newcommand{\tocsub}[2]{\par\noindent\hbox to \linewidth{\hspace*{1.5em}#1\dotfill\pageref{#2}}\par}
\newcommand{\tocgap}{\vspace{3pt}}

\begingroup
\setlength{\parskip}{0pt}

\tocsec{A\quad Implementation Details}{sec:implementation_details}
\tocsub{A.1\quad Training Setup}{sec:training_setup}
\tocsub{A.2\quad Pseudo Ground-Truth Construction}{sec:impl_detail_on_psuedo_gt}
\tocgap
\tocsec{B\quad Training Dataset}{supple:dataset_details}
\tocsub{B.1\quad Dataset Overview}{supple:dataset_details}
\tocsub{B.2\quad Dataset Preprocessing}{supple:dataset_processing}
\tocgap
\tocsec{C\quad Detailed Method Analysis}{supple:method_analysis}
\tocsub{C.1\quad Choice of Supervision Signal}{supple:supervision_signal}
\tocsub{C.2\quad Choice of Mechanism}{supple:supervision_mechanism}
\tocsub{C.3\quad Choice of Alignment Target}{supple:alignment_target}
\tocgap
\tocsec{D\quad Additional Analysis on Pseudo Ground-Truth}{supple:analysis_on_pseudo_gt}
\tocsub{D.1\quad Robustness to Pseudo-GT Errors}{supple:pseduo_gt_errors}
\tocsub{D.2\quad Generality across Pseudo-GT Sources}{supple:pseduo_gt_generality}
\tocsub{D.3\quad Behavior under Challenging Conditions}{supple:behavior_under_challenging_conditions}
\tocgap
\tocsec{E\quad Additional Ablation Studies}{supple:hyperparam_sensitivity}
\tocsub{E.1\quad Hyperparameter Sensitivity}{supple:hyperparam_sensitivity}
\tocsub{E.2\quad Additional Visualizations of Loss Component Ablations}{supple:attention_output}
\tocgap
\tocsec{F\quad Additional Details on Motivation Experiment}{supple:details_on_motiv_exp}
\tocgap
\tocsec{G\quad Runtime and Memory}{sec:runtim_memory}
\tocgap
\tocsec{H\quad CORAL on Other Baselines}{supple:coral_across_pose_injection_strategies}
\tocsub{H.1\quad Baselines with Different Pose Injection Strategies}{supple:coral_across_pose_injection_strategies}
\tocsub{H.2\quad An Additional Efficient Baseline}{supple:coral_on_efficient_baseline}
\tocsub{H.3\quad Baseline with a Different Backbone}{sec:flux2_result}
\tocgap
\tocsec{I\quad VLM-based Evaluation}{supple:vlm_eval}
\tocgap
\tocsec{J\quad Human Evaluation}{supple:human_eval}
\tocgap
\tocsec{K\quad Evaluation on Additional Datasets}{supple:additional_evaluation}
\tocsub{K.1\quad Motivation}{supple:additional_evaluation_motivation}
\tocsub{K.2\quad Evaluation Dataset Details}{supple:add_eval_details}
\tocgap
\tocsec{L\quad Person-To-Person Garment Transfer}{supple:applications}
\tocgap
\tocsec{M\quad Additional Qualitative Results}{supple:additional_results}
\tocgap
\tocsec{N\quad Limitations and Discussion}{supple:dicussion}

\endgroup

\noindent\rule{\linewidth}{0.5pt}
\clearpage
\section{Implementation Details}
\label{sec:implementation_details}

\subsection{Training Setup}
\label{sec:training_setup}
We initialize our model from FLUX.1-Fill-dev and finetune the attention layers. We train with the AdamW~\cite{loshchilov2019decoupledweightdecayregularization} optimizer using a learning rate of $2\times10^{-5}$ and a batch size of 16 for a total of 8.4k steps on 8 NVIDIA B200 GPUs. We train and report results at a resolution of $1024\times768$. For training with $\mathcal{L}_{\text{CORAL}}$, same learning rate and batch size are used, with loss weights $\lambda_{\text{corr}} = 0.01$ and $\lambda_{\text{ent}} = 0.1$.

\subsection{Pseudo Ground-Truth Construction}
\label{sec:impl_detail_on_psuedo_gt}
We construct pseudo ground-truth correspondences using DINOv3-B~\cite{siméoni2025dinov3} and set the cycle-consistency threshold $\gamma$ as 3. The DiT operates on latents from an $8\times$ VAE with a $2\times2$ patchify, so each attention token corresponds to a $16\times16$ pixel region, giving a $64\times48$ token grid at $1024\times768$. DINOv3-B has a patch size of 16 and receives the same $1024\times768$ input, so its token grid is also $64\times48$. The pseudo ground-truth is thus defined directly on the grid on which the loss operates, with no resampling. When a matcher has a different patch size, we adjust its input resolution so that it produces the same token grid, rather than interpolating its matches afterwards. We compute $\mathcal{L}_{\text{CORAL}}$ over all layers throughout training and average it across layers, with both terms restricted to query tokens that pass the cycle-consistency check (Eqs.~\ref{eq:corr_loss} and~\ref{eq:entropy_loss}).

\section{Training Dataset} %
\label{supple:dataset_details}
\subsection{Dataset Overview} VITON-HD~\cite{choi2021vitonhdhighresolutionvirtualtryon} comprises 11,647 training images and 2,032 test images and focuses on upper-body garments captured in a studio setting. DressCode~\cite{morelli2022dresscodehighresolutionmulticategory} consists of 48,392 training set and 5,400 test set, covering 13,563 upper, 7,151 lower and 27,678 dresses. The person images are generally full-body centered, with face region excluded. In both datasets, garment product images are typically captured on plain, white backgrounds and presented as flat product shots. Overall, person images in both benchmarks are captured under controlled backgrounds and lighting, and most samples present static, frontal poses with limited viewpoint variation.

\subsection{Dataset Preprocessing}
\label{supple:dataset_processing}
We follow prior works~\cite{chong2025catvtonconcatenationneedvirtual, huang2024incontextloradiffusiontransformers} for input prompts, using a fixed input prompt for VITON-HD and adding the garment category for DressCode. For VITON-HD, we use both the masks from benchmarks and ones preprocessed by prior work~\cite{choi2024improvingdiffusionmodelsauthentic} in a 1:1 ratio. For DressCode, we adopt masks preprocessed by prior work~\cite{choi2024improvingdiffusionmodelsauthentic}, and at each training step we apply a random directional dilation with a radius uniformly sampled from 1 to 20 pixels to augment the masks. Input masks can cover cloth-agnostic regions such as hair or bags, which makes it difficult to preserve their appearance from the input under occlusions. While one could refine the masks to exclude such regions, this comes with practical risks: if the masks become too tight, they can reveal the originally worn garment, while accurately segmenting fine structures such as hair across diverse poses and viewpoints is often costly and error-prone. Instead, we extract these regions using human-parser~\cite{li2019selfcorrectionhumanparsing} or SAM3~\cite{carion2025sam3segmentconcepts}, and provide them as additional guidance by stitching them into the DensePose~\cite{güler2018denseposedensehumanpose} condition.

\clearpage
\section{Detailed Method Analysis}
\label{supple:method_analysis}

We provide a detailed analysis of CORAL's design from three perspectives: what kind of 
signal to supervise with (Sec.~\ref{supple:supervision_signal}), how 
correspondence should be used in the model (Sec.~\ref{supple:supervision_mechanism}), 
and where in the generation model to apply supervision 
(Sec.~\ref{supple:alignment_target}). We discuss CORAL's design choices 
along these aspects and analyze why they are effective.

\subsection{Choice of Supervision Signal: With What Kind of Signal?}
\label{supple:supervision_signal}

We can supervise person-garment alignment with 
different signals. We compare 3 forms used in related approaches and motivate CORAL's choice.

\paragraph{Photometric Reconstruction.}
Photometric reconstruction is the most widely used form 
in VTON, computing reconstruction loss between the 
warped or generated output and the ground-truth person 
image. This is dominant in GAN-based 
methods~\cite{xie2023gpvtongeneralpurposevirtual, 
choi2021vitonhdhighresolutionvirtualtryon} that 
supervise on warped outputs from explicit flow 
estimators, and is also adopted in recent 
diffusion-based VTON 
methods~\cite{zhou2024learningflowfieldsattention}. However, 
photometric reconstruction relies on the assumption 
that pixel intensities are preserved between 
corresponding regions, which is often violated in 
practice due to lighting variations, material 
reflectance differences, and non-rigid deformations of 
garments on the 
body~\cite{wang2018characteristicpreservingimagebasedvirtualtryon}.

\paragraph{Global Semantic Alignment.}
A natural question is whether attention should disperse 
across all semantically similar regions, since they would 
share the same semantic identity. Recent works such as 
REPA~\cite{yu2025representationalignmentgenerationtraining} 
align the model's intermediate features with semantic 
descriptors from vision foundation models. However, 
multiple regions in a garment often share similar 
semantic features (e.g., repeated stripe patterns), 
and attending to all such regions introduces ambiguity 
in spatial correspondence. In contrast, geometric 
correspondence enforces a single, geometrically 
consistent match, providing a more reliable signal for 
generation.

\paragraph{Local Geometric Correspondence \textbf(Ours).}
CORAL adopts local geometric correspondence as its 
supervision signal. From features extracted by a vision 
foundation model, we 
compute pairwise similarity between every person token 
position and every garment token position, and extract 
for each person token the garment position of maximum 
similarity. This yields, for each person token, a 
target position in the garment that the model's 
attention should be directed toward. 
The supervision is then applied to the model's Q-K 
matching pattern. Unlike global semantic alignment, 
where the supervision target is the model's 
intermediate features, our supervision target is 
positional: it ties each person token to a specific 
garment location rather than to a semantic identity 
shared across multiple regions. This resolves the 
dispersion ambiguity faced by both photometric 
reconstruction and global semantic alignment. Since 
supervision operates on the matching pattern rather 
than the features themselves, it preserves the model's 
generative representations.

To empirically compare these supervision signals, we 
reproduce reconstruction-based supervision and global 
semantic alignment on our DiT baseline. As shown in 
Tab.~\ref{tab:supervision_signal}, local geometric 
correspondence supervision (CORAL) yields the best 
performance across all metrics, validating our choice.

\begin{wraptable}{r}{0.6\columnwidth}
\vspace{-10pt}
\centering
\resizebox{\linewidth}{!}{
\setlength{\tabcolsep}{6pt}
\begin{tabular}{lccc}
\toprule
Supervision Signal & SSIM \up & LPIPS \down & FID \down \\
\midrule
Baseline (no supervision)                       & 0.889 & 0.055 & 5.543 \\
Reconstruction-based              & 0.900 & 0.051 & 5.542 \\
Global semantic alignment & 0.887 & 0.059 & 5.724 \\
\textbf{Local geometric correspondence (Ours)}  & \textbf{0.908} & \textbf{0.048} & \textbf{5.009} \\
\bottomrule
\end{tabular}}
\vspace{5pt}
\caption{\textbf{Comparison of supervision signals on 
our DiT baseline.} Local geometric correspondence 
supervision yields the best performance compared to 
reconstruction-based and global semantic alternatives 
reproduced on the same DiT backbone.}
\label{tab:supervision_signal}
\vspace{-15pt}
\end{wraptable}

\paragraph{Implementation Details for Reproduction.}
Both alternatives are reproduced by replacing the 
supervision target in our loss 
formulation~\eqref{eq:corr_loss} with different 
quantities, while keeping the attention map $A$ 
and training configuration unchanged. For \textbf{photometric reconstruction}, we leverage 
reference RGB values from the person image. The 
input images are downsampled to match the attention 
map resolution:
\begin{equation}
\mathcal{L}_{\text{photo}} = \frac{1}{|\mathcal{P}|} 
\sum_{j \in \mathcal{P}} \left\| 
\sum_{i \in \mathcal{G}} A_{ij} \, \tilde{I}_g[i] - 
\tilde{I}_p[j] \right\|_2^2,
\end{equation}
where $\tilde{I}_g, \tilde{I}_p$ denote the 
bilinear-downsampled garment and person images at 
the latent resolution. For \textbf{global semantic alignment}, we replace 
the position-based target with semantic features 
extracted by 
DINOv3~\cite{siméoni2025dinov3}:
\begin{equation}
\mathcal{L}_{\text{align}} = \frac{1}{|\mathcal{P}|} 
\sum_{j \in \mathcal{P}} \left\| 
\sum_{i \in \mathcal{G}} A_{ij} \, \phi(I_g)_i - 
\phi(I_p)_j \right\|_2^2,
\end{equation}
where $\phi$ denotes the DINOv3 feature extractor 
and $\phi(\cdot)_i$ is the feature at token $i$.

\subsection{Choice of Mechanism: How should Correspondence be Used?}
\label{supple:supervision_mechanism}

The main paper empirically demonstrates that feature injection approaches yield limited gains on our DiT baseline. Here we analyze the architectural reasons: conditioning, which feature injection is one mechanism of, is inherently limited in DiT-based VTON, while direct supervision is more effective.

\paragraph{Conditioning Approaches in Prior VTON.}
Prior correspondence-based VTON methods~\cite{chen2024wearanywaymanipulablevirtualtryon, wan2025incorporatingvisualcorrespondencediffusion} integrate sparse correspondences into 
the model through conditioning: garment features 
extracted at matched positions are added to the model as additional input, and the model is trained only with 
the diffusion loss. The model is not explicitly guided 
to preserve the correspondence information carried by 
the added features, as illustrated in Fig.~\ref{fig:motivation}

\paragraph{Information Flow in UNet vs DiT Attention.}
Such conditioning approaches may operate on UNet-based 
dual-branch architectures, where garment 
information reaches the person branch through a single 
pathway: Garm-UNet features fed into augmented 
self-attention. Formally, $h_g^l = \text{GarmUNet}^l(I_g)$ 
depends solely on the garment image, so a correspondence 
feature added along this pathway has a consistent, 
predictable effect with no alternative route to bypass 
it.

In contrast, DiT-based VTON employs unified full 
attention where garment, person, pose, and text tokens 
become entangled at every layer: $h_g^l = f^l(I_g, I_p, 
\text{text}, \text{pose})$. Unlike UNet, unified 
attention offers many alternative routes for person 
tokens to access garment information, including indirect 
paths through tokens that have already incorporated it. 
A correspondence feature added through conditioning is 
only one route among many: if the model can obtain 
garment information through other routes, the 
correspondence signal is effectively bypassed without 
any explicit incentive to preserve it.

\paragraph{Empirical Verification.}
This is empirically verified in 
Section~\ref{sec:prior_comparison} of the main paper, 
where SPM-Diff~\cite{wan2025incorporatingvisualcorrespondencediffusion} is reproduced on our 
DiT baseline. Both the original sparse formulation 
with $\sim$25 matched points and a dense extension 
that uses correspondences at all positions yield 
negligible or negative improvement over the baseline, 
consistent with our architectural analysis above.

\paragraph{Implementation Details for Reproduction.}
We implement a feature injection conditioning 
variant on our DiT baseline, following the design 
of SPM-Diff~\cite{wan2025incorporatingvisualcorrespondencediffusion}. 
Projected DINOv3 features are injected into 
attention layers by adding flow-warped garment 
features to person-side query and key, and original 
garment features to garment-side key, with 
per-layer zero-initialized linear projections. As 
the inference code is not publicly available, we 
implement this conditioning from the paper 
description. We test two densities: a sparse 
variant using $\sim$25 matched points (as in the 
original formulation) and a dense variant using 
correspondences at all positions.

As shown in Tab.~\ref{tab:spm_diff_comparison}, 
both variants yield negligible or negative 
improvement compared to the baseline, consistent 
with our analysis. In contrast, CORAL's direct 
supervision explicitly enforces correspondence 
within attention, yielding substantial improvement 
across all metrics.

\clearpage
\subsection{Choice of Alignment Target: Where to Apply Supervision?}
\label{supple:alignment_target}

Given that direct supervision is adopted, the final 
question is where within the generation model the 
supervision should be applied. We compare two alignment 
targets and motivate CORAL's choice of Q-K matching.

\paragraph{Intermediate Features vs Q-K Matching.}
We compare two alignment targets within DiT: aligning 
the model's intermediate features with feature 
descriptors from vision foundation models, as in 
prior
approach~\cite{yu2025representationalignmentgenerationtraining}, 
versus aligning the Q-K similarity between person and 
garment tokens, as in CORAL. The two targets differ in 
two key aspects.

First, in what they preserve in the model. Prior 
analyses~\cite{nam2024dreammatcher} show that query-key 
similarities largely determine the spatial arrangement 
of generated image elements while value features mainly 
carry appearance details. Since intermediate features 
jointly encode both spatial and appearance information, 
aligning them directly can disrupt the model's learned 
appearance representation. CORAL supervises only the 
query-key matches, correcting spatial alignment without 
altering the value pathway, thus preserving the model's 
appearance representation.

Second, in their training assumptions. REPA
alignment is designed to accelerate training from 
scratch rather than to fine-tune a pretrained 
model~\cite{zhang2025videorepalearningphysicsvideo}. 
Hard alignment pulls intermediate features toward 
external descriptors, disrupting the inherent 
generative representations learned during pretraining 
and leading to slower convergence during fine-tuning. 
CORAL operates only on attention similarities, leaving 
the underlying features intact.

\paragraph{Implementation Details for Feature Alignment.}
We build the intermediate feature alignment baseline 
upon the same architectural modifications adopted in our 
main baseline described in 
Sec.~\ref{DIT-based VTON Baseline}. Following 
REPA~\cite{yu2025representationalignmentgenerationtraining}, 
we attach a lightweight projection head to the output of 
each transformer block, where each head is a three-layer 
MLP with SiLU activations that maps DiT hidden states 
into the DINOv3~\cite{siméoni2025dinov3} embedding 
space. We use DINOv3-B as the external encoder, 
consistent with our default configuration.

Given a garment image $I_{\mathrm{g}}$ and a person 
image $I_{\mathrm{p}}$, we concatenate them horizontally 
to form $\hat{I}\in\mathbb{R}^{H\times 2W\times 3}$, and 
extract DINOv3 features $\phi_{\mathrm{g}}, 
\phi_{\mathrm{p}} \in \mathbb{R}^{hw\times 768}$ from 
the frozen encoder. We concatenate them as:
\begin{equation}
\hat{\phi}=[\, \phi_{\mathrm{g}} \, \| \, \phi_{\mathrm{p}} \,]\in\mathbb{R}^{2hw\times 768}.
\end{equation}

Let $\mathbf{h}$ be an intermediate hidden representation 
from the diffusion transformer $\mathbf{v}_\theta$, and 
let $f_\omega(\cdot)$ be a trainable projection head. We 
define a patch-wise alignment loss:
\begin{equation}
\mathcal{L}_{\mathrm{REPA}}(\theta, \omega) := - \mathbb{E}_{\hat{I}, \epsilon, t} \left[\frac{1}{K}\sum_{k=1}^{K}\operatorname{sim} \big( \hat{\phi}_k, f_\omega(\mathbf{h}_k) \big) \right], 
\label{eq:repa}
\end{equation}
where $k$ indexes patches and $K$ is the number of 
patches. We compute this loss at the output of each 
transformer block using its corresponding projection 
head, and average it across layers.

The final objective adds the projection loss to the 
standard flow-matching loss:
\begin{equation}
    \mathcal{L}_{\mathrm{total}}
    = \mathcal{L}_{\mathrm{velocity}}
    + \lambda_{\mathrm{REPA}} \, \mathcal{L}_{\mathrm{REPA}},
\label{eq:feature_matching_loss}
\end{equation}
where we set $\lambda_{\mathrm{REPA}} = 0.1$ in all 
experiments.

\begin{table}[h]
\vspace{-5pt}
\resizebox{1\linewidth}{!}{
\centering
\small

\begin{tabular}{clcccccc}
\toprule
\multirow{2}{*}{\textbf{}} & \multirow{2}{*}{\textbf{Methods}} &
\multicolumn{4}{c}{\textbf{Paired}} & \multicolumn{2}{c}{\textbf{Unpaired}} \\
\cmidrule(lr){3-6} \cmidrule(lr){7-8}
& & \mc{SSIM \up} & \mc{LPIPS \down} & \mc{FID \down} & \mc{KID \down} & \mc{FID \down} & \mc{KID \down} \\
\midrule
(I)  & Baseline & 0.889 & 0.055 & 5.543 & 0.870 & 9.641 & 1.323 \\
(II)  & (I) + Feature Alignment (REPA)~\cite{yu2025representationalignmentgenerationtraining}
& 0.885 (\textcolor{gray}{${-0.45\%}$}) & 0.087 (\textcolor{gray}{${-58.18\%}$})  & 13.718 (\textcolor{gray}{${-147.48\%}$}) & 2.103 (\textcolor{gray}{${-141.72\%}$}) & 15.621 (\textcolor{gray}{${-62.03\%}$}) & 3.684 (\textcolor{gray}{${-178.46\%}$})\\
(III) & (I) + Query-Key Correspondence Alignment \textbf{(Ours)}
&\textbf{0.907} (\textcolor{teal}{+2.02\%}) & \textbf{0.048} (\textcolor{teal}{+12.73\%}) & \textbf{4.962} (\textcolor{teal}{+10.48\%}) & \textbf{0.565} (\textcolor{teal}{+35.06\%}) & \textbf{8.763} (\textcolor{teal}{+9.11\%}) & \textbf{0.880} (\textcolor{teal}{+33.48\%})  \\
\bottomrule
\end{tabular}
}
\vspace{+5pt}
\captionof{table}{\textbf{Alignment Target Ablation on VITON-HD \cite{choi2021vitonhdhighresolutionvirtualtryon}.}}
\label{tab:alignment_target_abl}
\end{table}

\paragraph{Comparison and Analysis.}
\begin{figure}[!h]
    \centering
    \includegraphics[width=1\linewidth]
    {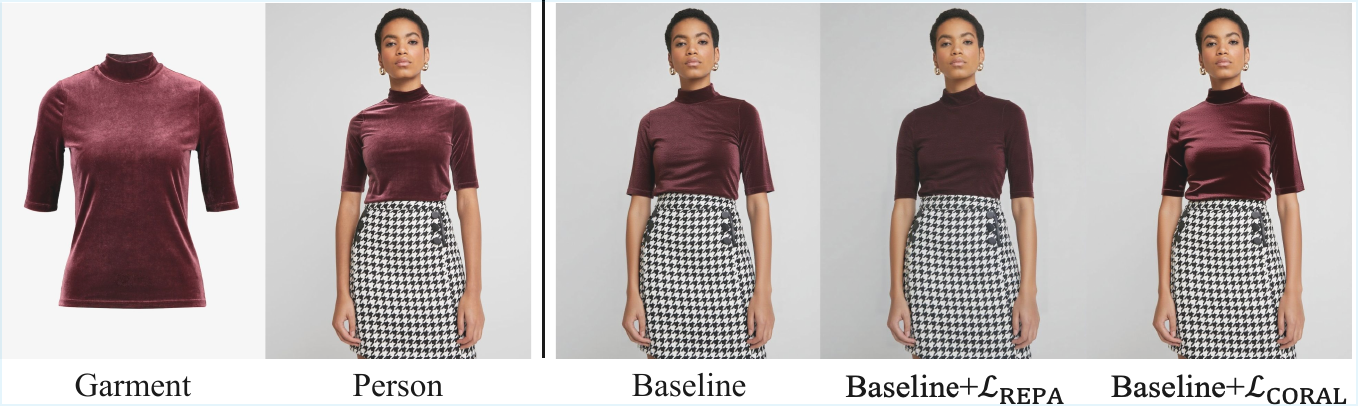} 
    \caption{\textbf{Intermediate Feature Alignment vs. Ours.}}
    \label{fig:feature selection}
\end{figure}

Tab.~\ref{tab:alignment_target_abl} compares the two 
alignment targets, showing that CORAL achieves better 
try-on quality with higher SSIM and lower LPIPS. Since 
intermediate features jointly encode both spatial and 
appearance information, directly aligning them affects 
the inherent representations learned during pretraining. 
By supervising query-key matches, CORAL strengthens 
person-garment alignment while largely preserving the 
model's learned appearance representation. In contrast, 
feature-level alignment shows a larger change in 
intermediate features (Fig.~\ref{fig:feature selection}), 
followed by a substantial performance drop across all 
metrics in Tab.~\ref{tab:alignment_target_abl}.

\clearpage
\section{Additional Analysis on Pseudo Ground-Truth}
\label{supple:analysis_on_pseudo_gt}
\subsection{Robustness to Pseudo-GT Errors.} 
\label{supple:pseduo_gt_errors}
A natural concern is that the pseudo-GT correspondence 
may contain errors in challenging cases (e.g., complex 
poses, occlusions, ambiguous garment regions), which 
could propagate into the training signal. We address 
this concern through cycle consistency filtering: a 
correspondence between a person token and a garment 
position is retained only if the round-trip mapping 
(person $\to$ garment $\to$ person) returns to within 
a threshold of the original person position. 
Correspondences that fail this check are excluded from 
the supervision signal.

\begin{wraptable}{r}{0.6\columnwidth}
\vspace{-10pt}
\centering
\resizebox{\linewidth}{!}{
\setlength{\tabcolsep}{6pt}
\begin{tabular}{cccccc}
\toprule
Threshold & Avg.\ \#reliable & SSIM \up & LPIPS \down & FID \down \\
\midrule
$\infty$ (no filtering)  & 6.2M & 0.905 & 0.057 & 5.098 \\
6                        & 5.8M & 0.907 & 0.050 & 5.021 \\
\textbf{3 (default)}     & 4.2M & \textbf{0.908} & \textbf{0.048} & \textbf{5.009} \\
\bottomrule
\end{tabular}}
\vspace{5pt}
\caption{\textbf{Effect of cycle-consistency threshold 
on VITON-HD.} Stricter filtering retains fewer but 
cleaner correspondences and yields the best performance.}
\label{tab:cycle_filtering}
\end{wraptable}

Tab.~\ref{tab:cycle_filtering} reports the effect of 
varying the cycle-consistency threshold on VITON-HD. 
Stricter filtering (threshold of 3) yields the best 
performance, confirming that cycle consistency 
effectively removes erroneous correspondences. Even 
without filtering ($\infty$), CORAL still improves 
over the baseline, indicating that the supervision 
also tolerates a moderate amount of noise.

\subsection{Generality across Pseudo-GT Sources.}
\label{supple:pseduo_gt_generality}
\begin{wraptable}{r}{0.55\columnwidth}
\vspace{-10pt}
\centering
\resizebox{\linewidth}{!}{
\centering
\setlength{\tabcolsep}{6pt}
\begin{tabular}{lccc}
\toprule
Pseudo-GT Source & SSIM \up & LPIPS \down & FID \down \\
\midrule
Baseline (no supervision)         & 0.889 & 0.055 & 5.543 \\
DINOv2 ~\cite{oquab2024dinov2learningrobustvisual}                            & 0.905 & 0.055 & 5.099 \\
SD + DINOv2  ~\cite{zhang2023talefeaturesstablediffusion}                     & 0.907 & 0.049 & 5.087 \\
LoMa ~\cite{nordström2026lomalocalfeaturematching}                             & 0.891 & 0.054 & 5.540 \\
DIFT   ~\cite{tang2023emergentcorrespondenceimagediffusion}                           & 0.902 & 0.055 & 5.321 \\
\textbf{DINOv3 (default)} ~\cite{siméoni2025dinov3}        & \textbf{0.908} & \textbf{0.048} & \textbf{5.009} \\
\bottomrule
\end{tabular}}
\vspace{5pt}
\caption{\textbf{Generality across pseudo-GT sources on 
VITON-HD.} CORAL improves over the baseline regardless 
of which correspondence source is used.}
\label{tab:encoder_comparison}
\vspace{-10pt}

\end{wraptable}

\begin{figure}[!h]
    \centering
    \includegraphics[width=1\linewidth]{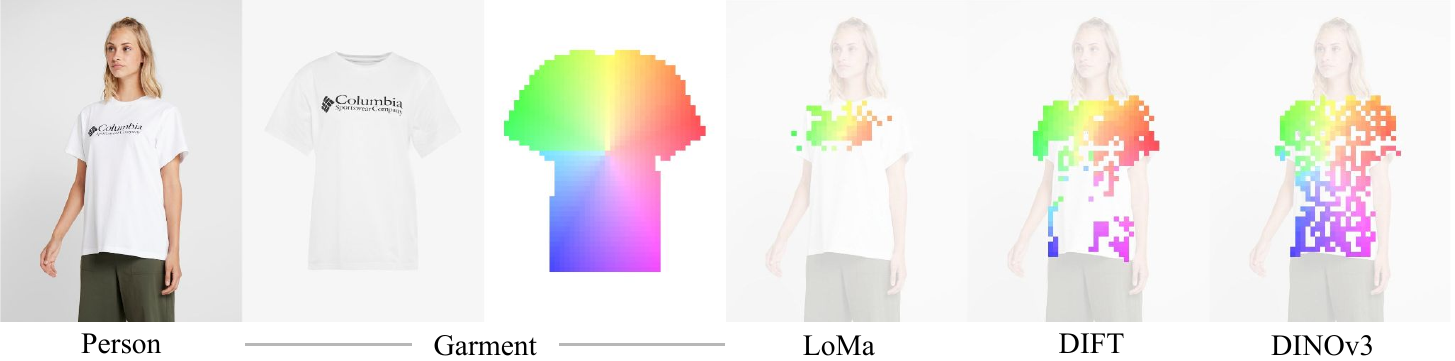}
    \caption{\textbf{Visualization of different pseudo-correspondence 
sources.} For each spatial position on the garment, 
color-coded with a rainbow position map, we visualize 
the matched location on the person predicted by 
different correspondence sources: LoMa, DIFT, and 
DINOv3. DINOv3 produces the most spatially coherent 
mappings, while LoMa yields sparser and less 
consistent matches. Our method is compatible with 
any of these sources, with DINOv3 providing the strongest supervision signal.}
    \label{fig:psuedo_gt_vis}
\end{figure}

CORAL relies on an external matching method to provide 
pseudo ground-truth correspondences. The choice of this 
matching method is independent of CORAL's design: any 
method that produces reliable correspondences between 
person and garment images can be \textbf{plugged in}. To verify 
this, we evaluate four alternatives spanning two 
categories: vision foundation models repurposed for 
correspondence (DINOv2~\cite{oquab2024dinov2learningrobustvisual}, DINOv3 
(our default)~\cite{siméoni2025dinov3}, and DINOv2 combined with 
diffusion features~\cite{zhang2023talefeaturesstablediffusion}), and 
specialized dense matching methods such as 
LoMa~\cite{nordström2026lomalocalfeaturematching} and DIFT~\cite{tang2023emergentcorrespondenceimagediffusion}.

Fig.~\ref{fig:psuedo_gt_vis} visualizes the 
correspondences extracted from each method on the same 
person-garment pair. While the matching quality varies 
across methods, all of them produce reasonable 
person-garment correspondences that can serve as the 
supervision signal for CORAL.

Tab.~\ref{tab:encoder_comparison} reports the 
quantitative results when each correspondence source is 
used. All four alternatives lead to substantial 
improvement over the baseline, confirming that CORAL is 
general across pseudo-GT sources. We also observe a 
slight performance increase with stronger correspondence 
sources, suggesting that future improvements in matching 
methods can directly benefit CORAL without any 
architectural modification.

\subsection{Behavior under Challenging Conditions.}
\label{supple:behavior_under_challenging_conditions}
\begin{figure}[!h]
    \centering
    \includegraphics[width=0.9\linewidth]
    {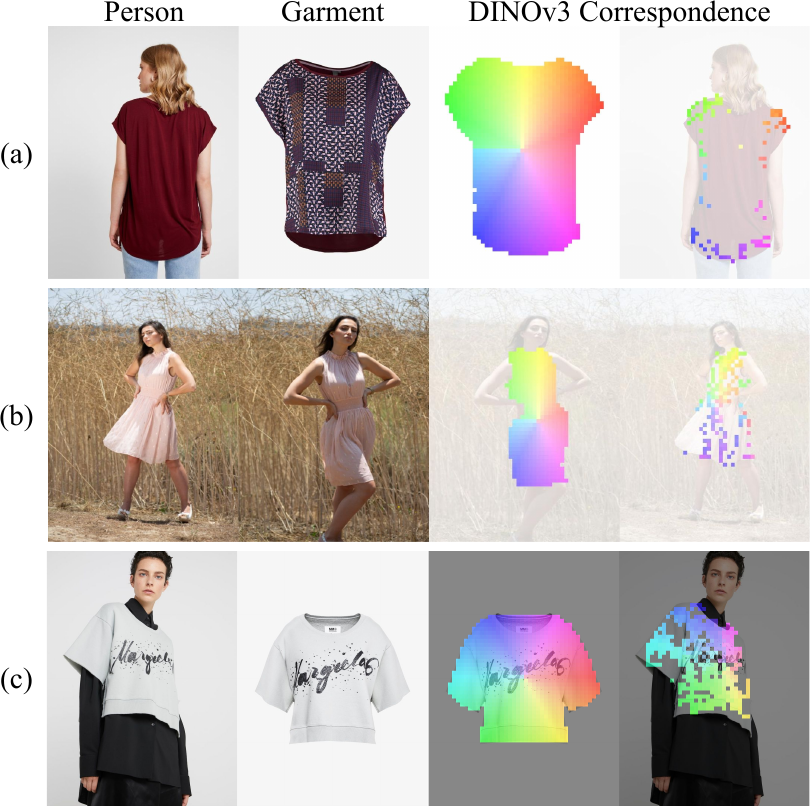} 
    \caption{\textbf{DINOv3 correspondence under extreme viewpoint, illumination variation, and layered garments.} For each garment position (color-coded with rainbow map), we visualize matched person locations predicted by DINOv3. \textbf{(a)} Opposite viewpoints from VITON-HD dataset (back-view person, front-view garment): DINOv3 distinguishes the two views, with correspondences remaining mainly along view-invariant boundary regions. \textbf{(b)} Significant illumination change: matches concentrate in regions with consistent appearance. \textbf{(c)} Layered garments from VITON-HD: matches on regions without a counterpart in the reference, such as the inner garment worn beneath it, are removed. Although coverage varies under such conditions, our supervision benefits from these partial yet reliable matches.}
    \label{fig:psuedo_under_challenging}
\end{figure}

DINOv3 correspondence quality varies with input 
characteristics. 
Fig.~\ref{fig:psuedo_under_challenging} shows three 
representative cases: \textbf{(a)} extreme 
viewpoint mismatch between person and garment 
(from VITON-HD), where DINOv3 distinguishes the 
two views and retains matches mainly along 
view-invariant boundary regions; \textbf{(b)} 
significant illumination change (from PPR10K), 
where matches concentrate in regions with 
consistent appearance; and \textbf{(c)} layered 
garments (from VITON-HD), where matches 
on regions without a counterpart in the reference 
are removed.

In all three cases, the removed regions are those without a reliable counterpart in the reference: the back of the body in (a), regions altered by strong illumination in (b), and the inner garment worn beneath the target garment in (c), which do not appear in a flat product photo. These regions are trained only with the diffusion loss, as in the baseline. Since they have no reliable position to learn, removing their matches does not lose useful supervision, and keeping them by loosening the filter hurts performance (Tab.~\ref{tab:cycle_filtering}). The check does not catch every case (Appendix~\ref{supple:dicussion}), but the remaining regions, such as view-invariant boundaries, consistently lit regions, and outer garment parts, still receive supervision and anchor the garment structure.

\begin{wraptable}{r}{0.7\linewidth}
    \centering
    \vspace{-5pt}
    \setlength{\tabcolsep}{8pt}
    \resizebox{\linewidth}{!}{%
    \begin{tabular}{lcccc}
        \toprule
        Threshold $\gamma$ & Avg.\ \# reliable points & SSIM \up & LPIPS \down & FID \down \\
        \midrule
        w/o pseudo-GT & --- & 0.897 & 0.068 & 52.201 \\
        $\infty$      & 3.8M & 0.899 & 0.064 & 51.995 \\
        $6$           & 2.5M & 0.901 & 0.061 & 51.892 \\
        \textbf{$3$ (default)} & \textbf{1.1M} & \textbf{0.929} & \textbf{0.050} & \textbf{51.423} \\
        \bottomrule
    \end{tabular}}
        \caption{\textbf{Robustness of cycle-consistency filtering on PPR10K.} Stricter thresholds yield better metrics, with $\gamma = 3$ providing the best balance between supervision density and pseudo-label quality.}
        \label{tab:cycle_filtering_ppr10k}
        \vspace{-5pt}
\end{wraptable}

To assess our method's behavior under in-the-wild 
conditions, we evaluate cycle-consistency 
filtering on 
PPR10K~\cite{liang2021ppr10klargescaleportraitphoto}, 
which contains heavy occlusions, illumination 
changes, and challenging poses. We sweep the 
threshold $\gamma$ across $\{3, 6, \infty\}$ 
(Tab.~\ref{tab:cycle_filtering_ppr10k}). The 
number of reliable correspondences after filtering 
is significantly lower on PPR10K (1.1M vs.\ 4.2M 
at $\gamma=3$ on VITON-HD), yet CORAL still 
improves over the no-supervision baseline across 
all threshold settings, and stricter filtering 
consistently yields better performance. This 
confirms that cycle-consistency filtering 
effectively prevents pseudo-label error 
propagation, and our supervision adapts gracefully 
to varying pseudo-GT density.

\clearpage
\section{Additional Ablation Studies}
\subsection{Hyperparameter Sensitivity}
\label{supple:hyperparam_sensitivity}

We analyze the sensitivity of CORAL to the choice of 
loss weights $\lambda_{\text{corr}}$ and 
$\lambda_{\text{ent}}$ in:
\begin{equation}
    \mathcal{L}_{\text{CORAL}} 
    = \lambda_{\text{corr}}\mathcal{L}_{\text{corr}} 
    + \lambda_{\text{ent}}\mathcal{L}_{\text{ent}}.
    \label{eq:coral_loss}
\end{equation}

\paragraph{Loss Weight Ablation.}
We sweep $\lambda_{\text{corr}}$ and 
$\lambda_{\text{ent}}$ across eight settings on 
VITON-HD, with all other hyperparameters fixed to 
the default training configuration. 
Tab.~\ref{tab:hyperparam_sensitivity} reports the 
results. CORAL maintains stable performance across a 
range of weight values, with the default setting 
($\lambda_{\text{corr}} = 0.01$, 
$\lambda_{\text{ent}} = 0.1$) providing the best 
overall balance. Performance degrades when either 
weight is too small (insufficient supervision 
strength) or too large (loss imbalance disrupting 
the diffusion objective). These results indicate 
that CORAL is not overly sensitive to weight 
selection within reasonable ranges, and the chosen 
defaults generalize without per-dataset tuning.

\begin{table}[h]
\centering
\setlength{\tabcolsep}{8pt}
\begin{tabular}{cc|ccc}
\toprule
$\lambda_{\text{corr}}$ & $\lambda_{\text{ent}}$ & 
SSIM \up & LPIPS \down & FID \down \\
\midrule
0    & 0    & 0.889 & 0.055 & 5.543 \\
0    & 0.1  & 0.905 & 0.052 & 5.482 \\
1.0  & 0.1  & 0.887 & 0.061 & 6.092 \\
0.1  & 0.1  & 0.908 & 0.049 & 5.032 \\
\rowcolor{gray!15} \textbf{0.01} & \textbf{0.1} & 
\textbf{0.908} & \textbf{0.048} & \textbf{5.009} \\
0.01 & 1.0  & 0.907 & 0.049 & 5.018 \\
0.01 & 0.01 & 0.906 & 0.050 & 5.449 \\
0.01 & 0    & 0.906 & 0.051 & 5.451 \\
\bottomrule
\end{tabular}
\vspace{3pt}
\caption{\textbf{Loss weight sensitivity on VITON-HD.} 
Default values are $\lambda_{\text{corr}}=0.01$ and 
$\lambda_{\text{ent}}=0.1$ (highlighted). Both losses 
are stable across a wide range of weights, with 
$\mathcal{L}_{\text{corr}}$ degrading at extreme 
values.}
\label{tab:hyperparam_sensitivity}
\vspace{-10pt}
\end{table}

\subsection{Additional Visualizations of Loss Component Ablations}
\label{supple:attention_output}
\begin{figure}[H]
\centering
\includegraphics[width=0.45\linewidth]{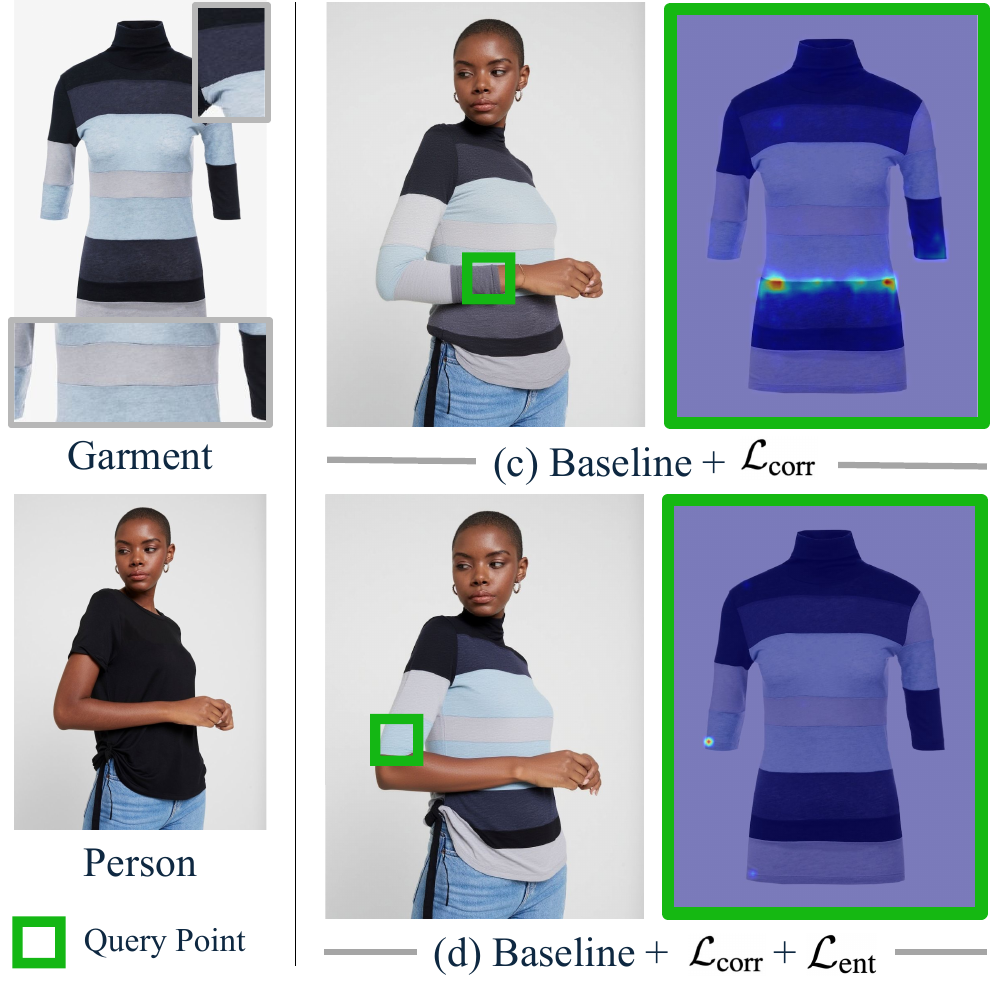}
\caption{\textbf{Additional attention visualizations for Fig.~\ref{fig:attn_map_vis}.} Queries are placed at the sleeve edge, where the outputs of (c) Baseline + $\mathcal{L}_{\text{corr}}$ and (d) CORAL differ.}
\label{fig:attention_output}
\vspace{-10pt}
\end{figure}

Fig.~\ref{fig:attention_output} provides additional attention visualizations for Fig.~\ref{fig:attn_map_vis}. Beyond the query points in Fig.~\ref{fig:attn_map_vis}, we place queries at the sleeve edge, where the outputs of (c) Baseline + $\mathcal{L}_{\text{corr}}$ and (d) CORAL differ. In (c), the attention spreads over the reference and reaches a dark band, whose fabric is drawn onto the arm. In (d), the attention narrows to the light-blue end of the reference sleeve. Since $\mathcal{L}_{\text{corr}}$ is satisfied once the attention averages out near its target, it leaves the attention spread, and content from other regions leaks into the output. Adding $\mathcal{L}_{\text{ent}}$ concentrates the attention on the corresponding position, which directly corrects the visible error.

\section{Additional Details on Motivation Experiment}
\label{supple:details_on_motiv_exp}
We provide additional details and visual analysis for the motivation
experiment in Fig.~\ref{fig:motivation}. The purpose of this experiment is
not to introduce an inference-time guidance method, but to examine whether
correspondence-aware query--key matching can affect the quality of
DiT-based VTON. To this end, we apply a minimal intervention to the baseline
model at inference time and visualize how the generated result changes.

\paragraph{Experimental setup.}
Given a selected query point on the person region, we obtain its corresponding location inside the garment region using the pseudo correspondence described in Sec.~\ref{supple:analysis_on_pseudo_gt}. We then modify only its query--key matching to garment keys around the matched location. All other tokens, model weights, and intermediate feature representations remain unchanged. This allows us to observe the effect of locally guiding person-to-garment matching without changing the overall model.

\paragraph{Implementation details.}
For the matched garment location, we apply the correspondence bias to a small \(5\times5\) neighborhood. This provides a localized guidance region while avoiding an overly restrictive single-token constraint. Our entropy-based objective shows a consistent effect in Fig.~\ref{fig:generalization_across_baselines}, reducing overly diffuse attention and bringing the attention entropy closer to the level observed in stronger baselines.

\begin{figure}[!h]
    \centering
    \includegraphics[width=0.85\linewidth]
    {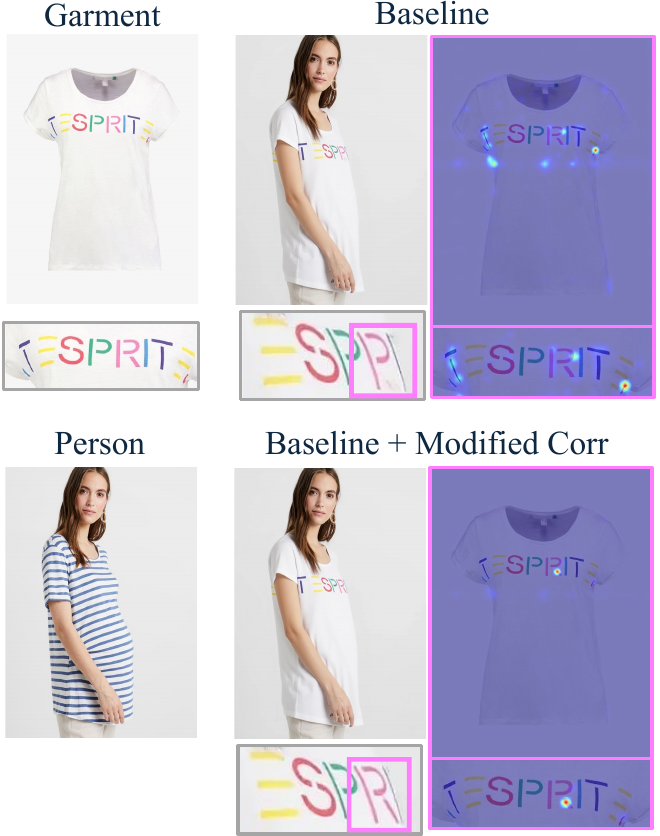} 
    \caption{\textbf{Zoomed-in Visualization of the Motivation Experiment.} \textcolor{magenta}{Pink} markers denote the query points (the bottom-right corner of the letter ``R'').}
    \label{fig:motivation_exp_details}
\end{figure}

\paragraph{Results.}
As shown in Fig.~\ref{fig:motivation_exp_details}, the baseline attention for the selected query point attends to an inaccurate or diffuse garment region. In the corresponding try-on result, this leads to imperfect transfer of local garment appearance, such as misplaced text or distorted patterns. After applying the local query--key bias, the selected person token attends more strongly to the corresponding garment region, and the generated result better preserves the local garment appearance around the queried area.

This experiment should be interpreted as a diagnostic analysis rather than a practical inference-time method. Since the intervention is applied only to selected query--key matching and does not update model parameters, the improved result suggests that query--key matching is a meaningful control point for person--garment correspondence in DiT-based VTON. Motivated by this observation, CORAL replaces manual inference-time intervention with train-time correspondence supervision, encouraging the model to learn correspondence-aware query--key matching directly.

\section{Runtime and Memory}
\label{sec:runtim_memory}
\begin{table}[h]
\centering
\small
\setlength{\tabcolsep}{4pt}
\begin{tabular}{cccccc}
\toprule
$\mathcal{L}_{\text{CORAL}}$ & GPU & GPU\# (T/I) & Batch (T/I) & Speed (T/I) & VRAM (T/I) \\
\midrule
 \ding{51}                   & H100  & 4 / 1 & 1 / 1 & 0.42 / 0.063 & 56 / 40 \\
\textcolor{gray}{\ding{55}} & H100  & 4 / 1 & 1 / 1 & 0.45 / 0.063 & 52 / 40 \\
\bottomrule
\end{tabular}
\vspace{+5pt}
\captionof{table}{\textbf{Speed and memory.} T denotes training and I denotes inference.
Speed is in img/sec and VRAM is in GB.}
\label{tab:runtime}
\end{table}

In Tab.~\ref{tab:runtime}, we report the hardware setup, throughput, and peak GPU memory for each baseline setting.

\clearpage
\section{CORAL on Other Baselines}
\label{supple:coral_across_pose_injection_strategies}
CORAL requires no architectural change, no additional module, and no change at inference. It can be attached to any DiT architecture that exposes a person-garment attention map for supervision, and the model is then fine-tuned with the loss. To demonstrate this, we extend our application to additional DiT baselines.

\subsection{Baselines with Different Pose Injection Strategies}
\label{supple:Baselines with Different Pose Injection Strategies}
\begin{figure}[!h]
  \centering
  \includegraphics[width=1\linewidth]{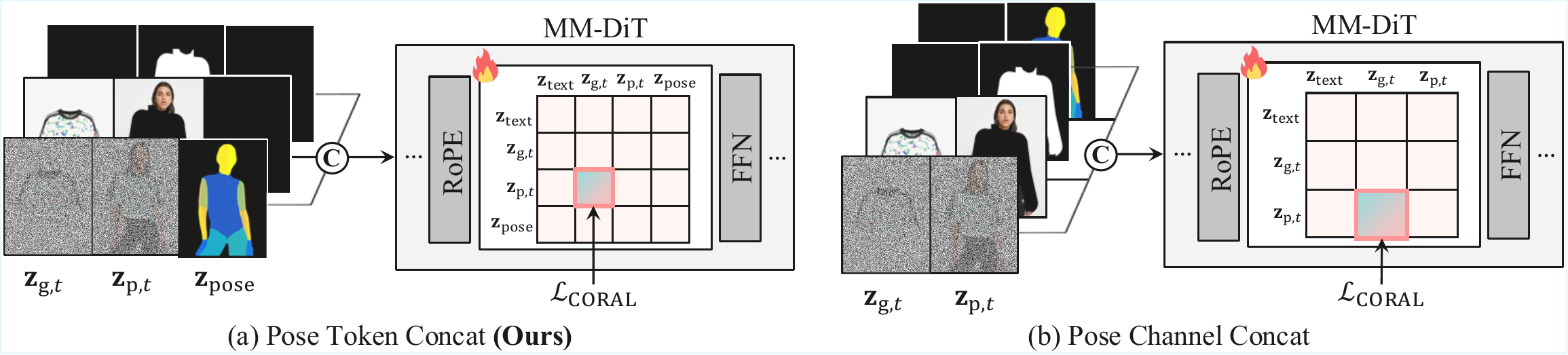}
  \caption{\textbf{Architecture Comparison.}}
  \label{fig:supple_baseline_arch}
\end{figure}

\paragraph{Pose Token Concat.} 
To leverage the DiT's token-based processing over the full sequence, we consider a baseline that injects pose as additional tokens. As shown in Fig.~\ref{fig:supple_baseline_arch} (a), we concatenate the pose condition  $\mathbf{z}_\text{pose}$ along the token dimension. This preserves positional structure in the token space and provides explicit pose correspondences to the model. Our main experiments use this \textbf{Pose Token Concat.} baseline. For more details, see Sec.~\ref{DIT-based VTON Baseline} of the main paper and Sec.~\ref{sec:implementation_details} of Appendix.

\paragraph{Pose Channel Concat.}
\label{supple:pose_channel_concat}
As illustrated in Fig.~\ref{fig:supple_baseline_arch} (b), we additionally design a  baseline that injects the pose condition $\mathbf{z}_\text{pose}$ along the channel dimension rather than the token dimension. Specifically, we first form a pose diptych $\mathbf{z}_\text{pose-diptych}$ by placing $\mathbf{z}_\text{pose}$ next to a zero-padded panel to match the diptych layout. We then concatenate $\mathbf{z}_t$, $\mathbf{z}_{\text{diptych}}$, $\mathbf{m}_{\text{diptych}}$, and the $\mathbf{z}_\text{pose-diptych}$ channel-wise, and feed the resulting tokens to the input projection layer within \texttt{FLUX.1-Fill-dev}. This increases the projection input dimension from 384 to 448, so we expand the projection layer and zero-initialize the newly added weights. Copying the original projection weights into the expanded channels led to degraded performance and unstable convergence.

\begin{table}[!h]
\vspace{-5pt}
\resizebox{1\linewidth}{!}{
\centering
\small
\begin{tabular}{cccccccc}
\toprule
\multirow{3}{*}{\textbf{Methods}}& \multirow{3}{*}{$\mathcal{L}_{\text{CORAL}}$} & \multicolumn{4}{c}{Paired} & \multicolumn{2}{c}{Unpaired} \\
\cmidrule(lr){3-6} \cmidrule(lr){7-8}
& & \mc{SSIM \up} & \mc{LPIPS \down} & \mc{FID \down} & \mc{KID \down} & \mc{FID \down} & \mc{KID \down} \\
\midrule
\multirow{2}{*}{Pose Token Concat} & \textcolor{gray}{\ding{55}} & 0.889 & 0.055 & 5.543 & 0.870 & 9.641 & 1.323 \\
& \ding{51} & \textbf{0.907} (\textcolor{teal}{+2.02\%}) & \textbf{0.048} (\textcolor{teal}{+12.73\%}) & \textbf{4.962} (\textcolor{teal}{+10.48\%}) & \textbf{0.565} (\textcolor{teal}{+35.06\%}) & \textbf{8.763} (\textcolor{teal}{+9.11\%}) & \textbf{0.880} (\textcolor{teal}{+33.48\%}) \\
\multirow{2}{*}{Pose Channel Concat} & \textcolor{gray}{\ding{55}} & 0.886 & 0.060 & 5.554 & 0.540 & 9.841 & 1.221 \\
& \ding{51} & 0.888 (\textcolor{teal}{+0.23\%}) & 0.059 (\textcolor{teal}{+1.67\%}) & 5.372 (\textcolor{teal}{+3.28\%}) & 0.744 & 9.323 (\textcolor{teal}{+5.26\%}) & 1.077 (\textcolor{teal}{+11.79\%}) \\
\bottomrule
\end{tabular}%
}
\vspace{+5pt}
\captionof{table}{\textbf{Effect of $\mathcal{L}_\text{CORAL}$ across Baselines with Different Pose Injection Strategies for VITON-HD~\cite{choi2021vitonhdhighresolutionvirtualtryon}.}}
\label{tab:channel_cat_tab}
\vspace{-20pt}
\end{table}

\begin{figure}[!h]
    \centering
    \includegraphics[width=1\linewidth]{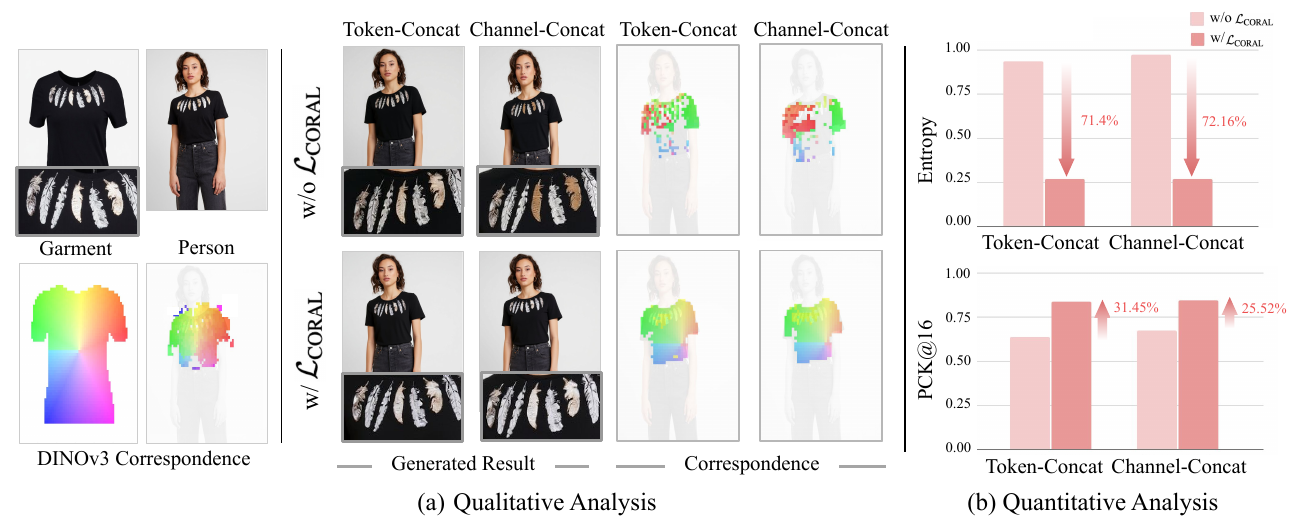} 
    \caption{\textbf{Effect of $\mathcal{L}_\text{CORAL}$ across Baselines with Different Pose Injection Strategies.}}
    \vspace{-15pt}
    \label{fig:Additional Baseline}
\end{figure}

\paragraph{Effect of $\mathcal{L}_\text{CORAL}$.} We evaluate CORAL on two pose-injection baselines, \textbf{Pose Token Concat.} and \textbf{Pose Channel Concat.}, with and without $\mathcal{L}_\text{CORAL}$. As reported in Tab.~\ref{tab:channel_cat_tab}, adding $\mathcal{L}_\text{CORAL}$ consistently improves VTON performance for both baselines, demonstrating that CORAL is model-agnostic across DiT variants that expose a person$\rightarrow$garment attention. Between the two baselines, Pose Token Concat performs notably better than Pose Channel Concat. This is expected because token concatenation preserves spatially aligned pose tokens that the DiT can attend to directly, whereas channel concatenation injects pose more indirectly through the input projection. Fig.~\ref{fig:Additional Baseline} (a) further shows that $\mathcal{L}_\text{CORAL}$ refines person$\rightarrow$garment mismatches in attention, leading to more accurate garment transfer: for instance, the fourth and fifth feather patterns no longer bleed into nearby brown tones and remain clean white as in the garment reference. Fig.~\ref{fig:Additional Baseline} (b) quantitatively supports that across both baselines, $\mathcal{L}_\text{CORAL}$ reduces attention entropy and increases PCK, indicating sharper and more correct correspondences. 

\subsection{An Additional Efficient Baseline}
\label{supple:coral_on_efficient_baseline}
We introduce an additional efficient baseline trained under a different setup to test whether CORAL’s improvements remain consistent beyond the settings in Sec.~\ref{supple:coral_across_pose_injection_strategies}.

\begin{wrapfigure}{r}{0.48\textwidth}
  \centering
  \vspace{-25pt}
  \includegraphics[width=1\linewidth]{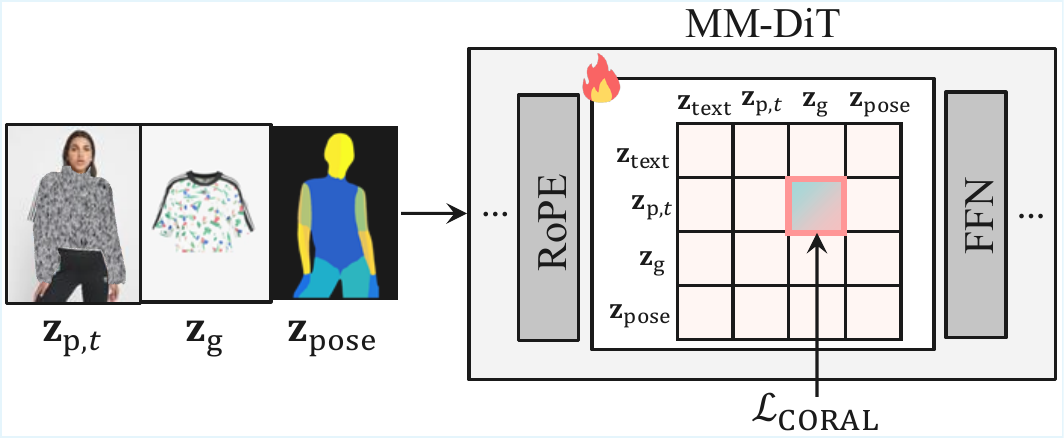}
  \caption{\textbf{Efficient Baseline Architecture.}}
  \label{fig:supple_baseline_omini_arch}
\end{wrapfigure}

\paragraph{Architectural Details.} Additional baseline is designed to be computationally efficient. Following the compact token integration method in prior work~\cite{tan2025ominicontrol2efficientconditioningdiffusion}, we first restrict denoising processes to the garment region in $\mathbf{z}_{\text{p}}$ using downsampled mask $\mathbf{m}_\text{e}$, since denoising the full person latent adds unnecessary computation and can also perturb garment-unrelated areas. At diffusion timestep $t$, the noisy person latent is defined as:
\begin{align}
\mathbf{z}_{\text{p}, t} &= \bigg((1-t)\mathbf{z}_\text{p} + t\mathbf{\epsilon} \bigg) \odot \mathbf{m}_\text{p} + \mathbf{z}_{\text{p}}  \odot (1-\mathbf{m}_\text{p}).
\end{align}
We also keep the garment latent clean to avoid unnecessary denoising process on the garment input. For pose conditioning, we concatenate the pose condition $\mathbf{z}_{\text{pose}}$ along the token dimension, as discussed in Sec.~\ref{sec:pose_injection}. 

\textbf{Implementation Details.} We initialize our model from \texttt{FLUX.1-dev} and finetune with LoRA~\cite{hu2021loralowrankadaptationlarge} with rank $128$. We train with Prodigy~\cite{mishchenko2024prodigyexpeditiouslyadaptiveparameterfree} optimizer using a learning rate of 1.0 and batch size of 1 on 2 NVIDIA A100 GPUs. We train and report results at a resolution of 1024 x 1024. For training with $\mathcal{L}_{\text{CORAL}}$, same learning rate and batch size are used, with loss weights $\lambda_{\text{corr}} = 0.01$ and $\lambda_{\text{ent}} = 0.1$. All other implementation details follow Sec.~\ref{sec:implementation_details} of Appendix.

\newcommand{\cmark}{\ding{51}}
\newcommand{\xmark}{\ding{55}}

\begin{table}[!h]
\vspace{-5pt}
\centering
\resizebox{1\linewidth}{!}{%
\small
\begin{tabular}{c *{12}{c}}
\toprule
\multirow{3}{*}{$\mathcal{L}_{\text{CORAL}}$}
& \multicolumn{6}{c}{VITON-HD} & \multicolumn{6}{c}{DressCode} \\
\cmidrule(lr){2-7} \cmidrule(lr){8-13}
& \multicolumn{4}{c}{Paired} & \multicolumn{2}{c}{Unpaired}
& \multicolumn{4}{c}{Paired} & \multicolumn{2}{c}{Unpaired} \\
\cmidrule(lr){2-5} \cmidrule(lr){6-7} \cmidrule(lr){8-11} \cmidrule(lr){12-13}
& \mc{SSIM \up} & \mc{LPIPS \down} & \mc{FID \down} & \mc{KID \down} & \mc{FID \down} & \mc{KID \down}
& \mc{SSIM \up} & \mc{LPIPS \down} & \mc{FID \down} & \mc{KID \down} & \mc{FID \down} & \mc{KID \down} \\
\midrule
\textcolor{gray}{\xmark} & 0.864 & 0.084 & 9.521 & 1.353 & 11.251 & 0.456
& 0.905 & 0.065 & 5.752 & 0.473 & 8.723 & 0.736 \\
\cmark & \textbf{0.871} (\textcolor{teal}{+0.81\%}) & 0.084 & \textbf{6.271} (\textcolor{teal}{+34.1\%}) & \textbf{0.563} (\textcolor{teal}{+58.5\%}) & \textbf{10.285} (\textcolor{teal}{+8.62\%}) & \textbf{0.315} (\textcolor{teal}{+30.9\%})
& 0.905 & \textbf{0.061} (\textcolor{teal}{+6.15\%}) & \textbf{5.221} (\textcolor{teal}{+9.22\%}) & 0.581 & \textbf{8.70} (\textcolor{teal}{+0.23\%}) & \textbf{0.193} (\textcolor{teal}{+73.78\%}) \\
\bottomrule
\end{tabular}%
}
\vspace{+5pt}
\label{tab:efficient_quan}
\captionof{table}{\textbf{Effect of $\mathcal{L}_\text{CORAL}$ on Additional Efficient Baseline for VITON-HD~\cite{choi2021vitonhdhighresolutionvirtualtryon} and DressCode~\cite{morelli2022dresscodehighresolutionmulticategory}}.}
\vspace{-20pt}
\end{table}

\paragraph{Effect of $\mathcal{L}_\text{CORAL}$.} Applying CORAL improves the efficient baseline across overall metrics, with particularly large improvements compared to our main baseline. In this efficient training setup with LoRA finetuning, the baseline performance is relatively weaker, yet CORAL still yields substantial gains. 
The largest improvements appear in distribution-based metrics, which indicates that CORAL also enhances overall realism of generated outputs, not only the perceptual fidelity. \\
Overall, the improvements across baselines confirm the effectiveness of $\mathcal{L}_\text{CORAL}$ and highlight its suitability for DiTs, where attention provides an explicit controllable correspondence mechanism.

\clearpage
\subsection{Baseline with a Different Backbone}
\label{sec:flux2_result}
\begin{wraptable}{r}{0.55\linewidth}
\vspace{-12pt}
\centering
\resizebox{\linewidth}{!}{%
\begin{tabular}{c *{6}{c}}
\toprule
\multirow{2}{*}{$\mathcal{L}_{\text{CORAL}}$}
& \multicolumn{4}{c}{Paired} & \multicolumn{2}{c}{Unpaired} \\
\cmidrule(lr){2-5} \cmidrule(lr){6-7}
& \mc{SSIM \up} & \mc{LPIPS \down} & \mc{FID \down} & \mc{KID \down} & \mc{FID \down} & \mc{KID \down} \\
\midrule
\textcolor{gray}{\xmark} & 0.894 & 0.053 & 5.433 & 0.529 & 9.682 & 0.860 \\
\cmark & \textbf{0.907} & \textbf{0.047} & \textbf{4.941} & \textbf{0.382} & \textbf{9.018} & \textbf{0.674} \\
 & \textcolor{teal}{+1.45\%} & \textcolor{teal}{+11.3\%} & \textcolor{teal}{+9.06\%} & \textcolor{teal}{+27.8\%} & \textcolor{teal}{+6.86\%} & \textcolor{teal}{+21.6\%} \\
\bottomrule
\end{tabular}%
}
\caption{\textbf{Effect of $\mathcal{L}_\text{CORAL}$ on a FLUX.2 Baseline for VITON-HD~\cite{choi2021vitonhdhighresolutionvirtualtryon}}.}
\label{tab:flux2}
\vspace{-10pt}
\end{wraptable}
To examine whether the gain generalizes to other backbones, we further attach
$\mathcal{L}_{\text{CORAL}}$ to a try-on model built on FLUX.2, a more recent
diffusion transformer. As shown in Tab.~\ref{tab:flux2}, CORAL improves every
metric in both the paired and unpaired settings.

\clearpage
\section{VLM-based Evaluation}
\label{supple:vlm_eval}
\begin{figure*}[!h]
    \includegraphics[width=\linewidth,]{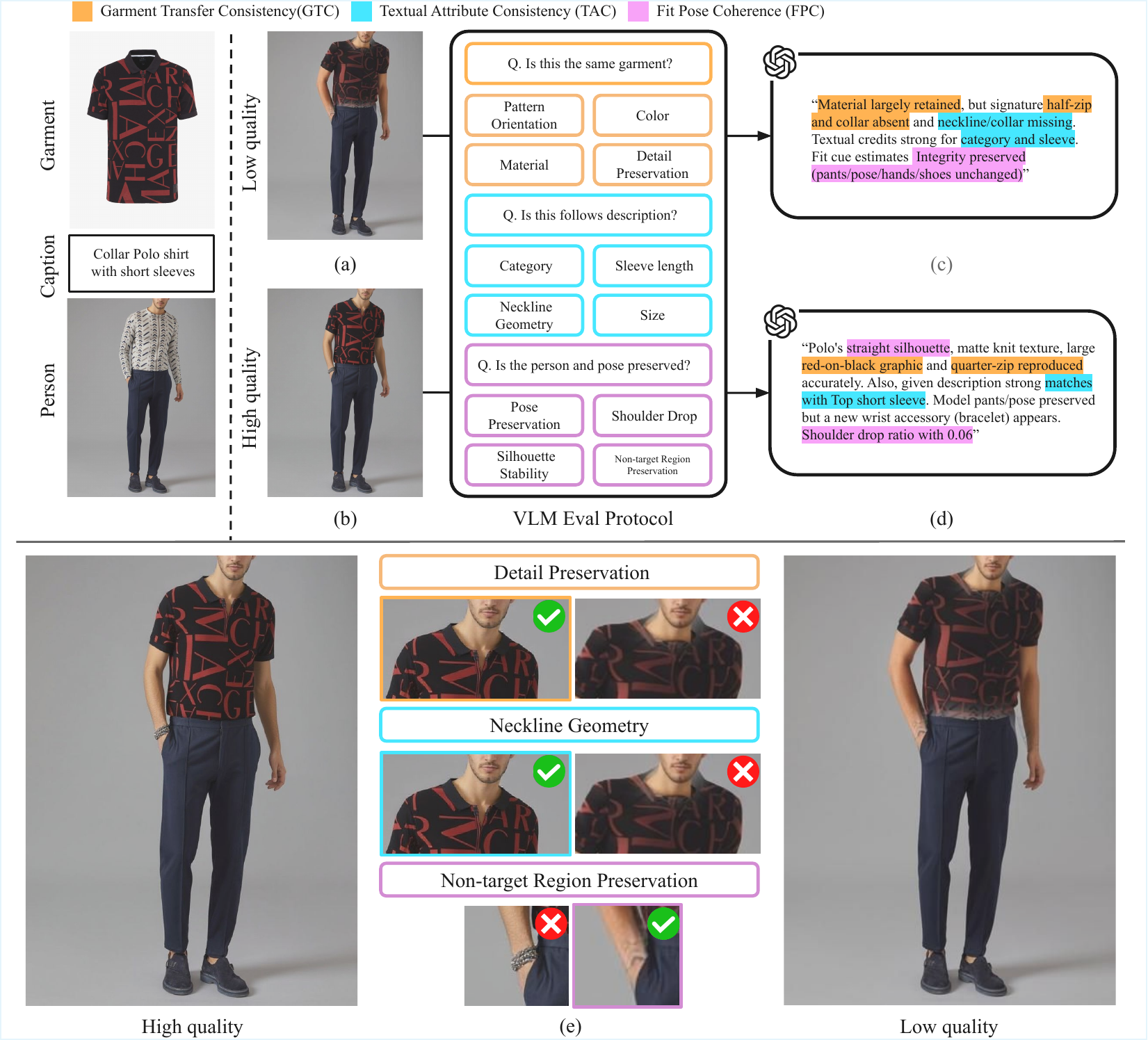}
    \caption{\textbf{VLM-based Evaluation Results Comparison.}}
\label{fig:vlm-eval-result-comparison}
\end{figure*}

\paragraph{Why VLM-based Evaluation is Needed?}
As discussed in Sec.~\ref{sec:exp-vlm-eval}, standard metrics such as FID~\cite{soloveitchik2022conditionalfrechetinceptiondistance}, KID~\cite{bińkowski2021demystifyingmmdgans}, SSIM~\cite{cong2022imagequalityassessmentgradient} and LPIPS~\cite{zhang2018unreasonableeffectivenessdeepfeatures} are insufficient for evaluating VTON, as they mainly reflect overall realism or low-level similarity and do not assess whether the specific reference garment is faithfully transferred and plausibly worn. Following Sec.~\ref{sec:exp-vlm-eval} of the main paper, we therefore propose a VLM-based evaluation protocol with three criteria, including Garment Transfer Consistency (GTC), Textual Attribute Consistency (TAC), and Fit Pose Coherence (FPC). For all VLM-based evaluations, we use GPT-4o as our evaluator. The full input prompt used for VLM evaluation is provided in Fig.~\ref{fig:vlm_eval}.

Fig.~\ref{fig:vlm-eval-result-comparison} motivates the need for this protocol. Although the two outputs in Fig.~\ref{fig:vlm-eval-result-comparison} (a) and (b) are generated from the same input garment and person, they differ in whether garment details, attributes and fit-pose are preserved. Our evaluator produces GTC, TAC, and FPC scores together with natural language justifications that explain which visual evidence supports each score, as shown in Fig.~\ref{fig:vlm-eval-result-comparison} (c) and (d). (e) provides the detailed view of the generated images aligned with each criterion. This design is motivated by recent findings~\cite{cheng2025evaluatingmllmsmultimodalmultiimage, jia2025exploringevaluatingmultimodalknowledge, chen2025reasoningerasurveylong} that VLMs~\cite{dai2023instructblipgeneralpurposevisionlanguagemodels, li2024llavanextinterleavetacklingmultiimagevideo} excel at image and multi-image reasoning, making them well suited to judge garment-specific fidelity beyond distribution-level similarity. 

\paragraph{VLM Evaluation Setup.}
For all VLM-based evaluations, we use GPT-4o as our 
evaluator. For each generated try-on result, the VLM 
receives 4 inputs: the input person image, the 
reference garment image, the generated output, and a 
caption describing the garment attributes (extracted 
from the reference garment using 
GPT-4o~\cite{openai2024gpt4technicalreport}). The VLM 
is prompted to produce four sub-scores covering 
\textit{textual alignment} (TAC), \textit{visual match} 
(GTC), \textit{fit quality}, and \textit{non-target 
integrity}, the latter two combined into FPC. Each 
sub-score is reported on both a continuous (0.0-5.0) 
and discrete (1-5) Likert scale, accompanied by a brief 
JSON-formatted justification. To prevent overrating in 
challenging cases, the prompt enforces hard score caps 
when severe failures are detected (e.g., category 
mismatch, silhouette mismatch). The full prompt 
specification is provided in Fig.~\ref{fig:vlm_eval}.

\paragraph{Definition of Each Criterion.}
We evaluate three aspects that existing metrics 
inadequately reflect:

\textit{(1) Garment Transfer Consistency (\textbf{GTC})} measures 
whether the output preserves the garment's visual 
appearance, including its overall shape and silhouette 
as well as details like patterns, logos, and texture.

\textit{(2) Textual Attribute Consistency (\textbf{TAC})} captures 
whether the output satisfies garment attributes specified 
in text or metadata, covering requirements that may be 
ambiguous from the reference image alone (e.g., length, 
neckline, category, and fit).

\textit{(3) Fit Pose Coherence (\textbf{FPC})} reflects whether 
the garment is worn naturally on the target person, 
consistent with the input person pose and body geometry. 
It also considers whether non-target regions (e.g., hair, 
skin, hands, and other clothing) are preserved without 
unintended changes.

\paragraph{Reliability of VLM-based Evaluation.}
The VLM-based rankings are broadly consistent with user 
study results (Sec.~\ref{supple:human_eval}). To provide 
more rigorous validation, we have GPT-4o score the same 
45 image-question pairs from our user study using the 
prompt described in Fig.~\ref{fig:vlm_eval}, with 
sub-scores aggregated into the same three criteria 
(GTC, TAC, FPC) used in the user study. From the 2,430 
total human ratings (54 participants $\times$ 45 
questions), we compute the average human score per 
question and measure Pearson correlation with GPT-4o 
scores across the 45 averaged data points, yielding 
$r=0.71$ ($p<0.001$, $n=45$). This represents 
a large effect size with GPT-4o explaining approximately 
50$\%$ of the variance in human scores ($r^2 \approx 0.50$), 
supporting the reliability of our VLM-based protocol as 
a proxy for human evaluation.

\clearpage
\section{Human Evaluation}
\label{supple:human_eval}
\begin{wrapfigure}{r}{0.5\textwidth} 
    \centering
    \includegraphics[width=1\linewidth]{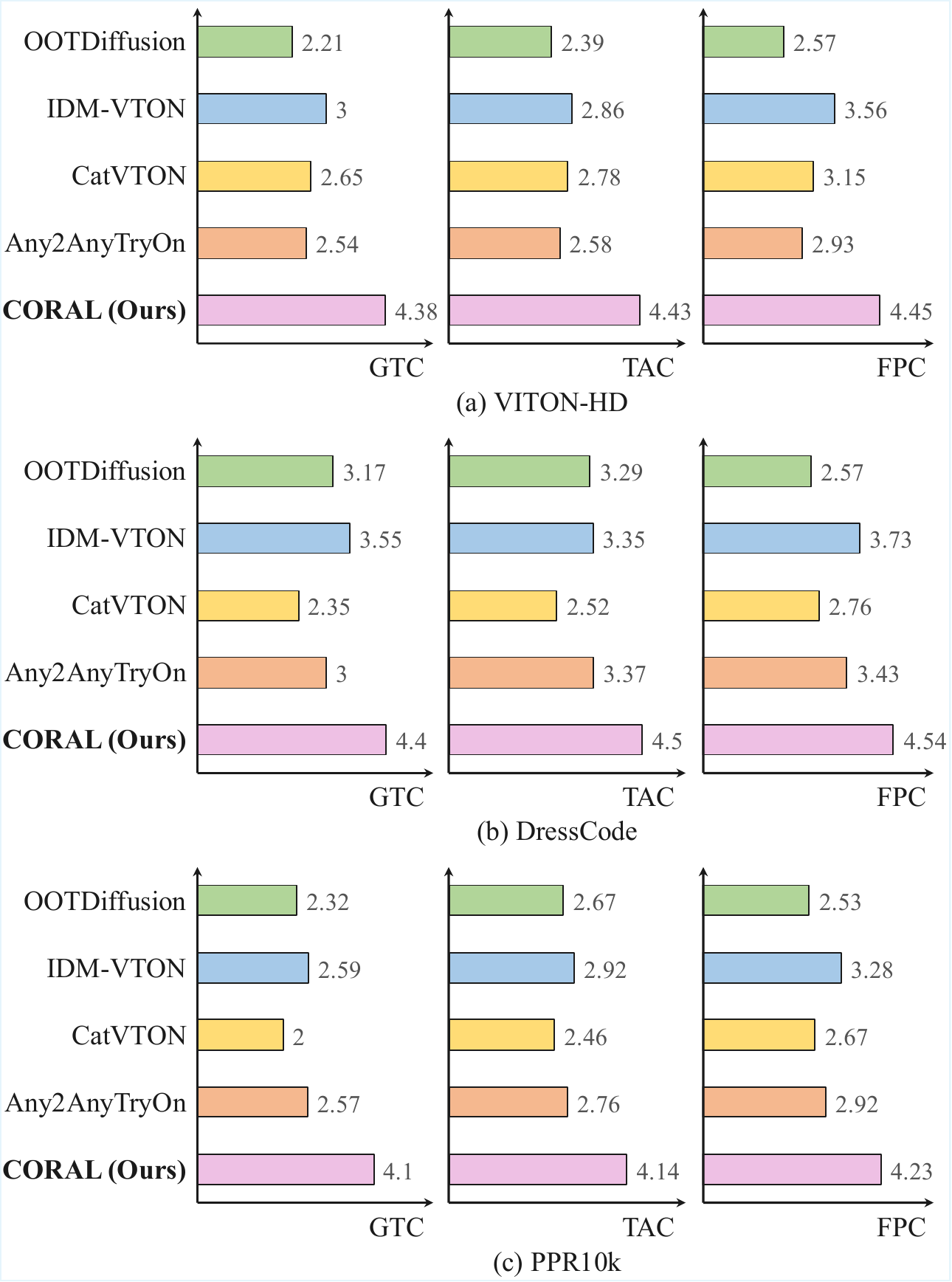}
    \vspace{-10pt}
    \caption{\textbf{User Study Results.}}
    \vspace{-10pt}
    \label{fig:user_study_results}
\end{wrapfigure}

\paragraph{Human Evaluation Details.}
We conducted a user study to measure virtual try-on quality under human perception and to validate that our VLM-based evaluation reflects human preference. Participants rated each generated result on the same three criteria used in Sec.~\ref{sec:exp-vlm-eval}, including Garment Transfer Consistency (GTC), Textual Attribute Consistency (TAC), and Fit-Pose Coherence (FPC), using a 5-point scale (from 1 to 5) for each criterion. Fig.~\ref{fig:user_study_prompt} presents example questions from the user study.

To make the criteria concrete and reduce ambiguity, we provided references for each question. For GTC, we showed the input garment image so participants could directly compare whether shape, patterns, and logos were preserved. For TAC, we provided a short textual description of the garment attributes, automatically extracted from the reference garment using a pretrained vision-language model~\cite{openai2024gpt4technicalreport}. For FPC, we showed the input person image so participants could judge whether the garment is worn naturally and follows the target pose and body geometry without altering non-target regions. 

A total of $54$ participants answered $45$ questions. For a fair comparison, we sampled outputs from a large pool sharing the same input person image and garment image across $5$ comparison models, including OOTDiffusion~\cite{xu2024ootdiffusionoutfittingfusionbased}, IDM-VTON~\cite{choi2024improvingdiffusionmodelsauthentic}, CatVTON~\cite{chong2025catvtonconcatenationneedvirtual}, Any2AnyTryOn~\cite{guo2025any2anytryonleveragingadaptiveposition} and CORAL (Ours) to ensure intra-rater reliability. 

\paragraph{Human Evaluation Results.}
Fig.~\ref{fig:user_study_results} shows the results, demonstrating that our model outperforms others across all criteria.

\paragraph{Correlation with VLM-Evaluation.}
We find that our VLM-based scores align well with the user study outcomes, indicating that the proposed VLM evaluation reflects human preference more faithfully than standard metrics. Across Tab.~\ref{tab:vlm_quan_all} and the user study, methods that receive higher GTC/TAC/FPC scores also tend to be preferred by human participants, and CORAL performs strongly in both evaluations. This agreement suggests that GTC, TAC, and FPC capture perceptually important failure cases in try-on quality.

\clearpage
\section{Evaluation on Additional Datasets}
\label{supple:additional_evaluation}
\subsection{Motivation} We evaluate whether the correspondences learned by CORAL generalize beyond standard VTON benchmarks and still leads to high-quality try-on results under input conditions that are less curated than the benchmark setting. For the person input, we consider photos with broader variation in poses, viewpoints, backgrounds, framing, and occlusions than typical benchmarks. For the garment input, we consider reference images that are not limited to clean product images.
\label{supple:additional_evaluation_motivation}

Most existing VTON benchmarks follow the in-shop garment transfer setting, where the reference garment is a flat product image and the target person is photographed in a controlled studio environment. However, this setting does not fully reflect real-world inputs. User-provided person photos are often taken in everyday environments with diverse framing, non-frontal viewpoints, more dynamic poses, and frequent occlusions by accessories or surrounding objects. Garment reference images can also differ substantially in practice. It is often inconvenient for users to search for and provide a clean product photo for every garment. Instead, they typically rely on readily available photos, such as a casually taken garment photo with cluttered backgrounds or a worn-garment image from another person, which introduces additional noises compared to product images.

For this reason, we additionally build paired and unpaired test sets to evaluate CORAL under more challenging conditions while keeping the evaluation reliable. We evaluate with worn-garment references, where the reference garment is worn by another person, which is typically more challenging than a clean product image but still provides sufficient garment cues for try-on.

\begin{figure*}[!h]
    \includegraphics[width=\linewidth,]{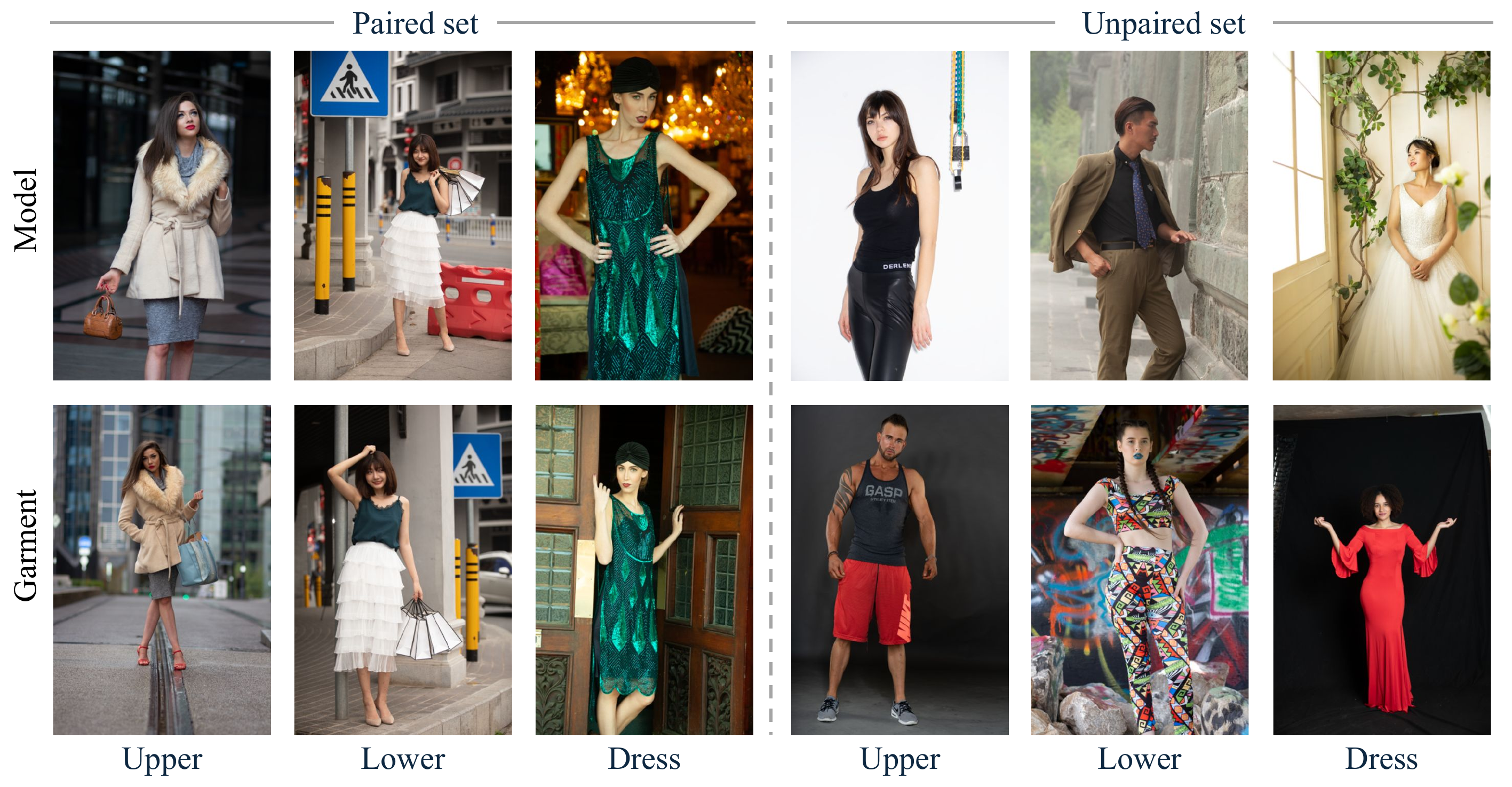}
    \vspace{-10pt}
    \caption{\textbf{Additional Evaluation Dataset Samples.}}
\label{fig:dataset_cmoparison}
\end{figure*}

\label{supple:add_eval_details}
\subsection{Evaluation Dataset Details}
We additionally build paired and unpaired evaluation sets using PPR10K~\cite{liang2021ppr10klargescaleportraitphoto}, since it offers high-resolution full-body portraits captured under diverse real-world conditions and provides grouped images of the same subject, which enables stable paired evaluation with consistent garment identity.

\paragraph{Dataset Overview.} PPR10K is a large scale, high quality portrait photo retouching dataset designed to cover diverse real world conditions. It contains 11,161 high resolution raw portrait photos organized into 1,681 groups, where each group includes 3 to 18 photos of the same subject captured in the same scene within a short time span. The dataset was curated to cover diverse shooting purposes such as weddings, graduations, and advertisements. It also includes diverse subjects (e.g., children and people from diverse regions), varied indoor and outdoor backgrounds, and a wide range of lighting conditions from day to night and across seasons. The portraits cover diverse framing from face-centric to full-body shots, and include various viewpoints excluding fully back-facing.

\paragraph{Dataset Preprocessing.} We first filter person images to keep upper-body to full-body shots. We remove face-centric crops by extracting 2D keypoints with OpenPose~\cite{8765346} and retaining images where the body is visible at least down to the knees. We further heuristically filter samples by viewpoints using IUV maps extracted from DensePose~\cite{güler2018denseposedensehumanpose} to exclude fully back and side views. For the reference garment setting, we exclude images where the garment region is too small to provide meaningful detail, as well as extreme cases such as overly dark or overexposed images and heavy occlusions. We use the same preprocessing pipeline as in the standard VTON benchmarks for extracting DensePose conditions. We obtain input masks following prior work~\cite{choi2024improvingdiffusionmodelsauthentic}. After preprocessing, the final split contains 358 dresses, 320 upper-body samples, and 195 lower-body samples. In the paired setting there are 309 samples (138 dresses, 86 upper-body, 85 lower-body) and in the unpaired setting there are 564 samples (220 dresses, 234 upper-body, 110 lower-body).

\section{Person-To-Person Garment Transfer}
\label{supple:applications}

\begin{wrapfigure}{r}{0.55\textwidth}
  \centering
  \vspace{-10pt}
  \includegraphics[width=\linewidth]{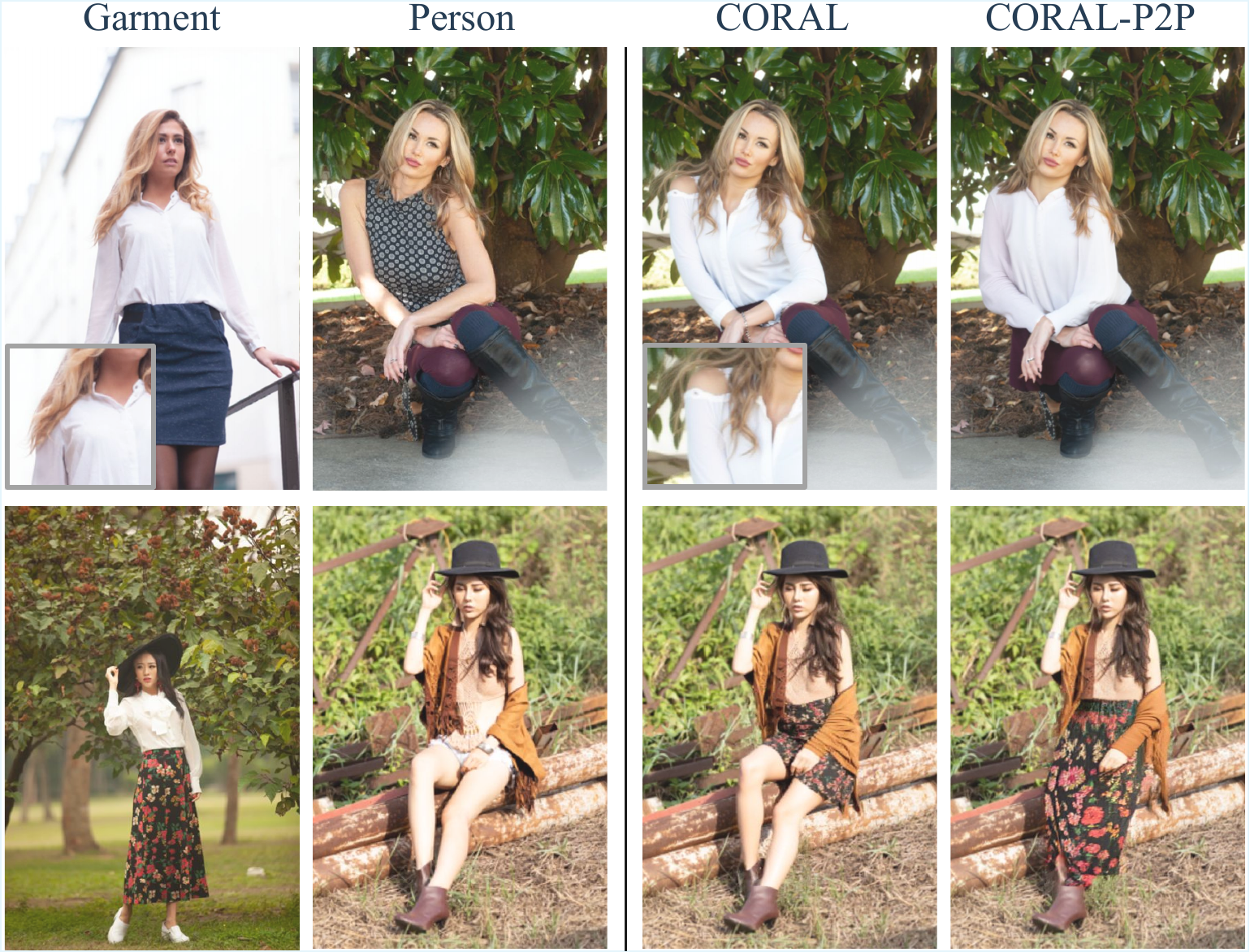}
  \caption{\textbf{Person-To-Person Garment Transfer Motivation.}}
  \label{fig:supple_p2p_motivation}
\end{wrapfigure}

\paragraph{Motivation.} As mentioned in Sec.~\ref{supple:additional_evaluation_motivation}, worn-garment transfer is a practically important setting where the garment is observed on another clothed person rather than as a clean product photo. We refer to this setup as \emph{person-to-person} garment transfer. Compared to in-shop product photos, worn-garment references are more common in real world and also show how the garment looks when worn, but it is harder to determine what should be transferred from in-the-wild reference images. A standard in-shop setting often addresses this by estimating a garment mask from the reference image of a clothed person, but this mask can be unreliable under heavy occlusions or complex backgrounds. As illustrated in the first row of Fig.~\ref{fig:supple_p2p_motivation}, occlusion by hair can remove garment regions in the mask, and the try-on result can transfer this missing shape, making the garment look cut out in the same area. Garment masks can also remove global shape information such as overall length and silhouette, which is particularly critical for lower-garment transfer. As shown in Fig.~\ref{fig:supple_p2p_motivation}, for skirts and dresses, a garment-only reference cropped by a mask can hide where the garment ends on the body, making it difficult to preserve the correct length. For this reason, we train a separate model for \emph{person-to-person} garment transfer using full reference images without estimating garment masks, and apply CORAL to improve person–garment correspondence under these in-the-wild references.

\paragraph{Architectural Details.} In the \emph{person-to-person} garment transfer setting, the garment reference image is provided through another clothed person image rather than a clean product image. We therefore denote the target and reference person images as $I_{\text{p}_{\text{tgt}}}$ and $I_{\text{p}_{\text{ref}}}$, and encode them with a VAE to obtain the corresponding latents $\mathbf{z}_{\text{p}_{\text{tgt}}}$ and $\mathbf{z}_{\text{p}_{\text{ref}}}$. Following the same diptych layout as in Sec.~\ref{sec:diptcy_formulation}, at diffusion timestep $t$, we construct the noisy diptych latent as:
\begin{align}
\mathbf{z}_t = \bigl[\, \mathbf{z}_{\text{p}_{\text{ref}},t} \,\Vert\, \mathbf{z}_{\text{p}_{\text{tgt}},t} \,\bigr].
\end{align}
For conditioning worn-garment reference image and the masked person image, we use the clean reference person latent and the masked target person latent as:
\begin{align}
\mathbf{z}_{\text{diptych}} = \bigl[\, \mathbf{z}_{\text{p}_{\text{ref}}} \,\Vert\, \mathbf{z}_{\text{p}_{\text{tgt}}} \odot (1-\mathbf{m}_\text{e}) \,\bigr],
\end{align}
where $\mathbf{m}_\text{e}$ is obtained by downsampling $M_e$ into the latent space. We use the same definition of the binary mask canvas $\mathbf{m}_{\text{diptych}}$ as in Sec.~\ref{sec:diptcy_formulation}.

Since DensePose~\citep{güler2018denseposedensehumanpose} can be estimated from both $I_{\text{p}_{\text{tgt}}}$ and $I_{\text{p}_{\text{ref}}}$, we extract two pose conditions aligned to the target and reference persons, denoted as $I_{\text{pose}_{\text{tgt}}}$ and $I_{\text{pose}_{\text{ref}}}$, and then encode them with a VAE to obtain $\mathbf{z}_{\text{pose}_{\text{tgt}}}$ and $\mathbf{z}_{\text{pose}_{\text{ref}}}$. Due to additional pose inputs for the reference worn-garment image, pose injection through token concatenation would increase the number of tokens, leading to higher computation and memory cost. We therefore inject the pose condition along the channel dimension instead. As explained in Sec.~\ref{supple:pose_channel_concat} of Appendix, we build the diptych layout for pose conditions, $\mathbf{z}_{\text{pose-diptych}} = [\,\mathbf{z}_{\text{pose}_{\text{ref}}} \,\Vert\, \mathbf{z}_{\text{pose}_{\text{tgt}}}\,]$, and then concatenate $\mathbf{z}_t$, $\mathbf{z}_{\text{diptych}}$, $\mathbf{m}_{\text{diptych}}$, and the $\mathbf{z}_{\text{pose-diptych}}$ channel-wise before the input projection layer in \texttt{FLUX.1-Fill-dev}. 

\begin{figure}[!h]
  \centering
  \includegraphics[width=1\linewidth]{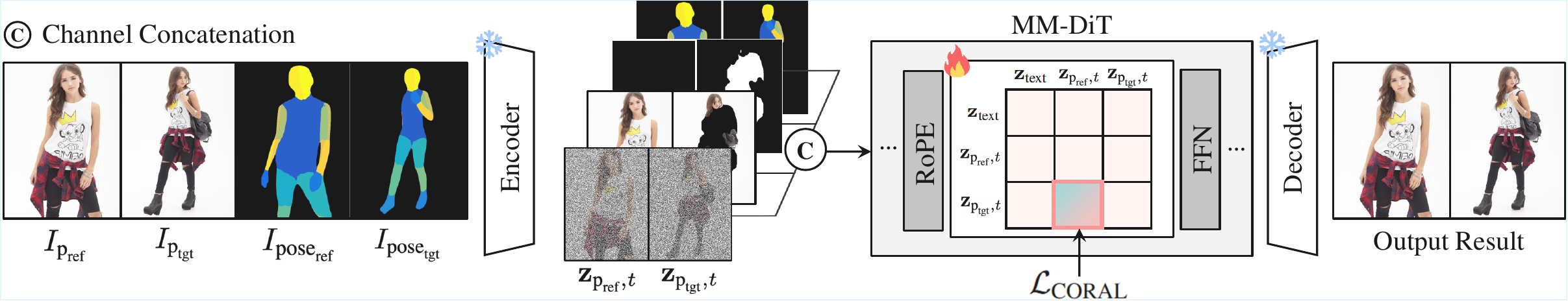}
  \caption{\textbf{Person-To-Person Architecture.}}
  \label{fig:supple_p2p_baseline}
\end{figure}

\paragraph{Training Dataset.} Existing VTON benchmarks are built on curated paired sets where the reference image is a flat in-shop product image, typically centered with a clean background. The worn-garment setting, however, does not come with an established benchmark. Even though some public paired datasets are available where multiple subject images share the same garment identity, some images have incorrect garment identity annotations, and there is no annotation indicating which image in each group should be used as the reference worn-garment image. In the worn-garment setting, many images are in-the-wild with different viewpoints, framing, and occasional occlusions, so some images may include garments that are not visible enough to be used as a reliable reference input.

To build paired data for \emph{person-to-person} garment transfer, we leverage DeepFashion~\cite{7780493} and an internal dataset. DeepFashion provides 14,120 pairs for the multi-view try-on task, and each pair typically includes multiple viewpoints across a range of garment categories, with reliable garment identity annotations. We filter out reference images with extreme viewpoints such as strong side or back views using DensePose~\cite{güler2018denseposedensehumanpose} IUV maps, and overly tight crops using human parsing maps~\cite{li2019selfcorrectionhumanparsing}. We also leverage an internal dataset covering tops, bottoms, and dresses, where images are paired by garment identity across different subjects, with garment identities annotated by human annotators. For in-the-wild images that are not full body, human parsing or OpenPose~\cite{8765346} can fail, so we additionally use SAM3~\cite{carion2025sam3segmentconcepts} to generate the input masks. In total, our final training set contains 23,678 paired samples.

\begin{wraptable}{r}{0.6\linewidth}
\vspace{-5pt}
\resizebox{1\linewidth}{!}{
\centering
\small
\begin{tabular}{l*{6}{c}}
\toprule
 \multicolumn{1}{c}{\multirow{2}{*}{\textbf{Methods}}}
& \multicolumn{4}{c}{Paired} & \multicolumn{2}{c}{Unpaired} \\
\cmidrule(lr){2-5} \cmidrule(lr){6-7} 
& \mc{SSIM \up} & \mc{LPIPS \down} & \mc{FID \down} & \mc{KID \down} & \mc{FID \down} & \mc{KID \down} \\
\midrule
\textbf{CORAL (w/o $\mathcal{L}_{\text{CORAL}}$)}  & 0.877  & 0.078 & 44.117 & 0.015 & 56.644 & 1.202 \\
\textbf{CORAL (w $\mathcal{L}_{\text{CORAL}})$}
& 0.915
& 0.060
& 43.648
&  0.011
&  53.164
&  1.101 \\
\midrule
\textbf{CORAL-P2P (w $\mathcal{L}_{\text{CORAL}})$}
& \textbf{0.923}
& \textbf{0.050}
&  \textbf{38.590}
&   \textbf{0.008}
&   \textbf{52.047} 
&  \textbf{1.1088}  \\
\bottomrule
\end{tabular}}
\captionof{table}{\textbf{Quantitative Comparison on PPR10K.}}
\label{tab:coral_p2p_tab}
\end{wraptable}

\paragraph{Quantitative Results.} As shown in Tab.~\ref{tab:coral_p2p_tab}, training the model for \emph{person-to-person} garment transfer with worn-garment references achieves better performance than using a model trained with in-shop product photos without retraining. In Tab.~\ref{tab:coral_p2p_tab}, \textbf{CORAL-P2P} uses the unmasked reference image at inference time, while the other \textbf{CORAL} settings use a garment-only reference extracted with a garment mask. In this more challenging setting, $\mathcal{L}_{\text{CORAL}}$ helps the model focus on the intended garment and avoid transferring details from irrelevant parts of the reference image, which is reflected in the quantitative results of Tab.~\ref{tab:coral_p2p_tab} across both paired and unpaired settings.

\begin{figure*}[!h]
\centering
    \includegraphics[width=0.91\linewidth]{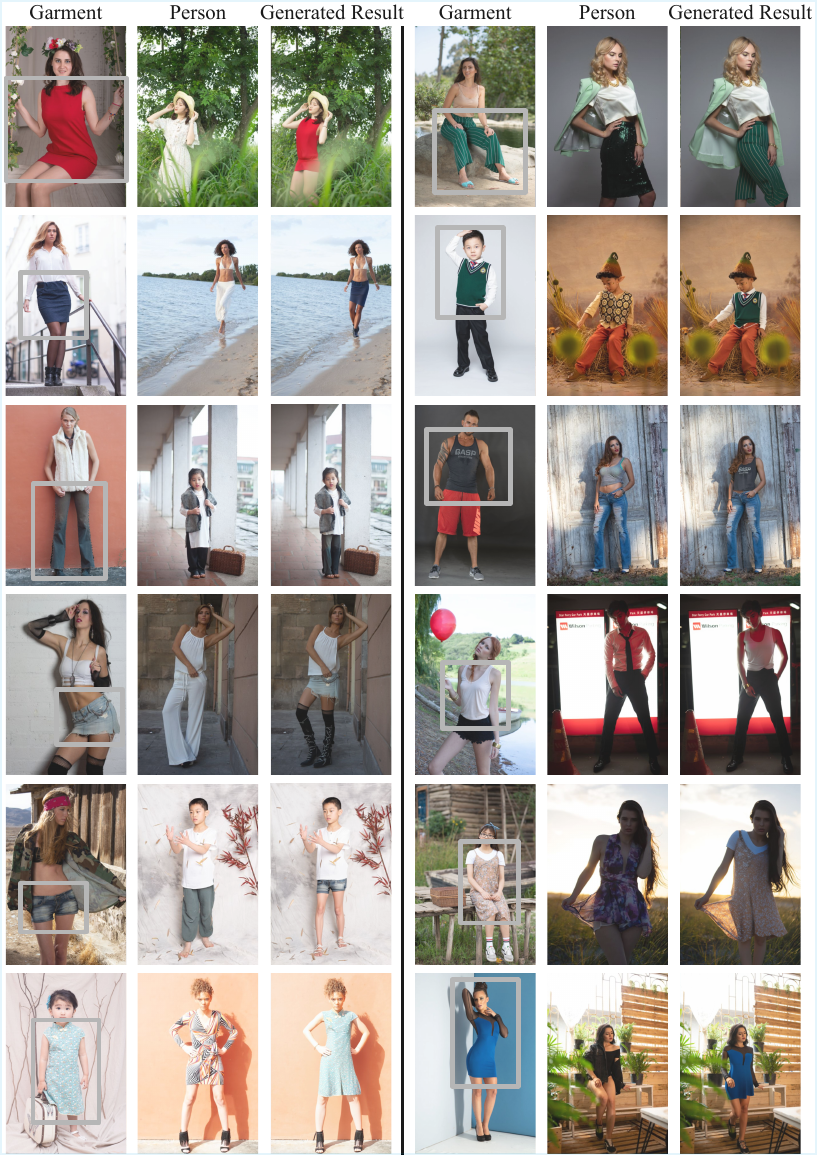}
    \caption{\textbf{Qualitative Results of \textbf{CORAL-P2P} on PPR10K.}}
\label{fig:p2p_additional_qual}
\end{figure*}

\paragraph{Qualitative Results.} We provide qualitative results for the \emph{person-to-person} garment transfer in Fig.~\ref{fig:supple_p2p_motivation} and Fig.~\ref{fig:p2p_additional_qual}. Models trained with in-shop product images often struggle when the garment reference is provided as a clothed person image rather than a clean product photo. As shown in Fig.~\ref{fig:supple_p2p_motivation}, the visible garment region in the reference image can be incomplete under occlusion, and the generated try-on result follows the edge of the visible part, making the garment look cut off or visually deformed. Training on \emph{person-to-person} pairs helps the model produce a more natural garment shape even under such occlusions, where the shoulder part in shirts is generated seamlessly with \textbf{CORAL-P2P}. For lower garments and dresses, garment length is hard to infer from a garment-only reference cropped by a mask, since it removes global information about where the garment ends on the body and its overall silhouette, which can lead to shortened hems or incorrect length, as shown in Fig.~\ref{fig:supple_p2p_motivation}. Instead, Fig.~\ref{fig:p2p_additional_qual} further shows that \textbf{CORAL-P2P} preserves overall length and silhouette by conditioning on the full reference person image, keeping the skirt length just above the ankles, as in the reference.

\clearpage
\section{Additional Qualitative Results}
\label{supple:additional_results}

\begin{figure*}[!h]
\centering
    \includegraphics[width=\linewidth]{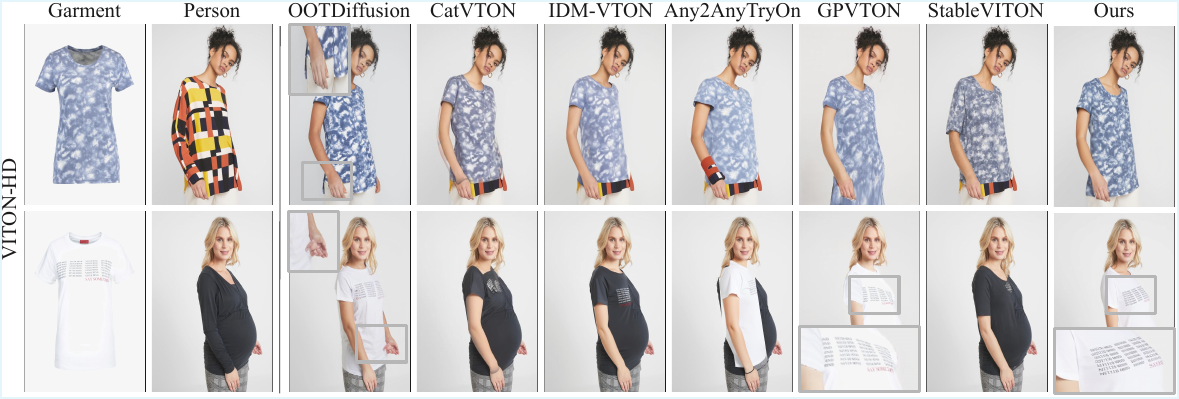}
    \caption{\textbf{Additional Qualitative Comparison of CORAL on VITON-HD~\cite{choi2021vitonhdhighresolutionvirtualtryon}}.}
\label{fig:additional_qual_vt}
\end{figure*}

\begin{figure*}[!h]
\centering
    \includegraphics[width=\linewidth]{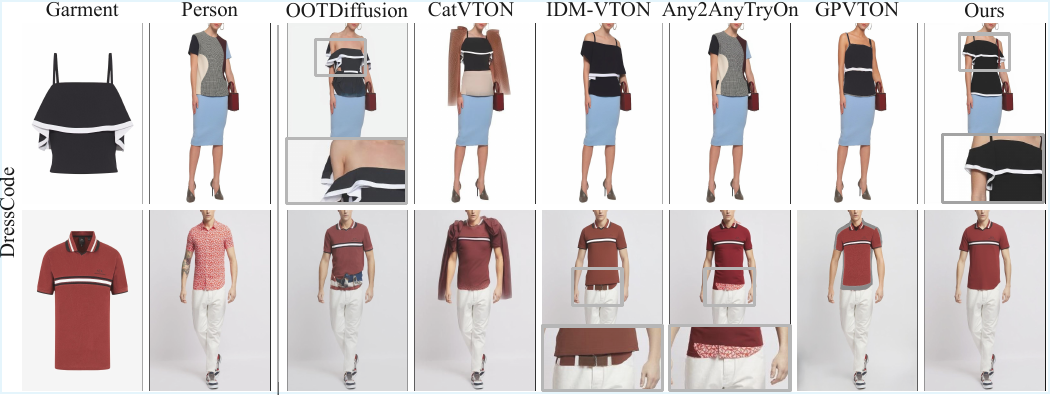}
    \caption{\textbf{Additional Qualitative Comparison of CORAL on DressCode~\cite{morelli2022dresscodehighresolutionmulticategory}.}}
\label{fig:additional_qual_dc}
\vspace{-15pt}
\end{figure*}

\begin{figure*}[!h]
\centering
    \includegraphics[width=\linewidth]{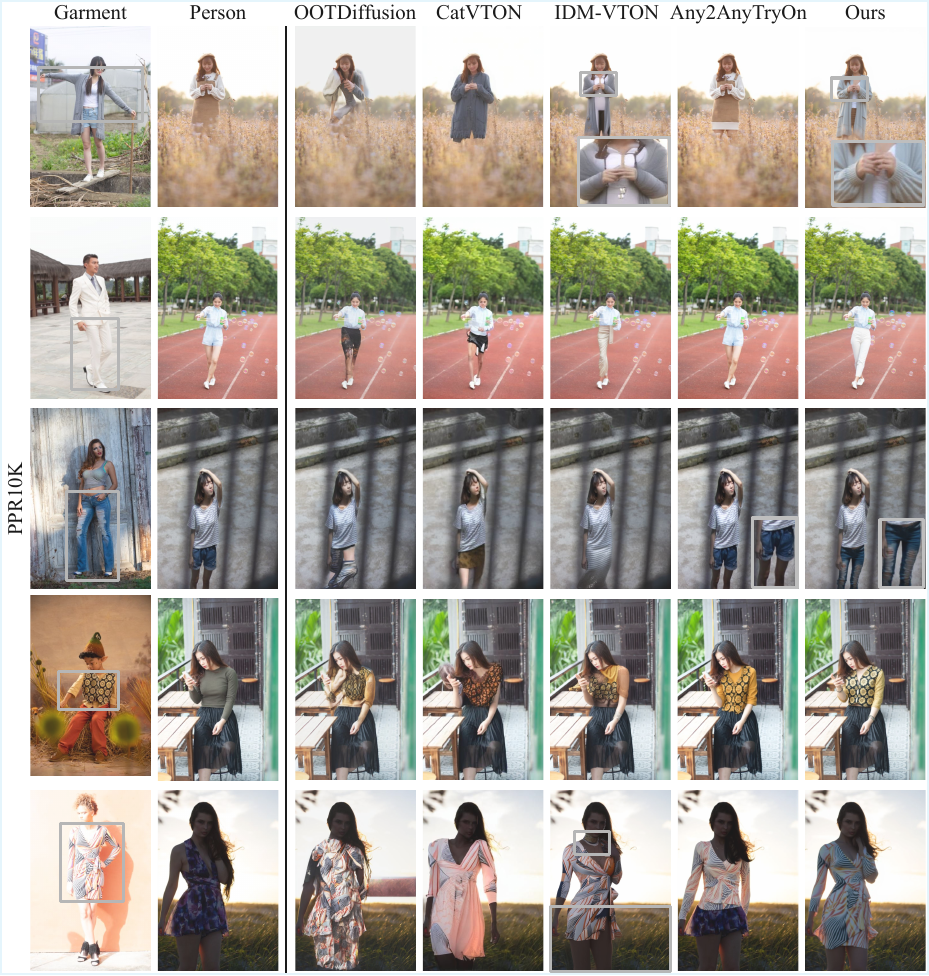}
    \caption{\textbf{Additional Qualitative Comparison of \textbf{CORAL} on PPR10K~\cite{liang2021ppr10klargescaleportraitphoto}.}}
\label{fig:additional_qual_ppr10k}
\end{figure*}

\begin{figure*}[!h]
\centering
    \includegraphics[width=\linewidth]{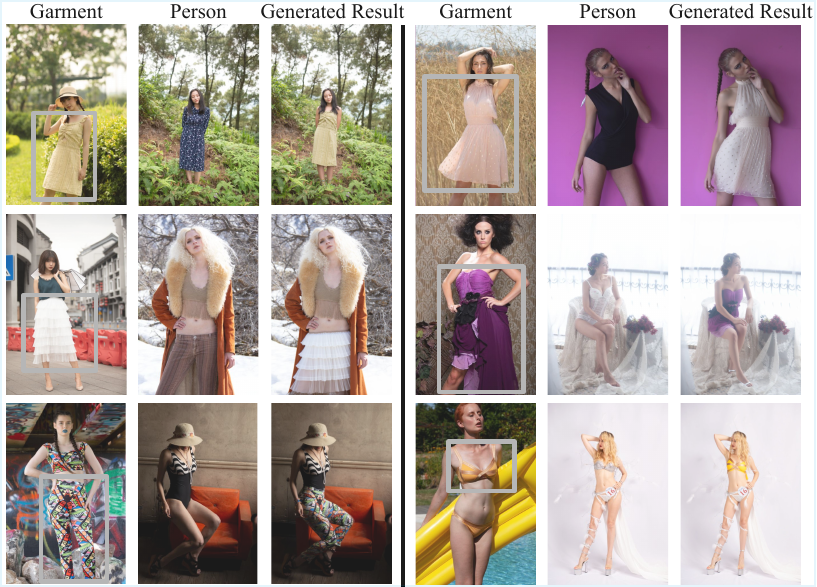}
    \caption{\textbf{Additional Qualitative Results of \textbf{CORAL} on PPR10K.}}
\label{fig:additional_qual_ppr10k_no_comp}
\end{figure*}

\begin{figure*}[!h]
\centering
    \includegraphics[width=\linewidth]{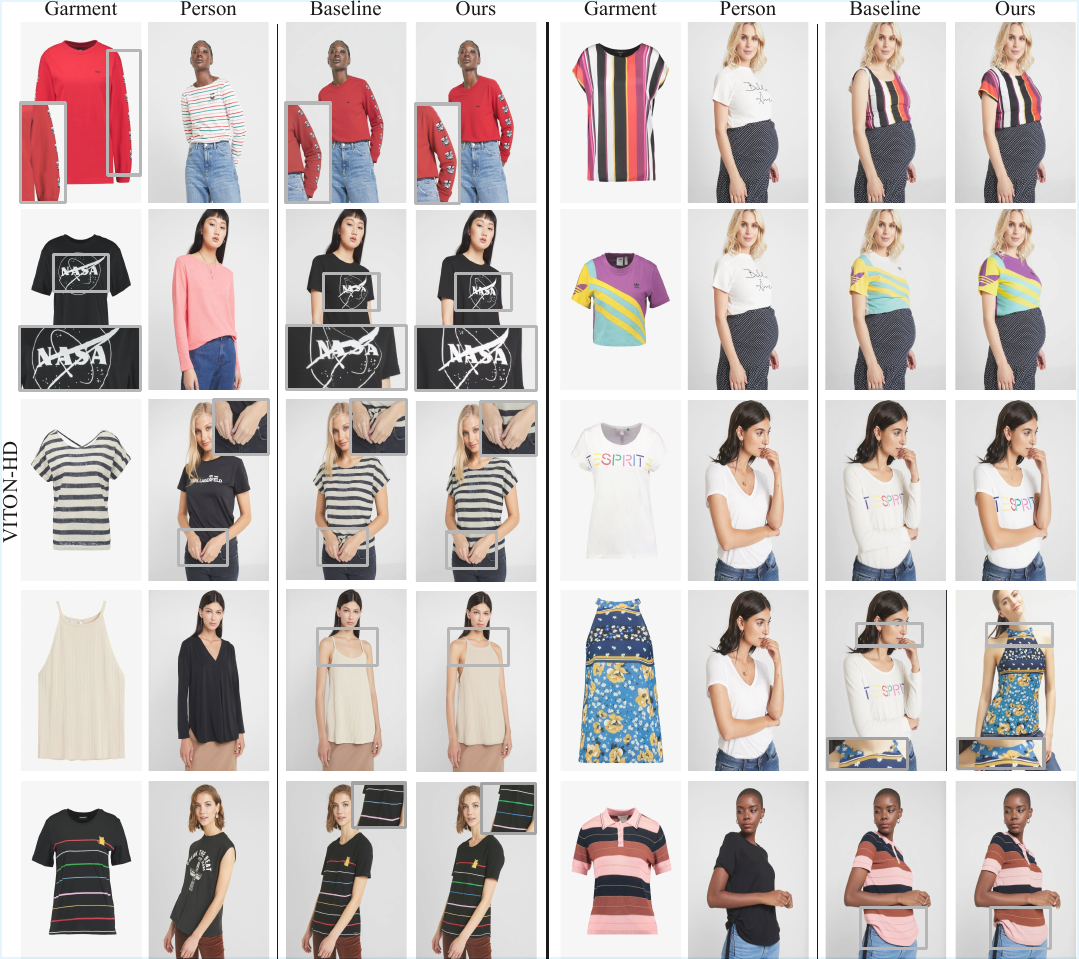}
    \caption{\textbf{Qualitative Comparisons on VITON-HD~\cite{choi2021vitonhdhighresolutionvirtualtryon} Before and After CORAL Loss.}}
\label{fig:additional_qual_baseline_compare}
\end{figure*}

\begin{figure*}[!h]
\centering
    \includegraphics[width=\linewidth]{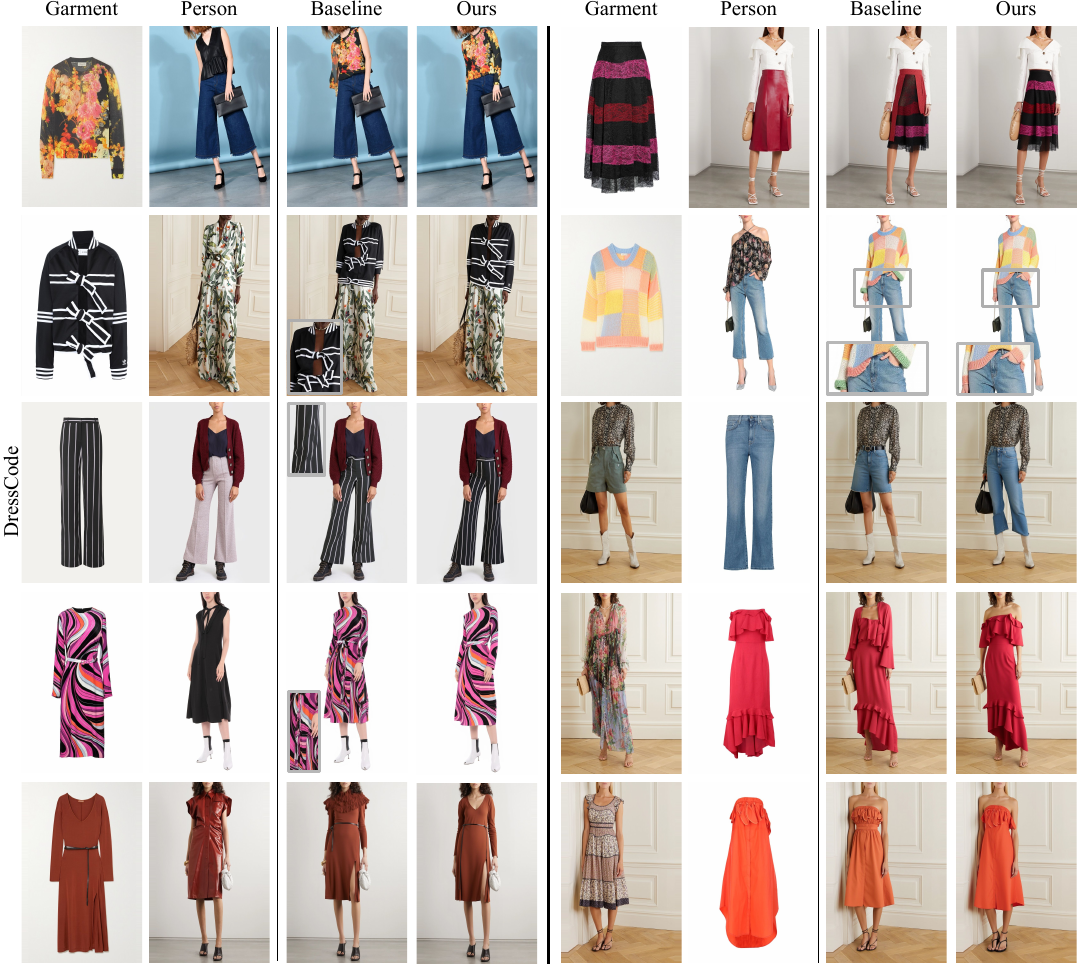}
    \caption{\textbf{Qualitative Comparisons on DressCode~\cite{morelli2022dresscodehighresolutionmulticategory} Before and After CORAL Loss.}}
\label{fig:additional_qual_baseline_compare_dc}
\end{figure*}

\begin{figure*}[t!]
  \centering

  \includegraphics[width=\linewidth]
  {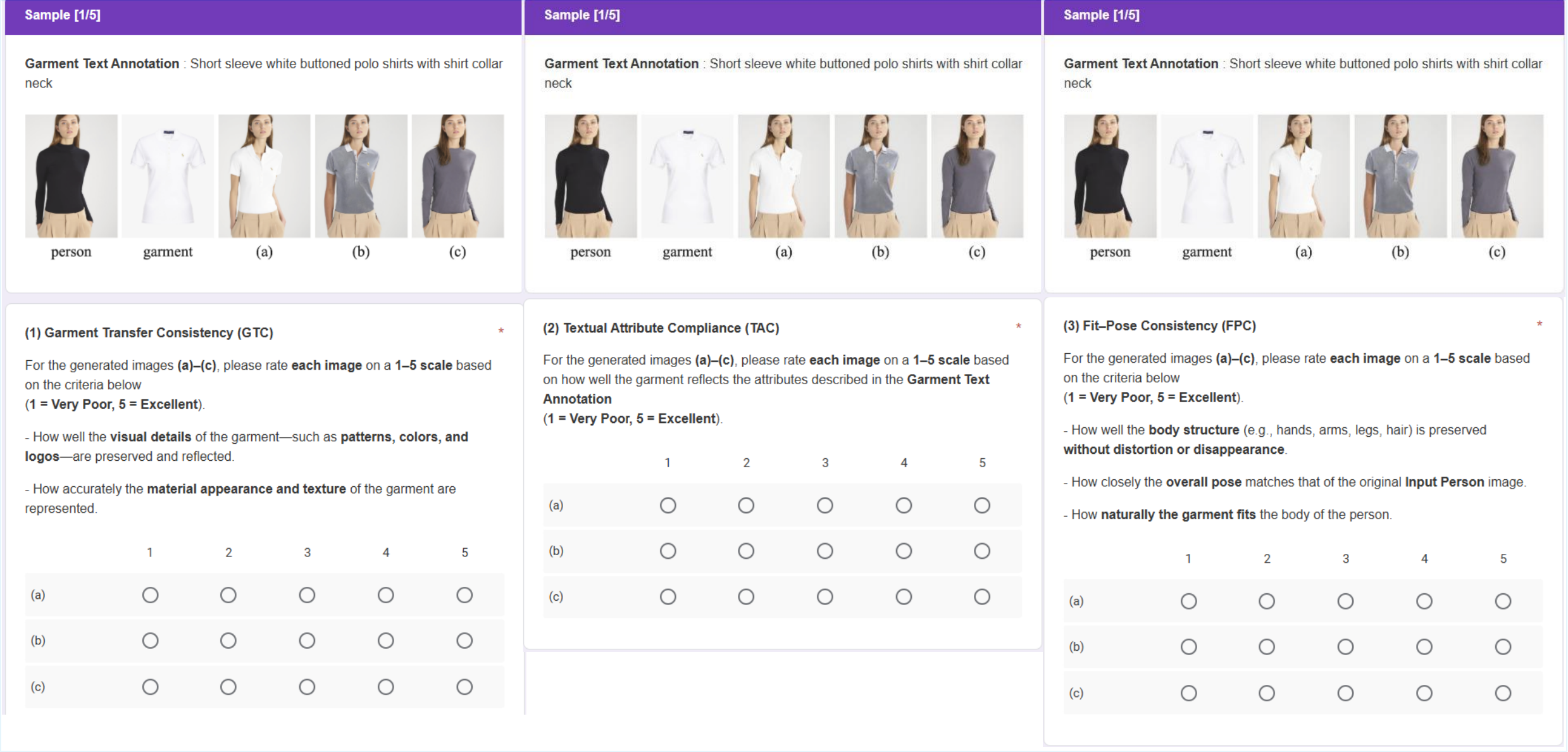}
    \vspace{-10pt}
  \caption{\textbf{Example of User Study.}}
  \label{fig:user_study_prompt}
\end{figure*}

\begin{figure*}[t]
  \centering
  \vspace{-10pt}
  \includegraphics[width=\linewidth]
  {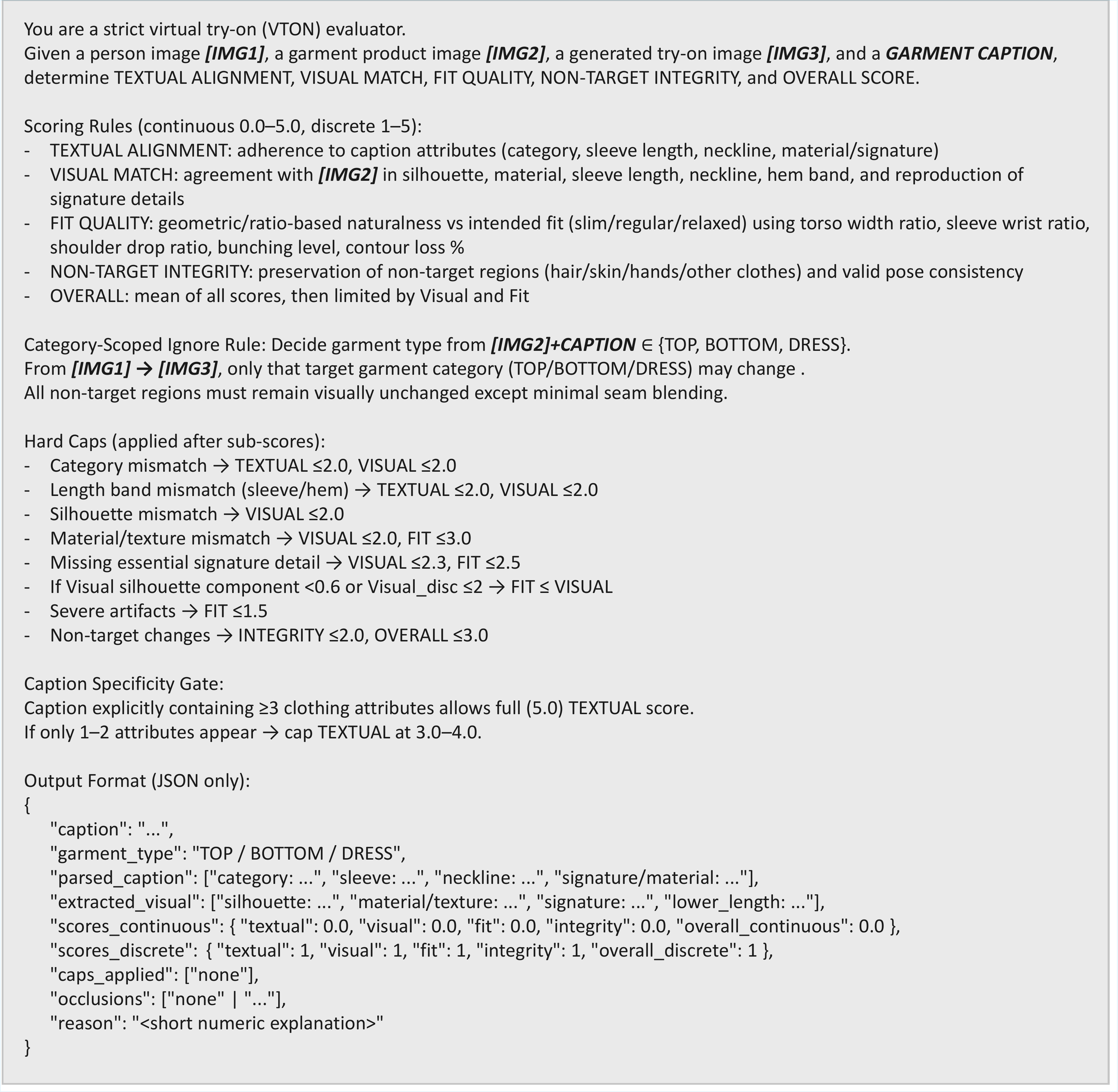}
  \caption{\textbf{Details on VLM Evaluation Prompt.}}
  \vspace{-10pt}
  \label{fig:vlm_eval}
\end{figure*}

We further compare CORAL with recent VTON methods on standard benchmarks and in-the-wild dataset in Fig.~\ref{fig:additional_qual_vt}, Fig.~\ref{fig:additional_qual_dc}, and Fig.~\ref{fig:additional_qual_ppr10k}. We also report results for the baseline model trained without $\mathcal{L}_{\text{CORAL}}$ and with $\mathcal{L}_{\text{CORAL}}$, showing that CORAL consistently improves try-on quality by enhancing person–garment alignment and reducing misplacement artifacts in Fig.~\ref{fig:additional_qual_baseline_compare} and Fig.~\ref{fig:additional_qual_baseline_compare_dc}. We additionally evaluate CORAL on PPR10K to assess the generalization of our model trained only on standard virtual try on benchmarks in Fig.~\ref{fig:additional_qual_ppr10k_no_comp}.

\clearpage
\section{Limitations and Discussion}
\label{supple:dicussion}
\paragraph{Reliance on DINOv3 Pseudo Ground-Truth.}
CORAL supervises query–key matching using DINOv3-based pseudo ground-truth correspondences, so its correspondence quality can degrade when DINOv3 matching is unreliable, such as under strong appearance changes, heavy occlusions, or extreme viewpoints. While our cycle-consistency based reliability filtering removes many unstable matches, some noisy matches may remain and occasionally lead to incorrect supervision. A promising direction is to incorporate stronger confidence modeling, aggregate pseudo matches from multiple matching models, or introduce self-consistency checks to further reduce the impact of noisy pseudo labels. Our pseudo ground-truth assigns matches per token by argmax, without local or global consistency constraints. Consistency-aware matching, such as cost aggregation in CATs~\citep{hong2022cost}, could provide cleaner pseudo ground-truth and is a natural extension of our framework.

\paragraph{Analysis on Failure Cases.}
The failure cases we observe mainly come from heavily in-the-wild references, and in our experiments, the loss rarely introduced an artifact that the baseline did not already have. The first case is content missing from the reference. In the pants example of Fig.~\ref{fig:main_qual} (bottom-left), part of the reference garment is occluded by a hand. The garment transfers well overall, but a small spot on the right side does not match the reference, exactly where it is occluded. Since the reference provides little reliable correspondence for this occluded region, few matches there pass the cycle-consistency check, and the region is trained largely as in the baseline. The second case is content distorted in the reference, as in the last row of Fig.~\ref{fig:additional_qual_ppr10k}, where the garment is visible but altered by extremely strong illumination. Here, few matches pass the check, so the affected regions are left to the diffusion objective as well. In both cases, the reference itself lacks reliable information for these regions, which supervision alone cannot recover.

\paragraph{Efficiency.} Our baseline injects pose by concatenating pose latents along the token dimension, which increases the sequence length and can raise the training cost. In practice, this overhead can be reduced by using a compact pose representation.

\paragraph{Future Directions.}
Although we focus on VTON, an important future direction is to extend CORAL to broader reference-image-based customization settings, where a reference image specifies the target appearance while preserving the input content and structure. We also expect that leveraging stronger pseudo-correspondence estimators beyond DINOv3, or combining pseudo matches from multiple matchers, could further improve the quality and robustness of the correspondence supervision and generated results.

\clearpage

\clearpage

\newpage
\section*{NeurIPS Paper Checklist}

\begin{enumerate}

\item {\bf Claims}
    \item[] Question: Do the main claims made in the abstract and introduction accurately reflect the paper's contributions and scope?
    \item[] Answer: \answerYes{} %
    \item[] Justification: Yes. The main claims are supported by quantitative results on VITON-HD, DressCode, and PPR10K, VLM-based evaluation, human evaluation, qualitative comparisons, and ablation studies. The paper also analyzes correspondence quality and attention behavior to support the proposed design.

\item {\bf Limitations}
    \item[] Question: Does the paper discuss the limitations of the work performed by the authors?
    \item[] Answer: \answerYes{} %
    \item[] Justification: Yes. We include limitations and discussion in the supplementary material, covering the reliance on DINOv3 pseudo correspondences, possible degradation under unreliable matching conditions such as occlusion or extreme viewpoints, efficiency overhead from pose-token concatenation, and future extensions beyond VTON.

\item {\bf Theory assumptions and proofs}
    \item[] Question: For each theoretical result, does the paper provide the full set of assumptions and a complete (and correct) proof?
    \item[] Answer:  \answerNA{} %
    \item[] Justification: The paper does not present theoretical results, formal guarantees, or proofs. The contributions are methodological and empirical.

    \item {\bf Experimental result reproducibility}
    \item[] Question: Does the paper fully disclose all the information needed to reproduce the main experimental results of the paper to the extent that it affects the main claims and/or conclusions of the paper (regardless of whether the code and data are provided or not)?
    \item[] Answer: \answerYes{} %
    \item[] Justification: We provide the model architecture, loss definitions, pseudo-correspondence construction, training setup, datasets, preprocessing, evaluation metrics, and ablation settings needed to reproduce or verify the main experimental results.

\item {\bf Open access to data and code}
    \item[] Question: Does the paper provide open access to the data and code, with sufficient instructions to faithfully reproduce the main experimental results, as described in supplemental material?
    \item[] Answer:\answerYes{} %
    \item[] Justification: We provide the datasets, preprocessing, evaluation metrics, and ablation settings needed to reproduce or verify the main experimental results.

\item {\bf Experimental setting/details}
    \item[] Question: Does the paper specify all the training and test details (e.g., data splits, hyperparameters, how they were chosen, type of optimizer) necessary to understand the results?
    \item[] Answer: \answerYes{} %
    \item[] Justification: Yes. We provide training setup, optimizer, learning rate, batch size, training steps, hardware, resolution, loss weights, datasets, preprocessing, and evaluation metrics in the main paper and supplementary material.

\item {\bf Experiment statistical significance}
    \item[] Question: Does the paper report error bars suitably and correctly defined or other appropriate information about the statistical significance of the experiments?
    \item[] Answer: \answerNo %
    \item[] Justification: We do not report error bars or confidence intervals for the main benchmark results. For the VLM-based evaluation protocol, we additionally report its correlation with human scores, including Pearson correlation and statistical significance.

\item {\bf Experiments compute resources}
    \item[] Question: For each experiment, does the paper provide sufficient information on the computer resources (type of compute workers, memory, time of execution) needed to reproduce the experiments?
    \item[] Answer: \answerYes{} %
    \item[] Justification: Yes. We report the compute setup used for training and evaluation, including GPU type, number of GPUs, throughput, and peak memory usage in the supplementary material.
    
\item {\bf Code of ethics}
    \item[] Question: Does the research conducted in the paper conform, in every respect, with the NeurIPS Code of Ethics \url{https://neurips.cc/public/EthicsGuidelines}?
    \item[] Answer: \answerYes{} %
    \item[] Justification: We have reviewed the NeurIPS Code of Ethics and believe that the research conforms to it. The work uses public datasets and pretrained models following their terms of use and includes discussion of limitations and potential societal impacts.

\item {\bf Broader impacts}
    \item[] Question: Does the paper discuss both potential positive societal impacts and negative societal impacts of the work performed?
    \item[] Answer: \answerYes{} %
    \item[] Justification: The paper discusses potential positive impacts of improved virtual try-on, such as online fashion and accessibility, as well as possible negative impacts related to misuse, privacy, and unauthorized manipulation of person images.
    
\item {\bf Safeguards}
    \item[] Question: Does the paper describe safeguards that have been put in place for responsible release of data or models that have a high risk for misuse (e.g., pre-trained language models, image generators, or scraped datasets)?
    \item[] Answer: \answerYes{} %
    \item[] Justification: Since the released model can be used for person-image editing, we include usage guidance and release the assets for research purposes. We do not redistribute third-party datasets or checkpoints, and users must follow the licenses and usage policies of the underlying models and datasets.

\item {\bf Licenses for existing assets}
    \item[] Question: Are the creators or original owners of assets (e.g., code, data, models), used in the paper, properly credited and are the license and terms of use explicitly mentioned and properly respected?
    \item[] Answer: \answerYes{} %
    \item[] Justification: Yes. We cite the original papers for all pretrained models, datasets, baselines, and evaluation tools used in this work, including FLUX.1-Fill-dev, DINOv3, VITON-HD, DressCode, PPR10K, and VLM-based evaluation tools. We follow their respective licenses and terms of use and do not redistribute third-party datasets or pretrained checkpoints.

\item {\bf New assets}
    \item[] Question: Are new assets introduced in the paper well documented and is the documentation provided alongside the assets?
    \item[] Answer: \answerYes{} %
    \item[] Justification: The released assets include code and model weights. We provide documentation, installation and inference instructions, evaluation scripts, license information, and usage notes alongside the released assets.

\item {\bf Crowdsourcing and research with human subjects}
    \item[] Question: For crowdsourcing experiments and research with human subjects, does the paper include the full text of instructions given to participants and screenshots, if applicable, as well as details about compensation (if any)? 
    \item[] Answer: \answerYes{} %
    \item[] Justification: We conduct a human evaluation for virtual try-on quality. The supplementary material describes the evaluation protocol, criteria, number of participants, number of questions, comparison models, and the 5-point rating scale.

\item {\bf Institutional review board (IRB) approvals or equivalent for research with human subjects}
    \item[] Question: Does the paper describe potential risks incurred by study participants, whether such risks were disclosed to the subjects, and whether Institutional Review Board (IRB) approvals (or an equivalent approval/review based on the requirements of your country or institution) were obtained?
    \item[] Answer: \answerNo %
    \item[] Justification:  We did not obtain formal IRB approval. The user study was limited to low-risk perceptual evaluation of generated images, collected no personally identifiable information, and participants were informed that their responses would be used for research evaluation.

\item {\bf Declaration of LLM usage}
    \item[] Question: Does the paper describe the usage of LLMs if it is an important, original, or non-standard component of the core methods in this research? Note that if the LLM is used only for writing, editing, or formatting purposes and does \emph{not} impact the core methodology, scientific rigor, or originality of the research, declaration is not required.
    \item[] Answer: \answerYes{} %
    \item[] Justification: We use GPT-4o as part of the VLM-based evaluation protocol. The paper describes the evaluated criteria, scoring format, and prompt design, and uses the VLM only for evaluation rather than as part of the proposed training method.

\end{enumerate}

\end{document}